**OpenPodcar2: a robust, ROS2 vehicle for self-driving research**

Rakshit Soni

Chris Waltham

Md Umar Ibrahim

Mark Crampton

Charles Fox

School of Engineering and Physical Science, University of Lincoln, UK

chfox@lincoln.ac.uk

**Abstract**

OpenPodcar2 is a robust, ROS2-interfaced, low-cost, open source hardware and software, autonomous vehicle platform based on an off-the-shelf, hard-canopy, mobility scooter donor vehicle. It is a modification of the previous OpenPodcar design, which extends it with robust electronics and ROS2 interfacing, to enable both research and also potential deployment use cases. The platform consists of (a) hardware components: documented as a bill of materials and build instructions; (b) integration to the general purpose OSH R4 mechatronics board and a Gazebo simulation of the vehicle, both presenting a common ROS2 interface (c) higher-level ROS2 software implementations and configurations of standard robot autonomous planning and control, including the nav2 stack which performs SLAM and enacts commands to drive the vehicle from a current to a desired pose around obstacles. OpenPodcar2 can transport a human passenger or similar load at speeds up to 15km/h, for example for use as a last-mile autonomous taxi service or to transport delivery containers similarly around a city center. It is small and safe enough to be parked in a standard research lab robust enough for some deployment cases. Total build cost was around 7,000USD from new components, or 2,000USD with a used Donor Vehicle. OpenPodcar2 thus provides a research balance between real world utility, safety, cost and robustness.

**Metadata overview**

| | |
|---|---|
| **Main design files:** | https://github.com/Rak-r/OpenPodcar2 [will move to zenodo on paper acceptance] |
| **Target group:** | Autonomous vehicle researchers |
| **Skills required:** | Mechanical and electronics components assembly; electrical testing; drilling steel; soldering; crimping. Skilled technician level. Drilling steel may require safety certification. |
| **Build cost:** | 7000 USD (from all new components); 2000 USD (with a used donor vehicle) |
| **Replication:** | The vehicle has been disassembled and reassembled by a new skilled technician not involved the original design using the draft build instructions; answers to questions asked during the build have been incorporated into the final build instructions. |
| **Keywords:** | autonomous vehicle, self-driving car, open source hardware, R4, robotics |

# Introduction

Autonomous Vehicles (AVs), are researched in academic robotics and increasingly deployed by industry. Both academic research and industrial deployment now require long-term operation, going beyond short experimental runs to robust operation without failures over days, weeks or months. In research robotics, long-term operation is especially valuable in enabling rigorous testing of adaptation models and techniques, such as on-line learning of pedestrian and traffic flow patterns and in variances to weather and lighting conditions. Sustained long-term operation requires not only advances in vehicle hardware and sensing but also a robust and modular open software infrastructure. Such an infrastructure must support continuous execution, standardized interfaces between system components, and reproducible integration across hardware platforms and research environments.

Robot operating system (ROS2) open source software ecosystem has developed to enable robust automation over long-terms, and includes an underlying message-passing system upon which a larger collection of increasingly standard AI components are stacked, including the navigation stack, `nav2`. Like ROS2, `nav2` has evolved from a previous version 1 research system into a robust tool for long-term industrial use. It includes software implementations of standard mapping, localization, planning and control robotics algorithms. ROS2 and `nav2` together enable robotics research to explore alternate algorithms for each of these stages while holding the rest constant, and to quickly and easily reproduce work between labs in software and simulation.

However, reproduction of research remains harder for hardware robotics. A ROS2-based simulation can be open sourced, downloaded and run in another lab, but labs lack standard hardware upon which to perform more realistic evaluation of these algorithms. While ROS2 provides some portability between different physical robots – such as its use of a standard Twist command as an interface for controllers to output and vehicles to implement – different robots have different capabilities to implement these standards which lead to different measured results when experiment reproduction is attempted. Hardware companies have promoted closed-source products as attempted hardware standards, but do not provide design information for research modification or repair, and even if any of these was adopted unmodified, it would risk vanishing if the company discontinued the product or raised prices after standardisation.

OpenPodcar [12] is a previous OSH project which enables labs to standardize on an open design. It is based on a cheap and widely available closed-source donor vehicle – a hard-canopy mobility scooter, and its design is an modification which converts the donor for autonomous driving. OpenPodcar is intended for short-term operation for experiments lasting a few minutes or hours. It uses ROS1 software, and relatively simple electronics. It is suitable for short-term experiments, but significant extension is required to achieve robustness for long-term operation.

OpenPodcar2 extends OpenPodcar's design for robust long-term operation. It replaces OpenPodcar's entire electronics system with new electronics based on the open source hardware R4 [33]. R4 (Rapid Reproducible Robotics Research) is a general purpose medium-sized robotics (i.e. robots capable of transporting a human or similar load) control board, developed to high robustness. It provides a single firmware system and ROS2 interface, which can then be programmed entirely in ROS2 without end-users having to write any embedded code. It replaces the Donor Vehicle's proprietary, wigwag-based motor driver with open source hardware OSMC [5], enabling deep openness and understanding of the design. This simplifies safety, as wigwag relies on a potentially dangerous non-zero voltage to represent no motion, with zero volts for fast reverse, which can engage if voltage is lot due to technical failures, unless complex mitigations are used as in the Donor Vehicle's proprietary, undocumented driver. OpenPodcar2 also replaces OpenPodcar's entire ROS self-driving software stack with a new ROS2 stack, which are interfaced to R4 electronics and which are robust at the software level for long-term driving – ROS2 being targeted at robust industrial robotics. OpenPodcar2 replaces all physical connectors used in OpenPodcar with new robust connectors able to withstand tilts and vibrations of the vehicle. Unlike OpenPodcar, OpenPodcar2 is able to operate in rain. OpenPodcar2 replaces OpenPodcar's expensive lidar perception system with a cheaper depth camera based system, able to detect and track pedestrians with similar accuracy, using open-source YOLOv8 deep learning software. Figure 1 shows the new electronics

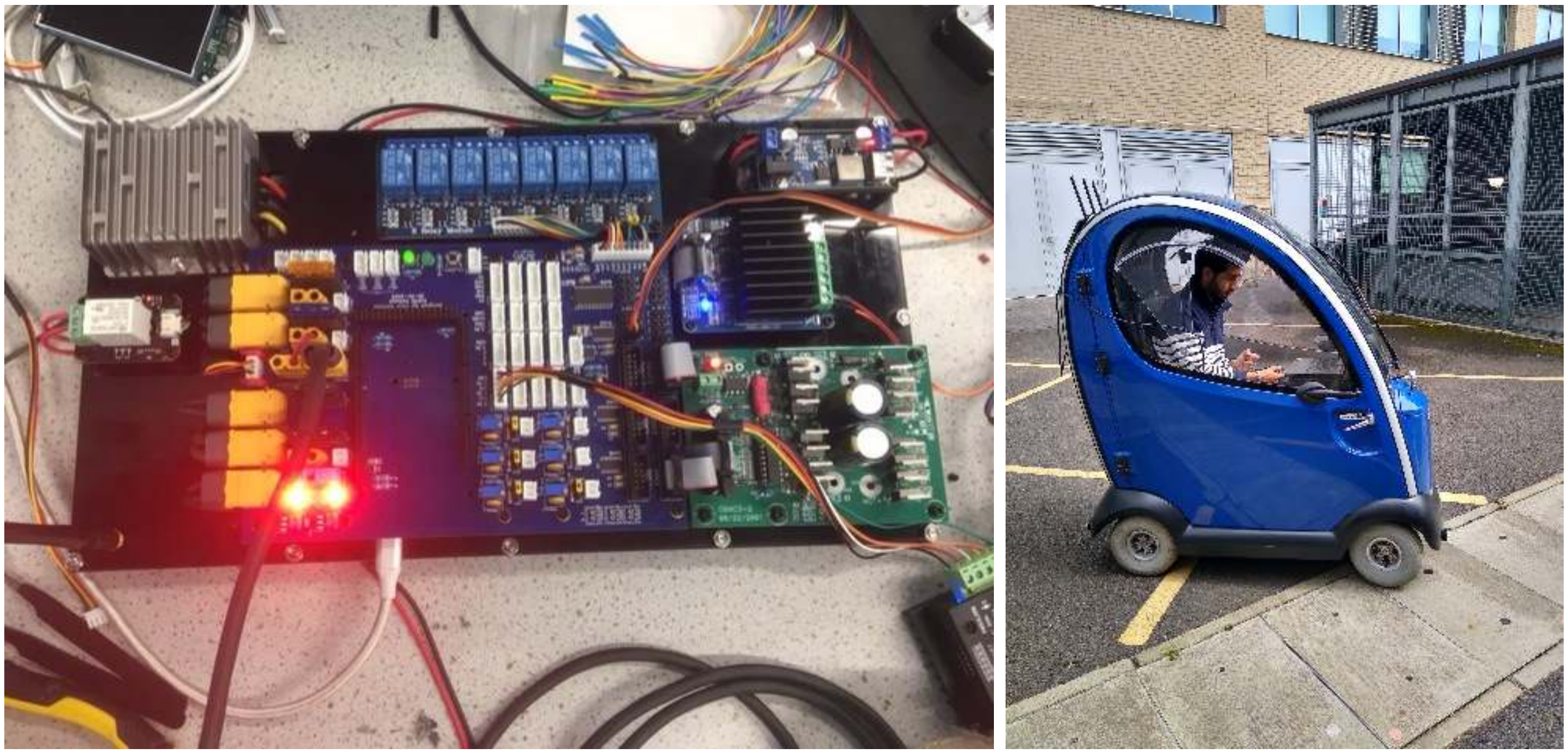

Figure 1: R4 interfaced to drivers and relays on the Mounting Board (left) and used to control OpenPodcar2 (right)

assembly and the complete OpenPodcar2 vehicle in action.

Related complete and built mechanical OSH designs for on-road, person-carrying cars exist, including PixBot [28], Tabby EVO [27] and the Autoware-interfaced iseAuto [30]. Building these full-size cars is a large task for experts and may require highly skilled processes such as welding, purchase of expensive components, and considerable storage and fabrication space. OpenPodcar is based on a proprietary but commodity mobility scooter which is cheaper and easier to convert than performing these builds. Several OSH RC-scale cars have been completed and built such as F1Tenth [1], AutoRally [14], BARC [15], MIT Racecar [2], MuSHR [3], [25], and [32], while similar sized OSH robots for education, research and survey include ROMR [26], ROBOTONT [29], 'Open source and open hardware mobile robot for developing applications in education and research' [7], and OpenScout [13]. These platforms are not large enough to drive on public roads or to transport people or goods like OpenPodcar. Open Source Ecology (OSE) [17] is an ambitious programme of projects which ultimately aims to develop fully OSH vehicles including a car and tractor. OSE is optimised for reliability and for users in developing countries so it uses hydraulic power rather than electric as used in OpenPodcar. But its vehicle designs are not yet complete. Autoware [18] is a heavyweight open source software project to construct a full ROS-based automation stack for on-road cars. Apollo [6] is an open source self-driving software stack and an open hardware interface which may be implemented on vehicles, as done in [19]. These systems could be software interfaced to run with OpenPodcar and OpenPodcar2.

# Overall implementation and design

## Mechanical vehicle

All original wiring and motor driver are removed from donor vehicle. The batteries are the only electrical components that are retained. (This cleaner approach contrasts with the previous OpenPodcar which relied on interfacing to parts of the donor vehicle's closed source electronics.)

Automated steering of the front wheels is performed by a linear actuator on the underside of the vehicle which moves the existing steering rod. Updating the OpenPodcar design, a new shorter actuator is used, whose motion is parallel with the steering rod.

The steering column is removed and replaced with a laptop stand for use by the passenger.

## Physical R4 interface

New electronics and the hardware-software interface are based on the R4 [33], an existing, peer-reviewed open source hardware general purpose medium-sized robot control board. R4 provides a standardized ROS2 interface to a variety of brushed and brushless motor controllers, sensors, and dead man's handle. It has previously been used as a component in the small mobile open source hardware robot, OpenScout [13], and its reuse here encourages portability of software across other R4-based robots. (For example, enabling OpenPodcar2's new automation stack to be easily ported to OpenScout.)

R4's user data-gram protocol (UDP) based communication interface connects the physical control board to the high-level ROS2 stack, enabling both manual teleoperation and autonomous navigation using `nav2`. R4 provides three sets of ROS2 hardware nodes: `/R4_Websockets-Clients` which is responsible to establish the connection to transfer the data packets from the R4 to ROS2 over the topic named `/R4` which gets unpacked by the second node `/R4_Publisher` to publish the individual component sub-messages mainly, steer voltage reading, and motor driver readings over the topics `/R4_AINSTEER` and `/R4_OSMC1` which are subscribed by ROS2 nodes to control the steering and speed of OpenPodcar2 via a gamepad for manual teleoperation and via ROS2's `nav2` for autonomous driving tasks. The output of either manual teleoperation or `nav2` is same and published over the standard ROS2 message type Twist over `/cmd_vel` topic and this information is then subscribed by the third node named as `/R4_Receiver` as illustrated in Figure 2, which transports the data back to hardware board to actually follow the requested commands.

A wifi router is included onboard the vehicle, and runs its own local network. The R4 communicates via UDP packets on this network, with a host computer which may be onboard the vehicle, such as a laptop, Raspberry Pi, and/or larger machines located physically separate from the vehicle.

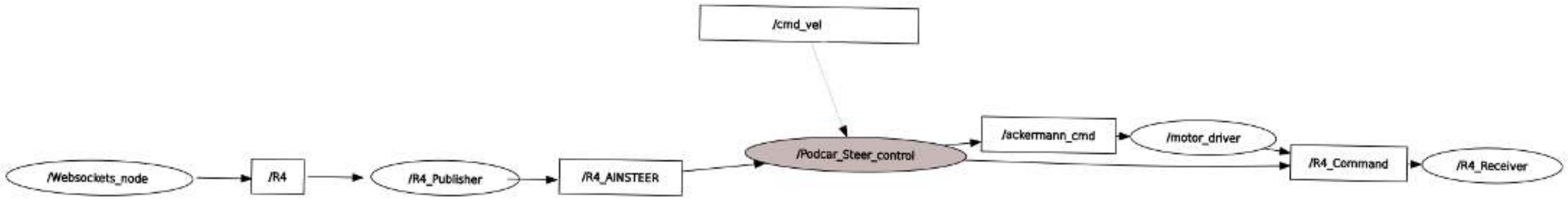

Figure 2: Podcar steer control node for physical podcar, wrapping R4 ROS2 nodes and reading cmd_vel

## Layered safety architecture

Safety is of the utmost importance for autonomous driving experiments, and OpenPodcar2's safety systems are designed to meet the formal Health and Safety and Research Ethics requirements of a UK university for autonomous driving around members of the public on its campus.

A *dead man's handle (DMH)* is included which is a last-resort immediate power cutoff to the main drive motor via a hardware relay. The DMH is independent of ROS2 and of R4 firmware, and instead breaks the power directly at hardware level. The experimenter holds a DHM trigger in their hand during all operation and letting go of this cuts off the power immediately for safety. The experimenter may either sit in the vehicle as a passenger or walk some distance away from it with the DHM wire between them.

The motor of the donor vehicle comes with an electromagnetic brake attachment, activated by the *absence* of any current sent to it. This mechanical safety system is retained from the original vehicle.

A higher-level *e-stop* is provided, which can implement more graceful shutdowns when all low-level systems are still working. The e-stop is a UDP signal transmitted to R4 from its ROS2 driver node, in response to any loss of the ROS2 network.

In manual drive mode, a secondary ROS2-level dead man's handle (DMH2) runs on one of the manual controller buttons, to prevent accidental control of the vehicle if the controller is unintentionally contacted.

In autonomous drive mode, the mapping, planning, and pedestrian detection algorithms work together to create ROS2-level plans which avoid hitting obstacles by steering around them, and commanding zero velocity then reversing and re-planning behaviours if an imminent collision is predicted.

### Sensors

OpenPodcar2 replaces the expensive lidar sensor of OpenPodcar with a cheaper single RGB-D camera, which is processed in such a way as to provide lidar data. (LiDAR can still be used if needed for higher accuracies, with some reconfiguration.) The depthcam is mounted on the front of the vehicle publishes RGB and depth image data as standard ROS2 `Image` messages, along with pointcloud data over `PointCloud2` messages. `PointCloud2` messages are produced directly by the RGBD camera, as for lidar sensors. These are formatted as a 2D grid of 20 byte chunks, ordered as the depth cam image. Each chunk contains the 3D spatial location of the contact in the camera's frame, and its RGB color. The `pointcloud_to_laserscan` library node is used to project it all into a 1D `LaserScan` message, including filtering out contacts on the floor.

### Software

OpenPodcar2 is supplied with a new self-driving software stack based on ROS2, detailed in the repository [And copied as Appendix D for review]. The stack is built from a curated selection of ROS2 ecosystem components, connected and configured for OpenPodcar2 and interfacing via R4's ROS2 interface.

The stack provides autonomous localisation and mapping using RTAB-Map, a 3D voxel map methods suited to RGBD camera sensing. Planning is provided by the SMAC planner from the nav2 ROS2 package. Pedestrian detection and tracking is provided by a YOLOv8 configuration.

Manual driving is also provided at the ROS2 level, from an XBox gamepad, with OpenPodcar2-specific conversion nodes provided to translate standard ROS2 joy messages into OpenPodcar2 Ackermann geometry and then to R4 ROS2 command messages.

A simulation of OpenPodcar2 is provided using the same ROS2 interface, implemented in the ignition Gazebo simulator.

## Quality control

### Safety

The design is based on a layered safety architecture. This includes a human-held dead man's handle as a last resort power cutoff. Above this, the R4 provides a heartbeat mechanism which checks for local and network device connectivity and cuts power if not present. At the higher ROS2 level, an e-stop is provided in manual gamepad control and the automation stack includes stopping on obstacle detection.

Overheating of the OSMC motor driver was found to be an issue at high speeds. The design includes a maximum speed cutoff which is chosen to prevent this. This was set by gradually increasing the OSMC power to find the failure point.

### Calibration

The Build Instructions include calibration processes for the depth cam and steering. Depthcam calibration is to position it accurately to align its images and ROS2 coordinates. Steering callibration is to map from desired ROS2 steering commands to physical wheel positions, which is affected by small build differences and by the vehicle's tyre state.

## General testing

To measure quality of mapping in a large environment, several SLAM tests were performed which demonstrate good quality maps and navigation. The maps of tight indoor lab space, large indoor corridor space and outdoor SLAM test are built using RTAB-Map SLAM package using visual odometry input from RGBD odometry from RTAB-Map. Figure 3(left) shows a SLAM map built using OpenPodcar2, driving around the first floor of a university building including a lab and several tight corridors. The SLAM system has also been tested with outdoor mapping in Figure 3(right). Both maps are close matches to real life, and include successful loop closures.

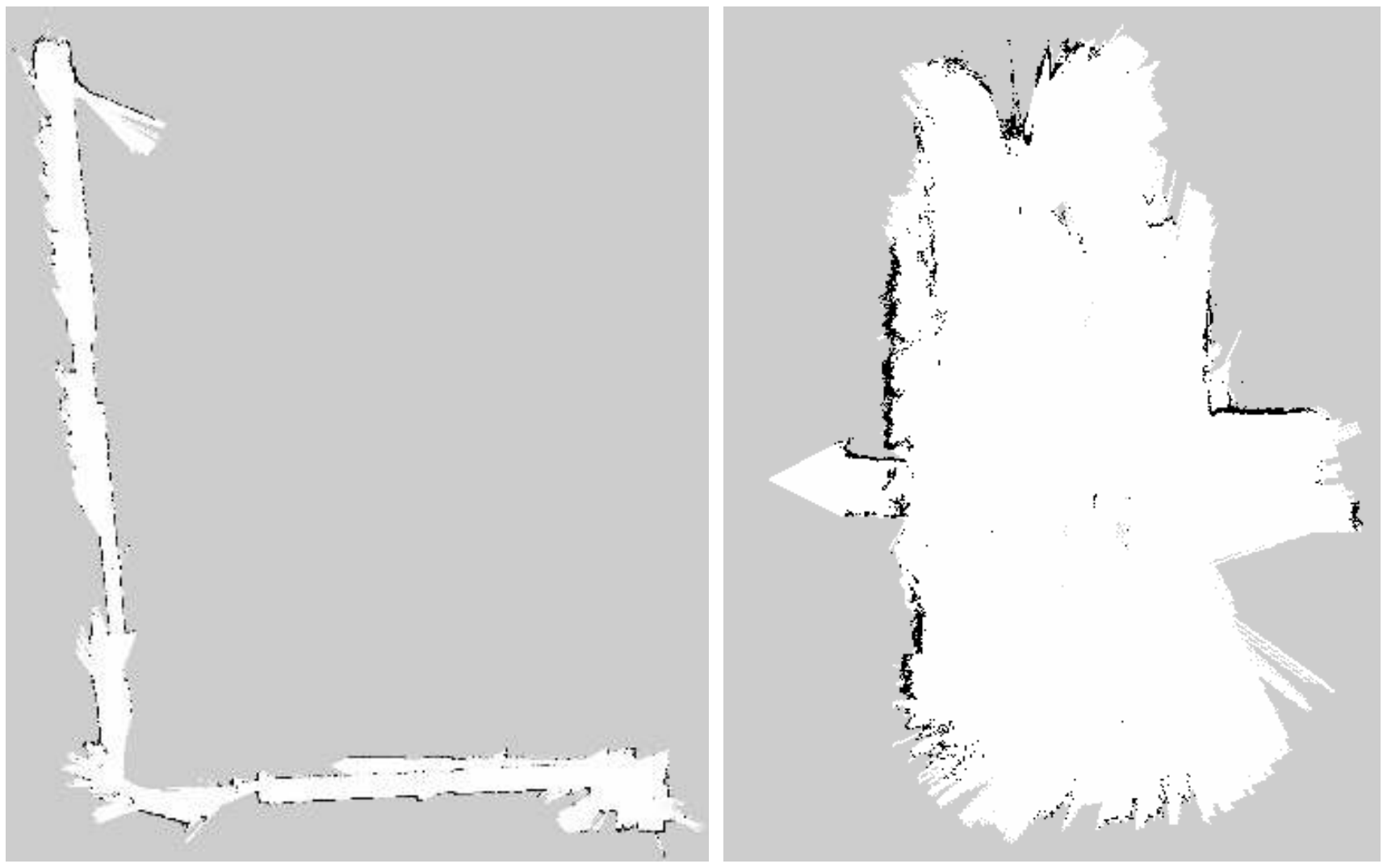

Figure 3: Successful RTAB SLAM indoor (left) and outdoor (right) 2D occupancy maps

To check the accuracy of the detection and tracking system using the RGBD camera, the resulting measurements of 3D coordinates were verified manually by placing pedestrians at known distances and monitoring the /yolo/detections topic. Measurements were within 200mm deviation, when tested indoors and outdoors with both static and dynamic pedestrians. Figure 4 shows a pedestrian tracked from RGBD sensor along with track history in the map frame.

Forward goals (inclusive of sharp turns) and near to straight reverse goals were achievable while performing obstacle avoidance using local costmap. In autonomous mode, OpenPodcar2 was tested with both with a pre-built map and SLAM mode. The long indoor area was explored while navigating autonomously as shown in Figure 5 and Figure 6, a 3D map of the same environment build using RTAB-Map.

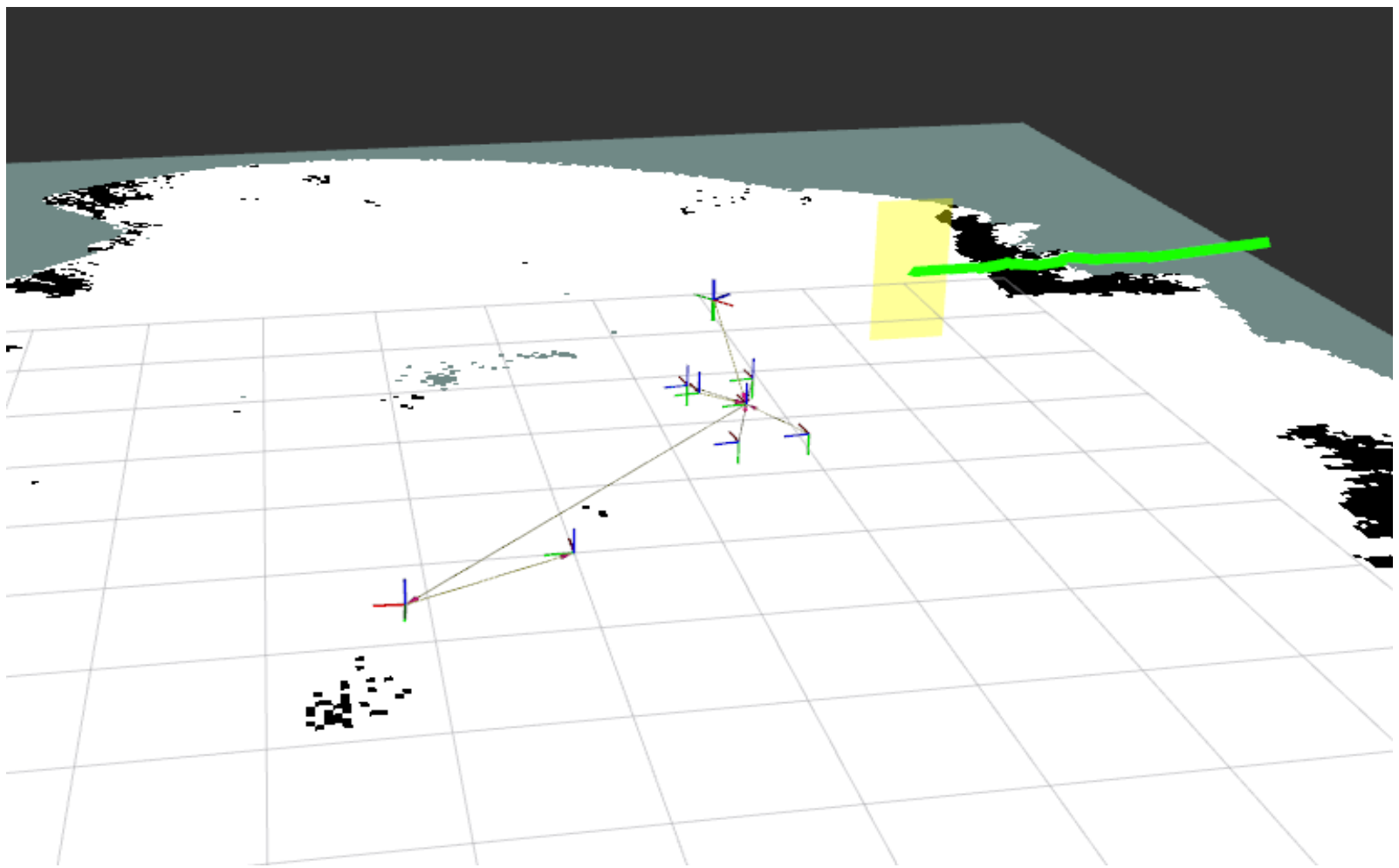

Figure 4: Pedestrian represented by 3D bounding box (yellow) with track history (green) crossing in front of the vehicle in rviz2 (map frame)

Accuracy of the distance to goal for long indoor plans was verified using a measuring tape, resulting in an reaching the final goal point with a tolerance of approximately 300-400 mm. This tolerance is due to the safety constraint imposed by the goal tolerance parameter and is measured from the center of the vehicle chassis. Three-point turn goals have been used to test the vehicle's ability to mimic car-like behavior. For forward goals, the vehicle followed the global plan but showed significant deviations on turns. In contrast, reverse goals (which involved executing a reverse maneuver followed by a forward and turn) were challenging for the DWB controller. DWB parameters namely; vx_samples and vtheta_samples to benefit the controller to generate more feasible trajectories but utilizes more computation, movement_time_allowance to provide more time for smooth motion. Another parameter to look for is sim_time which controls the forward simulation time for trajectory generation, and by increasing the number more longer trajectories are considered by the controller.

Parallel parking goals have been achieved with slight deviations in orientation. This slight deviation may be attributed to the over-predictive behavior of the DWB controller. Videos results from manual and autonomous drives are included in the project repository.

# Application

## Self-driving research

Many autonomous vehicle (AV) research groups face financial barriers when acquiring full-scale self-driving hardware platforms. OpenPodcar2 is designed to address this limitation by providing a low-cost, fully open-source hardware and software platform for AV research and education.

A key design decision is the replacement of expensive LiDAR sensors with a cost-effective RGB-D camera, while maintaining compatibility with standard LiDAR-based navigation pipelines through depth-to-laserscan conversion. This significantly reduces system cost without sacrificing integration with established ROS2 and Nav2 components.

Unlike many open-source AV software stacks that lack an associated physical reference platform, OpenPodcar2 provides openly documented hardware designs, integration details, and system architecture. This enables researchers to reproduce experiments on a real vehicle, extend the platform with additional sensors (e.g., supplementary RGB-D cameras, LiDAR units, IMUs, or GNSS modules), and modify the sensing configuration to achieve wider fields of view or higher perception redundancy.

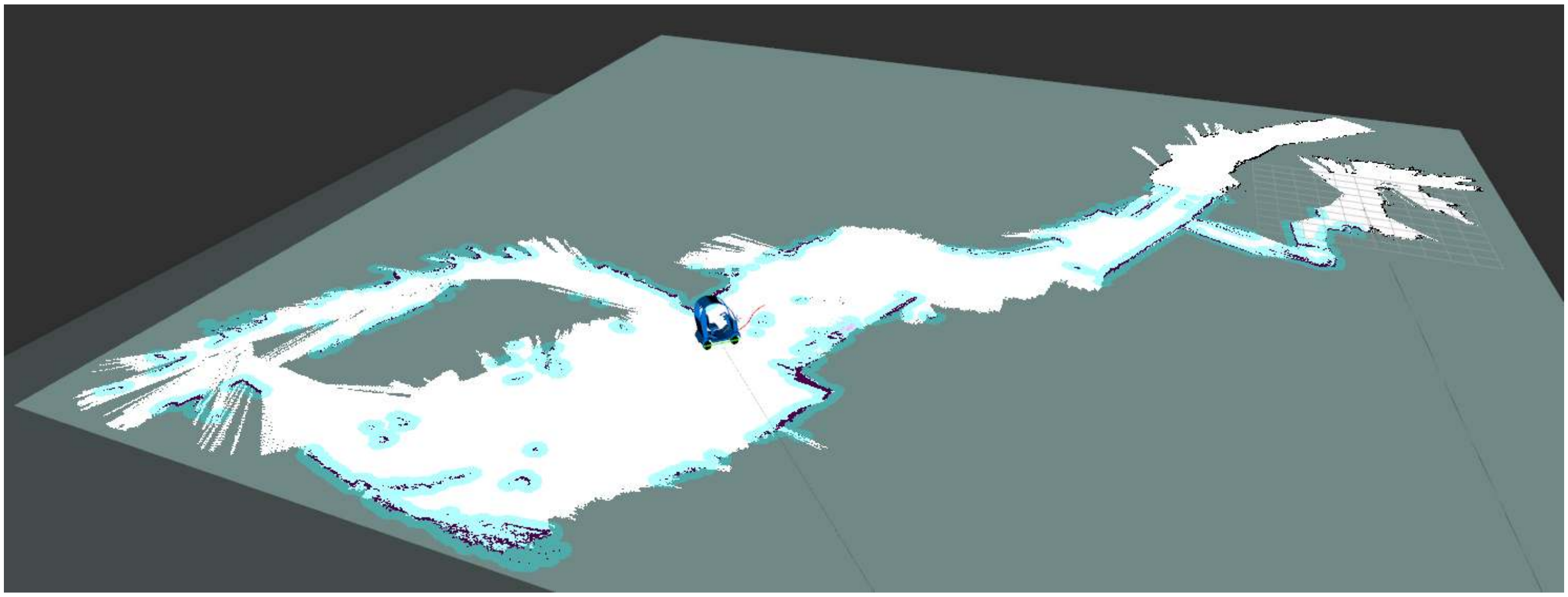

Figure 5: RTAB SLAM large indoor map with NAV2 Stack, following a successful large loop closure

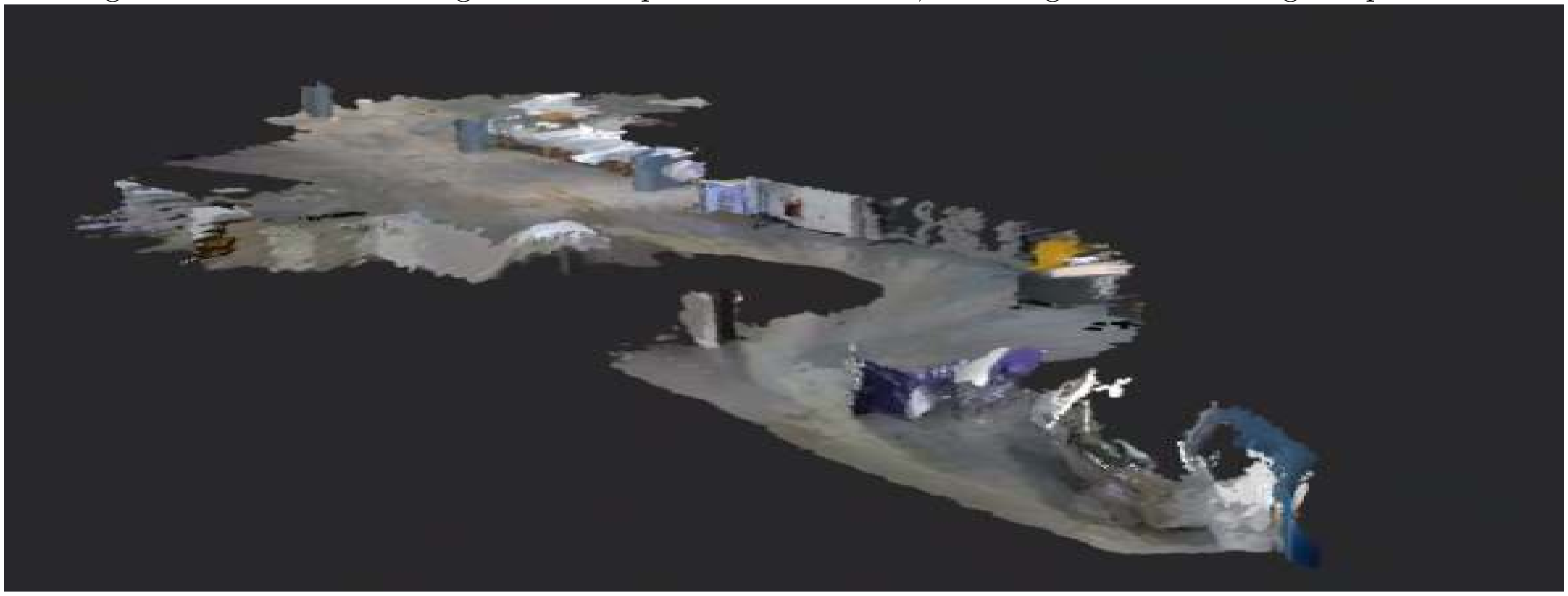

Figure 6: 3D map of the large indoor environment

The platform therefore supports experimental evaluation of localization, SLAM, planning, control, parking, and human-aware navigation algorithms under real-world conditions. By enabling repeatable testing on an accessible physical system, OpenPodcar2 contributes toward lowering the entry barrier to AV experimentation and supports reproducible research in autonomous driving.

## Human-robot-interaction

Both autonomous vehicle and social robotics research communities have interests in modeling human decision-making and interaction dynamics [8]. Game-theoretic approaches have been proposed to model strategic interaction between pedestrians and vehicles.

Previous empirical investigations of pedestrian–vehicle interaction were conducted in controlled laboratory settings [24, 9]. However, these studies reported limited ecological validity, as participants did not consis-tently exhibit realistic interaction behaviors. Subsequent experiments in immersive virtual reality environments demonstrated improved behavioral realism [10], and additional models incorporating human proxemics (interpersonal distance regulation) were developed to enhance interaction modeling [11].

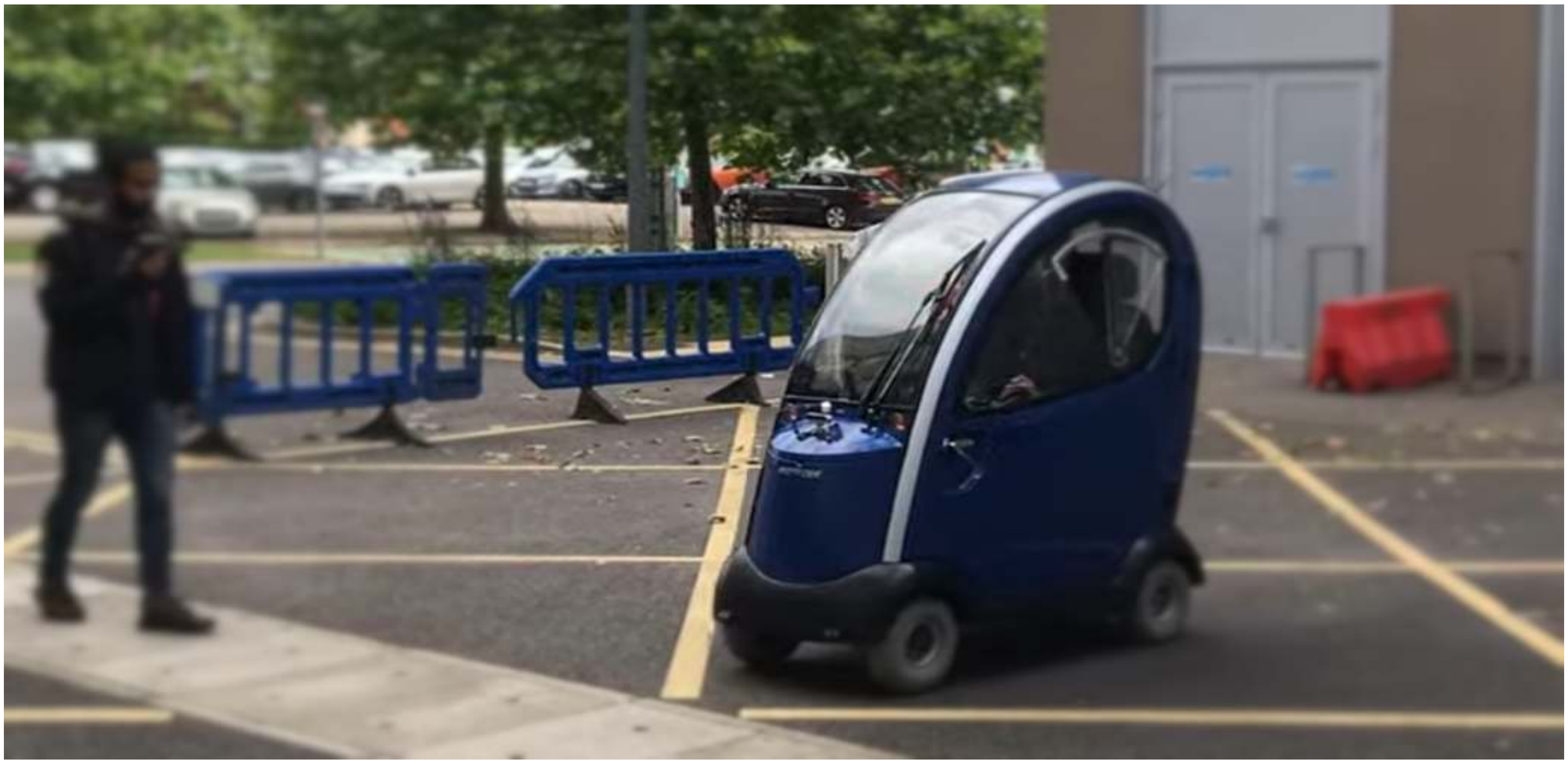

Figure 7: Real pedestrian crossing experiment with Openpodcar2.

OpenPodcar2 can bridge the gap between this virtual experimentation and real-world human-subject research. By deploying a previously developed game-theoretic interaction model into a functional autonomous system, the platform has enabled controlled physical experiments [31]. This study is among the first to demonstrate pedestrian–vehicle interactions using a completely open-source autonomous hardware architecture. Figure 7 showcases the interaction of OpenPodcar2 with real human subject for decision making.

# Build details

## Availability of materials and methods

All components are available on the open market and supplier links are provided in the bill of materials (BoM). The BoM includes licensing information for both interfaces and implementations of all components. The donor vehicle and some components are closed implementations but all present open interfaces. Where possible, formalized open standard interfaces such as ISO are used.

## Ease of build / design decision

The design is intended to be buildable by a skilled technician. The build includes drilling holes in steel which typically requires training and safety certification.

### Operating software and peripherals

The software is open source under GPL license and runs on Open Source Ubuntu Linux. Peripherals include an Intel depth camera and a wifi router which are closed implementations but implement open interfaces.

### Dependencies

The BoM includes all hardware dependencies and shows they have open interfaces. The main hardware subcomponent is the OSH R4 as previously published in JOH. The software stack builds on open source ROS2 and ROS2 standard nodes.

### BoM, Build instructions, User guide

The BoM is presented in the online repository [and included for review as Appendix A]. In addition to specifications and costs of components, it also includes links to suggested suppliers, and lists the openness of both their interface and the implementation. We suggest this is a useful convention for OSH BoMs, as differing definitions of OSH place different requirements on them. (Our preference is for a stronger definition of OSH than CERN-OHL licences, which requires interfaces to be open (i.e. non-patented and well-documented) and formally standardized (e.g. by published ISO standards) where possible, and preferring open implementations where possible. For example, the OpenPodcar2 donor vehicle is not an open implementation – it is a patented product – but it presents an open interface consisting of simple voltages send to the motors. )

Build instructions are presented in the online repository [and included for review as Appendix B]. As the design is based around a closed source donor vehicle we do not use a single CAD design file but rather show and explain steps of the build with occasional CAD images used to illustrate specific steps.

The User Guide is presented in the online repository [and included for review as Appendix C]. It assumes a working build and provides instructions for manual and autonomous drive modes in both simulation and real vehicle operation. A parameter tuning process is provided to compensate for small differences in builds.

## Discussion

OpenPodcar2 is a low-cost, robust, ROS2, open-source hardware and software-based autonomous mobility scooter, designed to facilitate the experimentation and testing of autonomous driving applications and local delivery applications. It sucessfully performs SLAM and navigation around indoor and outdoor campus environments, and pedestrian detection and tracking.

The Mounting Board is designed and separately CERN-OHL-W licenced to enable reuse in control of other similar Ackermann vehicles as well as the current donor vehicle. Future work could thus replace the donor vehicle with a deeper OSH mechanical implementation. A simple way to begin this process would be to base designs on the same main drive motor used in the current donor vehicle.

At the time of OpenPodcar2 `nav2` design, an MPPI controller server was in early release stages, using a similar foundation as its ROS1 predecessor `teb_planner` and supporting Ackermann models. When this matures it could be swapped into OpenPodcar2 and might provide more accurate control.

The pedestrian tracking system could be expanded by integrating a re-identification module for more accurate tracking and prediction.

To improve the system's localization accuracy, additional sensors can be integrated to enhance state estimation capabilities. Future work may also involve migrating the vehicle's software integration to an autonomous-driving-centric platform, such as Autoware. This transition would leverage Autoware's advanced tools and frameworks, which are designed specifically for autonomous vehicle applications, enabling improved path planning, perception, and control functionalities.

Future work could add high level ROS2 nodes to monitor battery levels, and gracefully shut down the vehicle or return it to a charging station in response. It could also connect more of the donor vehicle's features to R4 such as lights, horn, and windscreen wipers for automated control via the provided additional relay bank channels.

Users are encouraged and welcomed to contribute improvements to the design and to fork it for new applications.

## Declarations

Paper author contributions (CRediT): **Rakshit Soni**: Investigation (Software); Validation, Writing; **Chris Waltham**: Investigation (Mechanics, Electronics); **Md Umar Ibrahim**: Validation, Writing; **Charles Fox**: Methodology, Supervision, Writing.

Ethics: Human detection and tracking was conducted in accordance with University of Lincoln Research Ethics and Health and Safety Policies.

Funding statement: This research did not receive any specific grant from funding agencies in the public, commercial, or not-for-profit sectors.

Competing interests: The authors declare that they have no competing interests.

# Appendix A

# OpenPodcar2 Bill of Materials

| Name | Component | USD | Source | Interface | Implementation |
|---|---|---|---|---|---|
| Donor Vehicle | Phiseng TE-889XLSN mobility scooter (Branded as Shoprider Traverso) | 6000 new or 1000+ used | romamedical.co.uk/shoprider-traveso/ | generic (motor and brake control voltages; mechanical steering linkage) | patented |
| R4 | R4 OSH PCB robot control board | 300 | https://doi.org/10.5206/joh.v9i1.22878 | CERN-OHL-W | CERN-OHL-W |
| DepthCam | Intel RealSense D435 | 300 | store.intelrealsense.com/buy-intel-realsense-depth-camera-d435.html | ROS2 standard | closed |
| OSMC | OSMC motor driver v3-2 with 24V fan kit | 230 | www.robotpower.com/catalog/ | generic | public domain |
| Linear Actuator | Gimson Robotics GLA750-P 12V DC (100mm stroke, 240mm install) | 100 | gimsonrobotics.co.uk/products/gla-q40-12v-250n-compact-fast-travel-linear-actuator-with-encoder?variant=47116053905684 | generic | generic |
| Laptop Stand | Pyle | 60 | uk.redbrain.shop/p/00132017804903 | generic | generic |
| Power Relay | Ripca 12V, 200A REL-1/H-DUTY/200A/12V | 40 | parts.easycabin.co.uk/products/relay-200a-12volt-heavy-duty | generic | generic |
| DC/DC 24→12 | 20A | 40 | www.suremarineservice.com/Heat/Converters/DC2412-20C_2.html | generic | generic |
| Fuses (5A,12A,2A,10A,20A,100Ax2) | blade fuses and holders. | 40 | uk.rs-online.com/web/p/fuse-kits/2199556 | DIN 72581 standard | generic |
| Battery Charger | Nexpeak, 10A 12V/24V car battery charger | 25 | www.amazon.co.uk/Automatic-Temperature-Compensation-Motorcycle-Batteries-Red/dp/B094VQ88X2 | generic | closed |
| DBH12 | Dual h-bridge 12V motor driver | 25 | www.amazon.co.uk/Akozon-DC5-12V-0A-30A-Dual-channel-Arduino/dp/B07H2MDXMN | generic | closed |
| DC/DC 24→19 | 5A | 20 | www.amazon.co.uk/Converter-Waterproof-Regulator-Printers-Surveillance/dp/B087WWTSC4 | generic | generic |
| Wifi Router | Any, 12V powered | 20 | ... | IEEE Wifi standard | closed |
| R4 DMH Relay | SRD-05VDC-SL-C | 13 | www.amazon.co.uk/HUAREW-1-channle-optocoupler-isolation-triggering/dp/B0B52RPY43/ | generic | generic |
| 8-way Relay Bank | 8-Channel 5V Relay Module | 13 | https://www.sainsmart.com/products/8-channel-5v-relay-module | generic | generic |
| Acrylic Board, x3 | 400x200x5mm | 10 | www.amazon.co.uk/Perspex-Black-Acrylic-Plastic-Choose/dp/B09PRDRGXS/ | generic | generic |
| Circuit Breaker | DZ47-63 C10 | 10 | www.amazon.co.uk/RKURCK-DZ47-63-Low-voltage-Miniature-Circuit/dp/B0C3CQVWTS | ISO standard | generic |
| DMH | Philmore 30-825 SPST Hand Held Push Button Switch | 10 | https://www.amazon.com/Hand-Held-Button-Switch-30-825/dp/B00T6RCGNC | generic | generic |
| Alu Brackets | 3mm thickness alu plate; 2 pieces cut to 137x17mm and 152x25mm | 5 | https://www.amazon.co.uk/Aluminium-Sheet-plate-0-5mm-aluminium/dp/B0FHMDML7S/ | generic | generic |
| USB 3.0 | USB cable | 7 | https://www.amazon.co.uk/AAA-Products-USB-3-0-Cable/dp/B07778DHSM | generic | generic |
| DC/DC 24→5 | 10A | 5 | https://amzn.eu/d/4D8Er8h | generic | generic |
| Fan | Cooling fan for OSMC motor driver | 4 | https://www.ebay.co.uk/p/26035083262 | generic | public domain |
| Connectors | XT60 M and F, IDC headers, ribbon cables, Wago connectors, velcro, heatshrink, ring terminals, 10mm DIN rail, zip ties, JST (Three pin), blade fuse holders (2), thick wires (4mm$^2$, in red, black, yellow) | ... | ... | generic | generic |
| Nuts, bolts, washers, standoffs | M2-M6 metal and plastic | ... | ... | ISO standard | generic |
| Tools | hand toolkit, soldering, power drill, axle stands. | ... | ... | generic | generic |

## Appendix B

# OpenPodcar2 Build instructions

## Mechanical

### Donor Vehicle

Remove the doors from the Donor Vehicle, using a hex key to remove the bolts on the hinges. Remove the seat, following Donor Vehicle instruction manual. Remove the steering column, following the Donor Vehicle service manual instructions. Remove *all* original wiring, except for the batteries, motor and their connectors. Remove the original motor controller-driver as shown in Figure 8. It is located in a compartment under the seat on the driver's left (model versions may vary by Donor Vehicle production year).

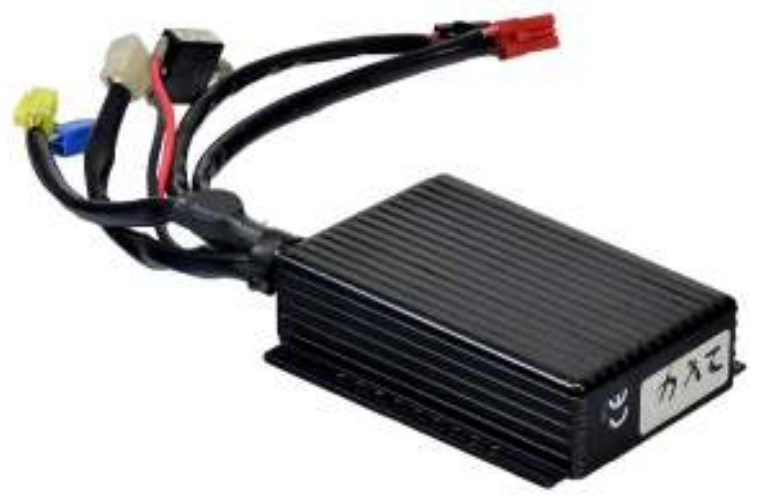

Figure 8: Original motor controller-driver removed from the Donor Vehicle

Install the 100A blade fuse in a blade fuse holder to the +24V battery output. This can be done using Wago connectors, or by soldering and heat-shrinks. Keep the wire from the battery to the fuse holder as short as possible. Cover the +24V terminal of the battery, e.g. with duct tape, to prevent accidental contact.

**All further connections to the battery +24V are made via this fuse. Do not, ever, connect anything to battery +24V before this fuse.**

Cut off the lower part of the Laptop Stand using a hacksaw. Insert the upper part of the Laptop Stand to the (non-rotating) base mount where the removed steering column was previously. Push it down firmly by hand and tap with a mallet to ensure a tight fit.

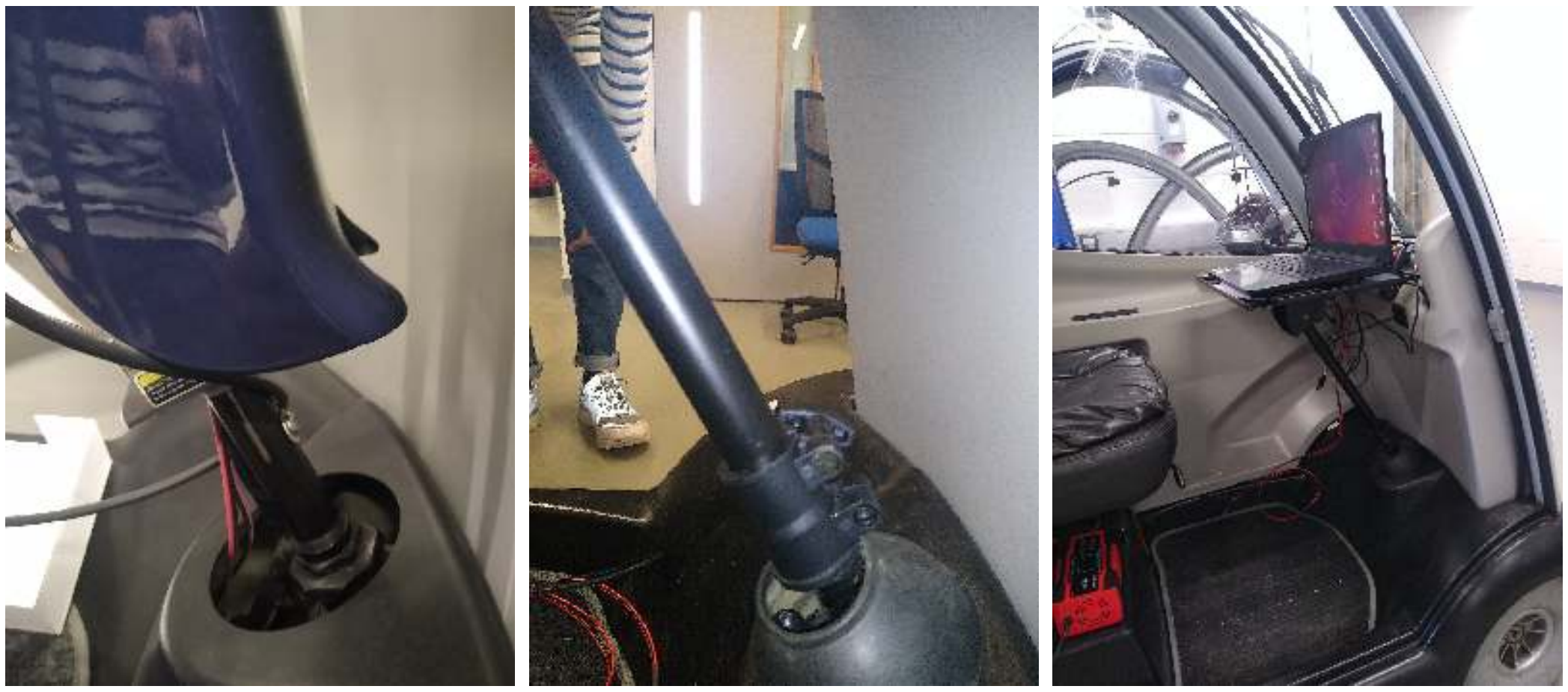

Figure 9: Original steering column; replaced with laptop stand

### Steering mechanism

Cut two brackets from 3mm aluminium plate as in Figure 12.

To access the underside of the Donor Vehicle, use two people to tilt the vehicle and place it on axle stands or similar devices to secure it in place as in Figure 10.

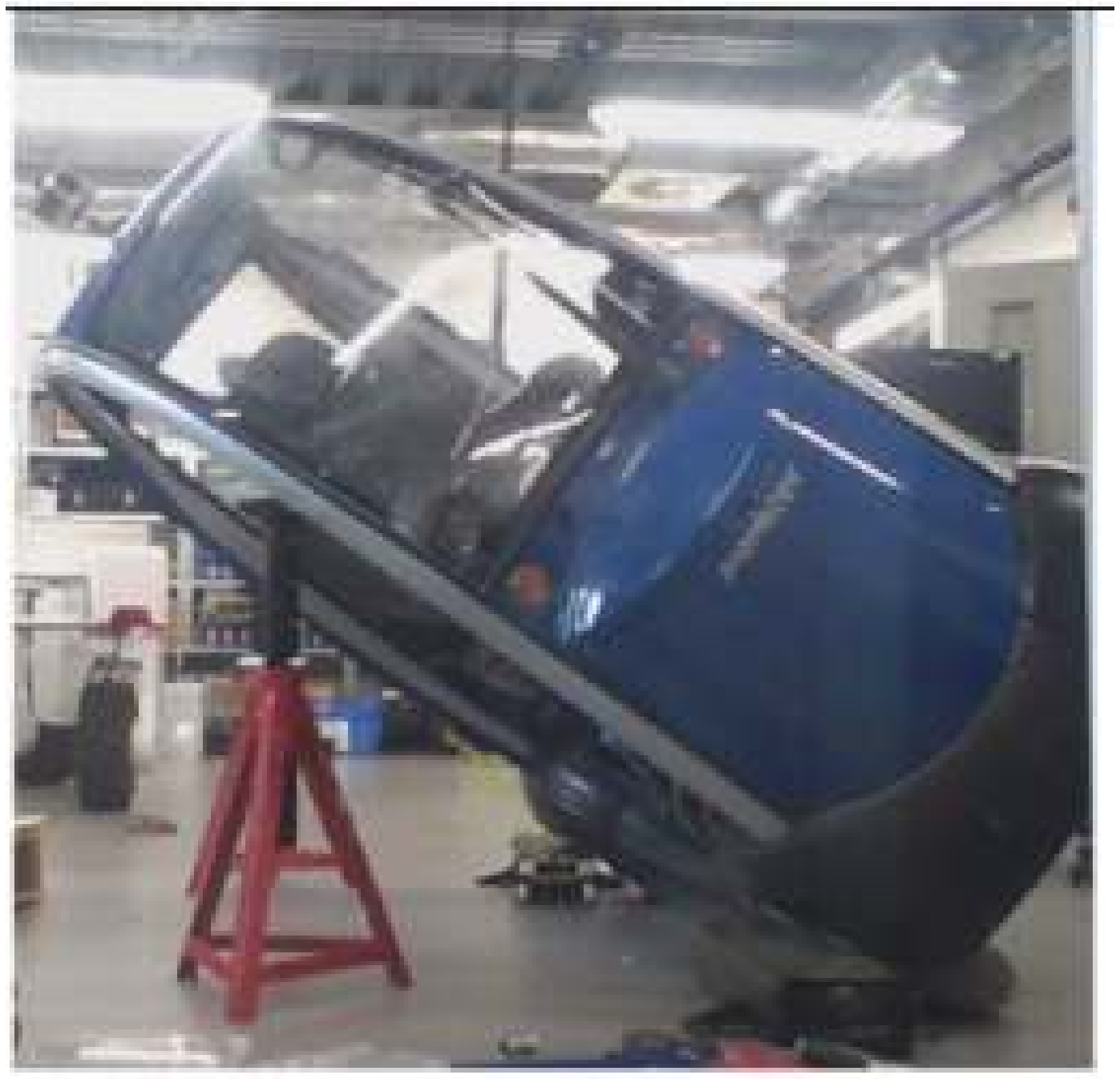

Figure 10: Support on axle stands

The Donor Vehicle has Ackermann steered geometry. To control this steering, a the Linear Actuator is mounted on the brackets near to the tie rod in the front. Due to mechanical limits, the Linear Actuator is not exactly in the middle of the front axle. Figure 11 shows a geometric representation of the steering with the Linear Actuator.

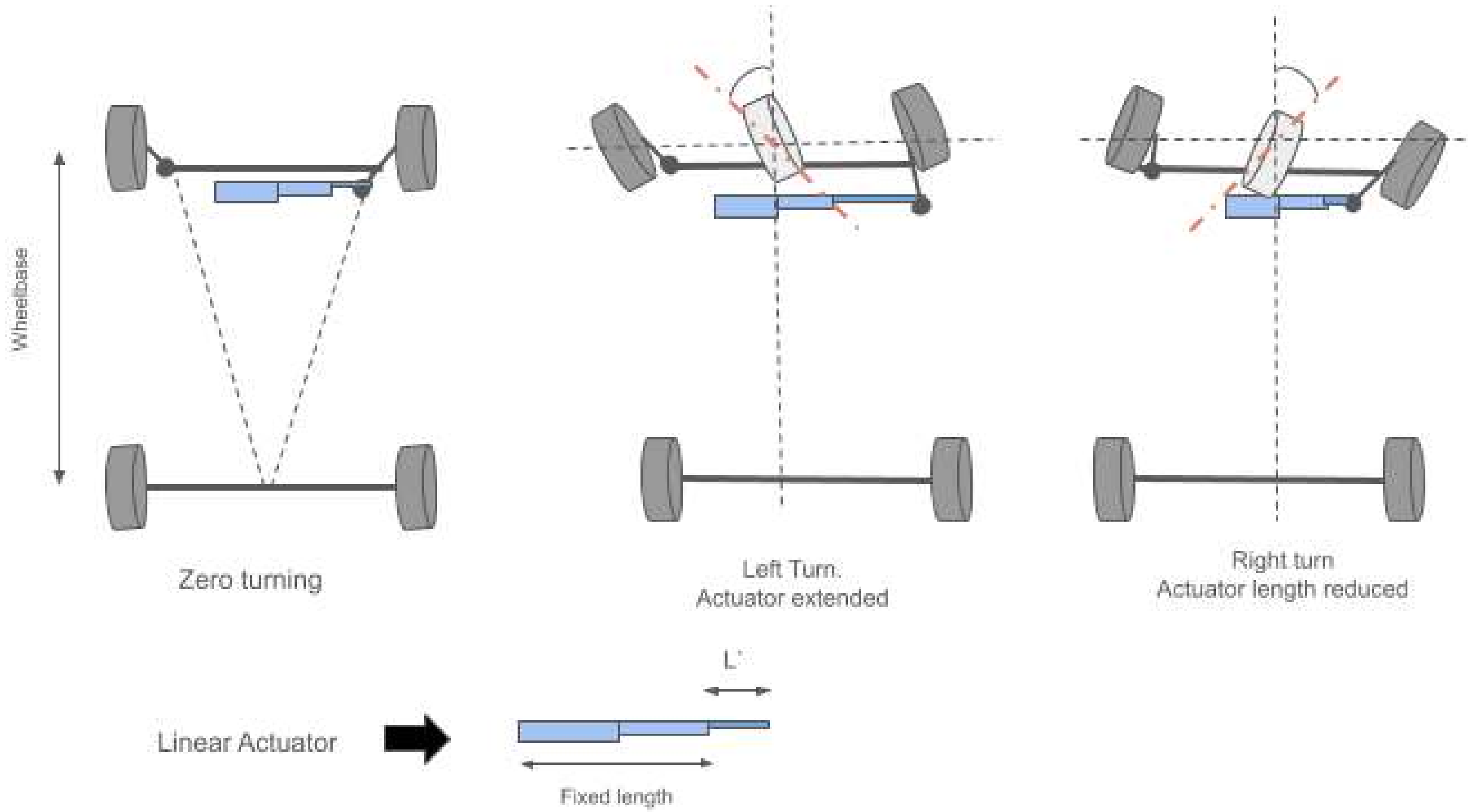

Figure 11: Underside view of front wheels' steering relationship

Drill holes into the steel structure of the chassis to attach the brackets. *Drilling steel can be dangerous so requires a trained operative.* Use long bolts with multiple nuts and washers as spacers to level the vertical positions.

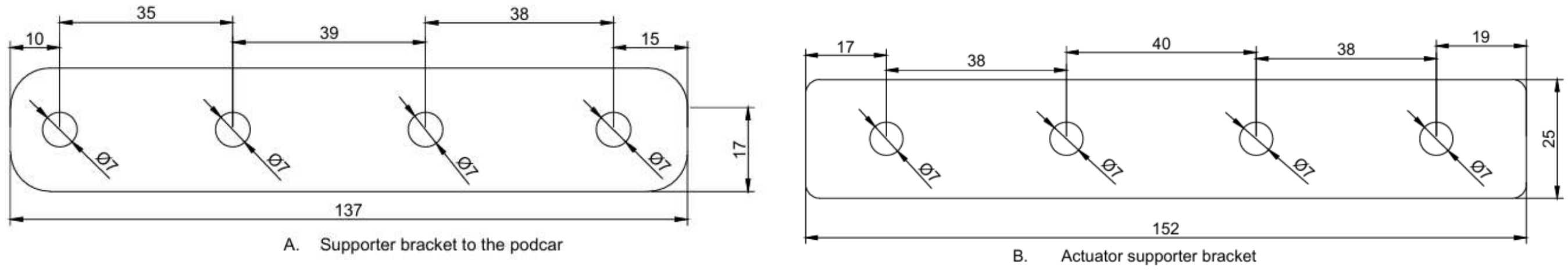

Figure 12: Steering bracket CAD

Install the Linear Actuator on the underside of the vehicle, using the brackets with nuts, bolts and washers for spacing to position its base, as shown in Figure 13. Connect the shaft end of the Linear Actuator to the existing hole present (from the original Donor Vehicle design) in the tie rod of the Ackermann steering of the Donor Vehicle, using a bolt, nut, and washers for spacing.

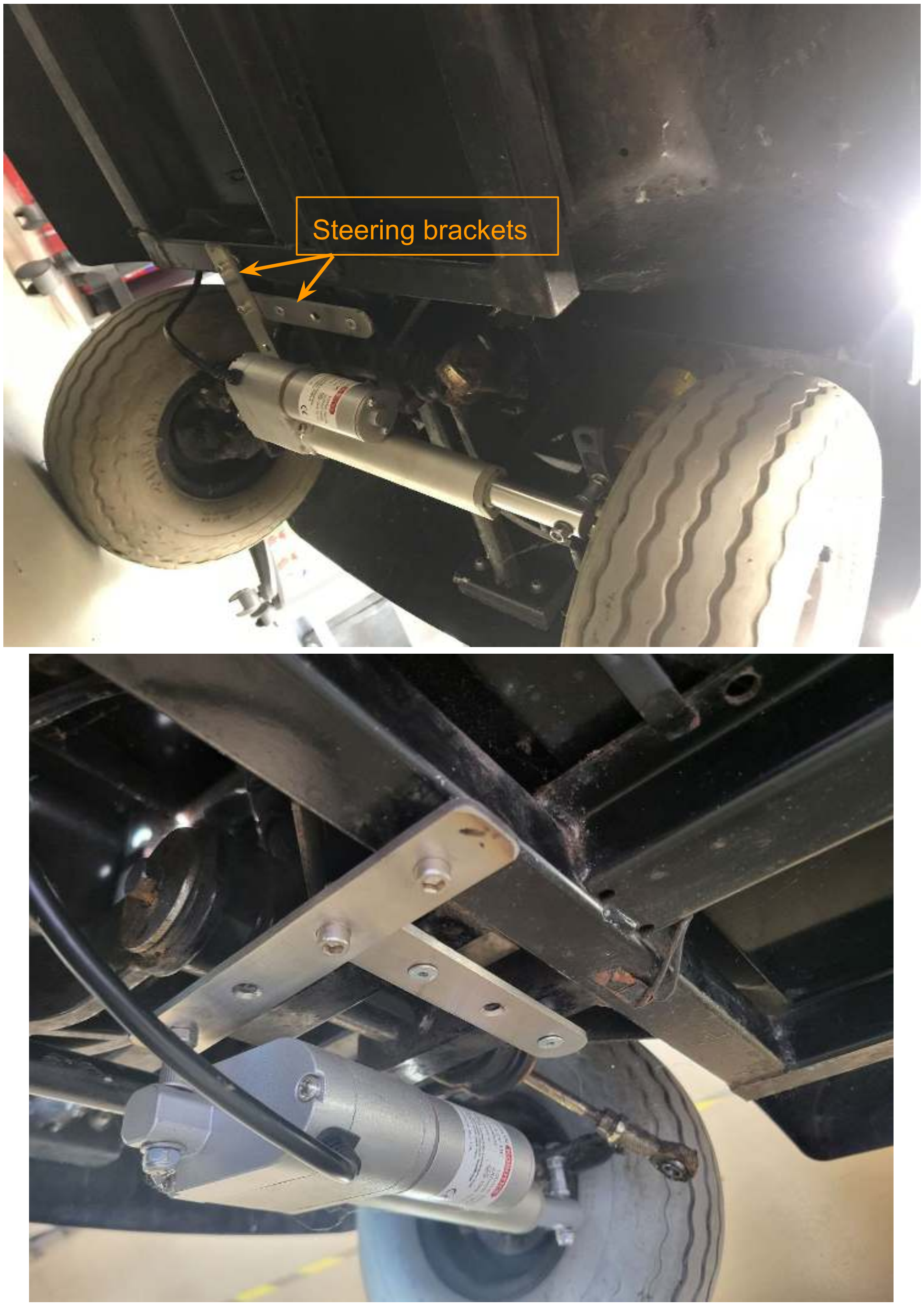

Figure 13: Steering linear actuator installation

Connect wires to the Linear Actuator and run them to the passenger compartment for later connection.

Use two people to un-tilt the vehicle and return it to its normal position.

### Depthcam physical install

Lift the Donor Vehicle's front panel (usually this is done to refill the windscreen washer fluid). Drill a hole in it and bolt on the Depthcam as in Figure 14.

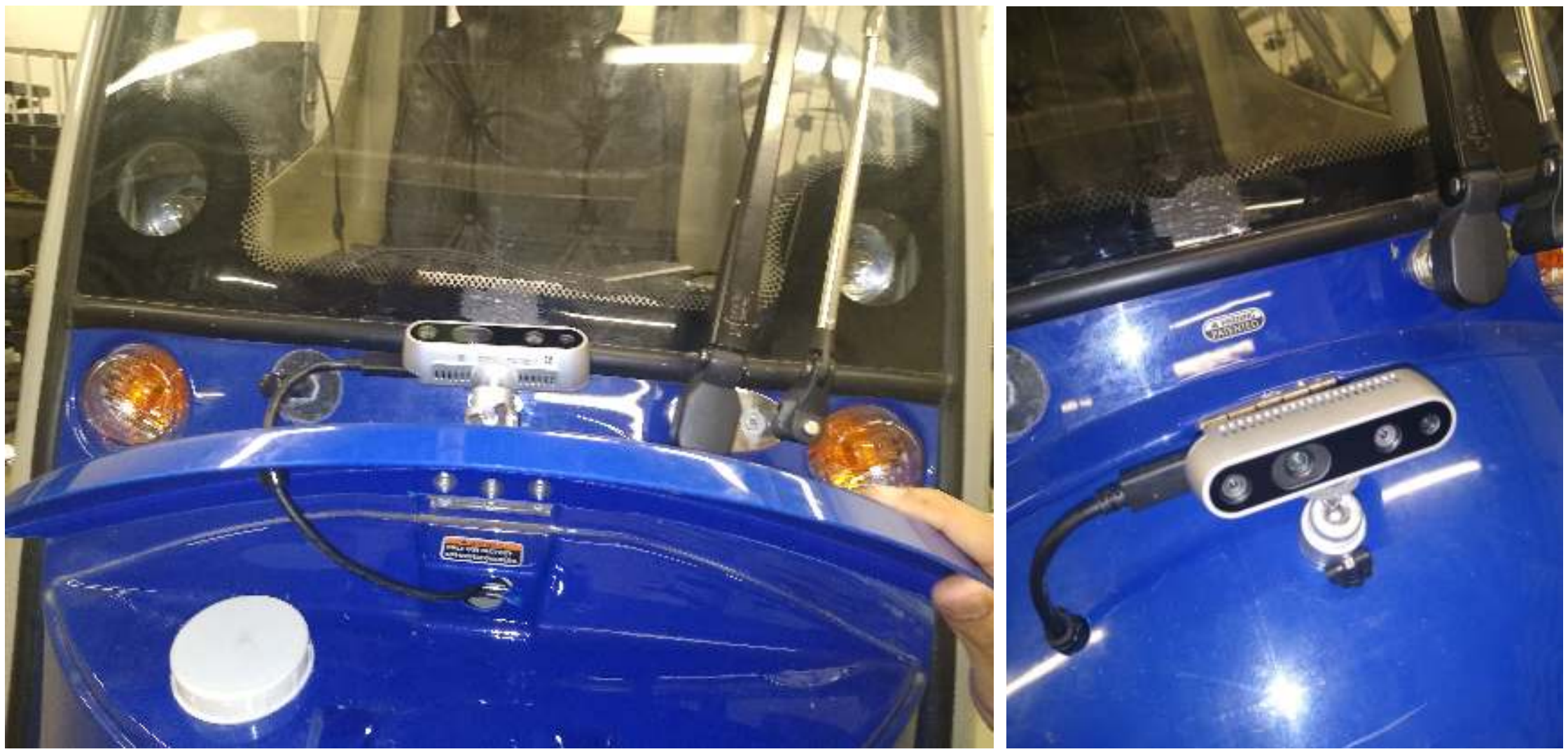

Figure 14: Depth camera installation

**Wifi Router install**

Drill holes in the upper rear of the Donor Vehicle, use them with cable ties to mount the Wifi Router to the vehicle exterior as shown in Figure 15.

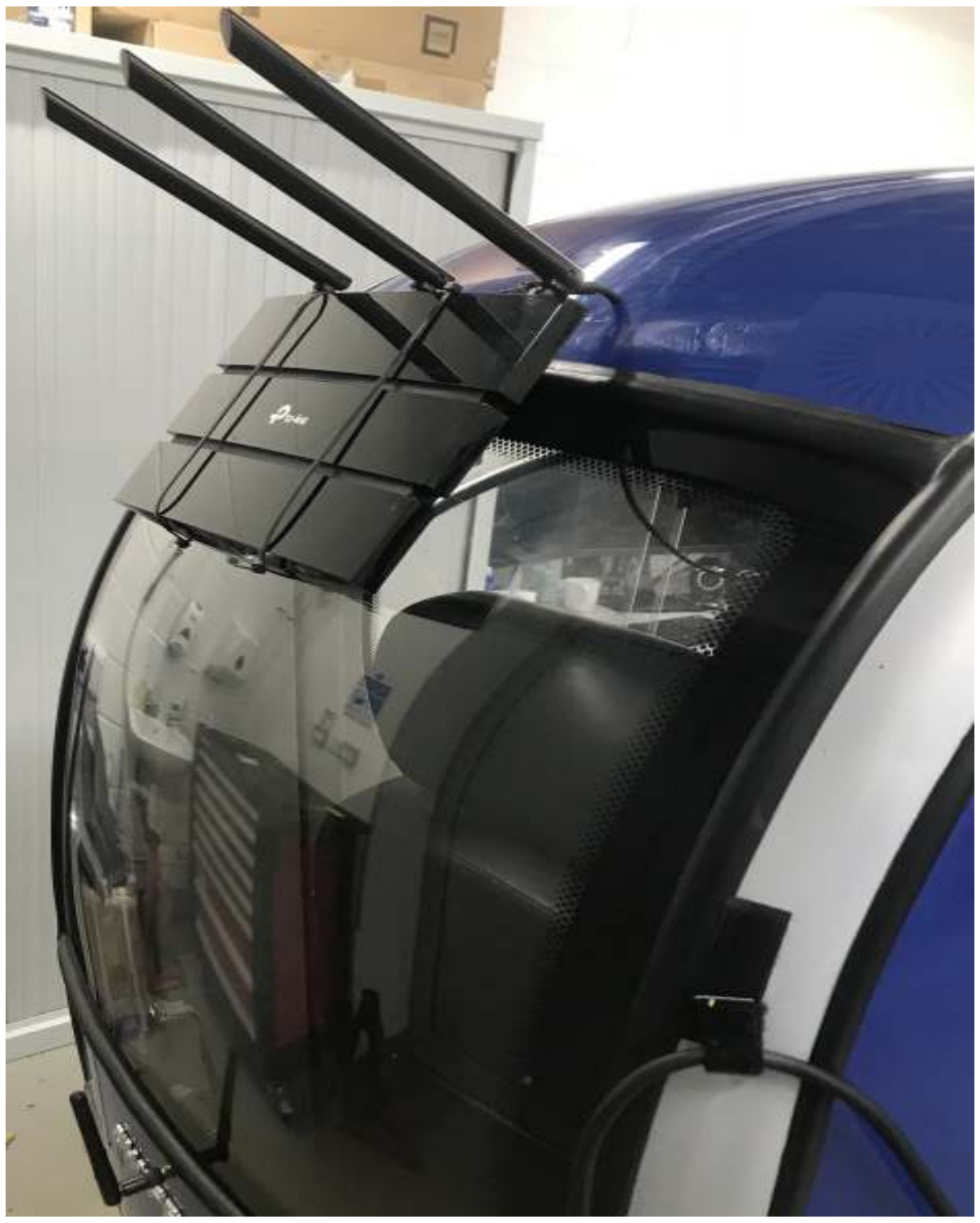

Figure 15: WiFi Router installation

**Mounting Board**

Cut three Acrylic Boards each to 400 mm × 215 mm (with 5 mm thickness). The boards may be cut from a single acrylic (e.g. Perspex brandname) sheet using a handsaw, laser cutter, or an external cutting service.

In each Acrylic Board, drill 5 mm diameter mounting holes around the perimeter, as in Figure 16.

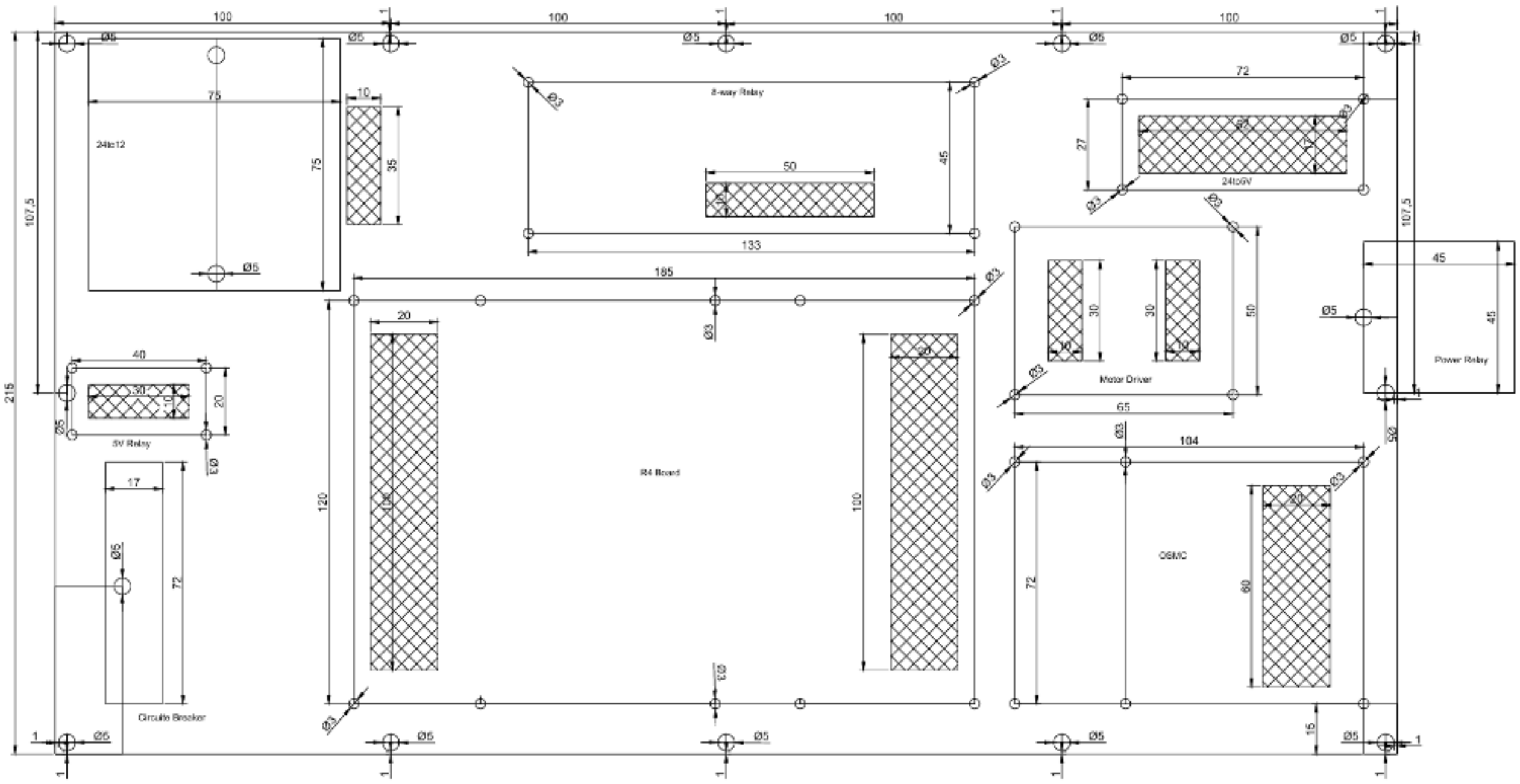

Figure 16: 2D design of mounting board

On one of the Acrylic Boards, arrange the required electronics as shown in Figure 17. Mark the positions of all component mounting holes on the board using a marker. After removing the components, drill the marked points to the required diameter.

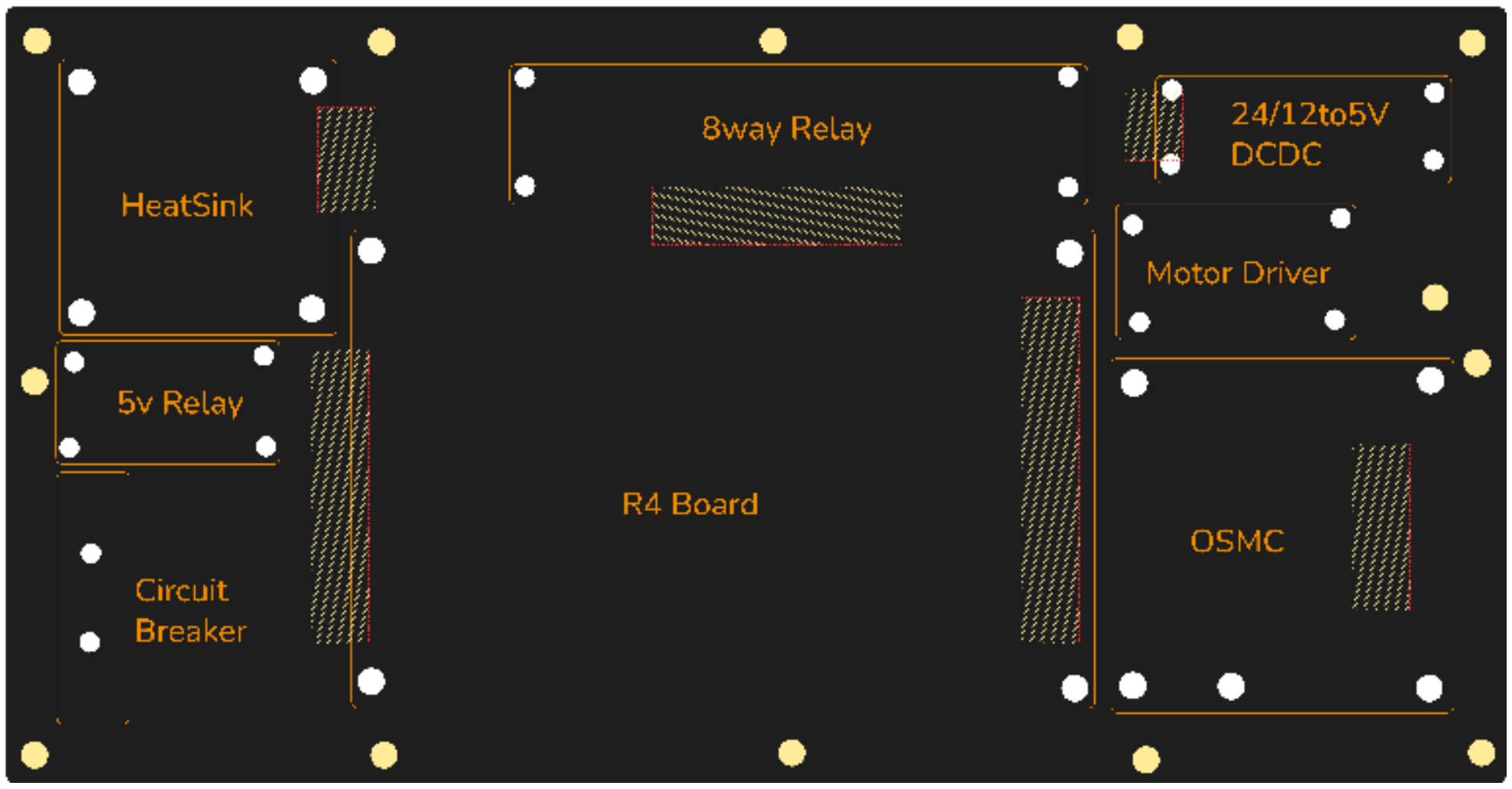

Figure 17: Components position on Mounting Board

To form the Mounting Board, stack the three Acrylic Boards vertically using metal bolts as spacers/standoffs at the perimeter holes as shown in Figure 18.

Connect four 500mm thick wires to a metal bolt used as one of the spacers, in the OSMC corner of the Mounting Board, using ring terminals, to form the GND distributor as in Fig. 19.

The space between the Acrylic Boards can be used to route wires and keep them tidy in the build.

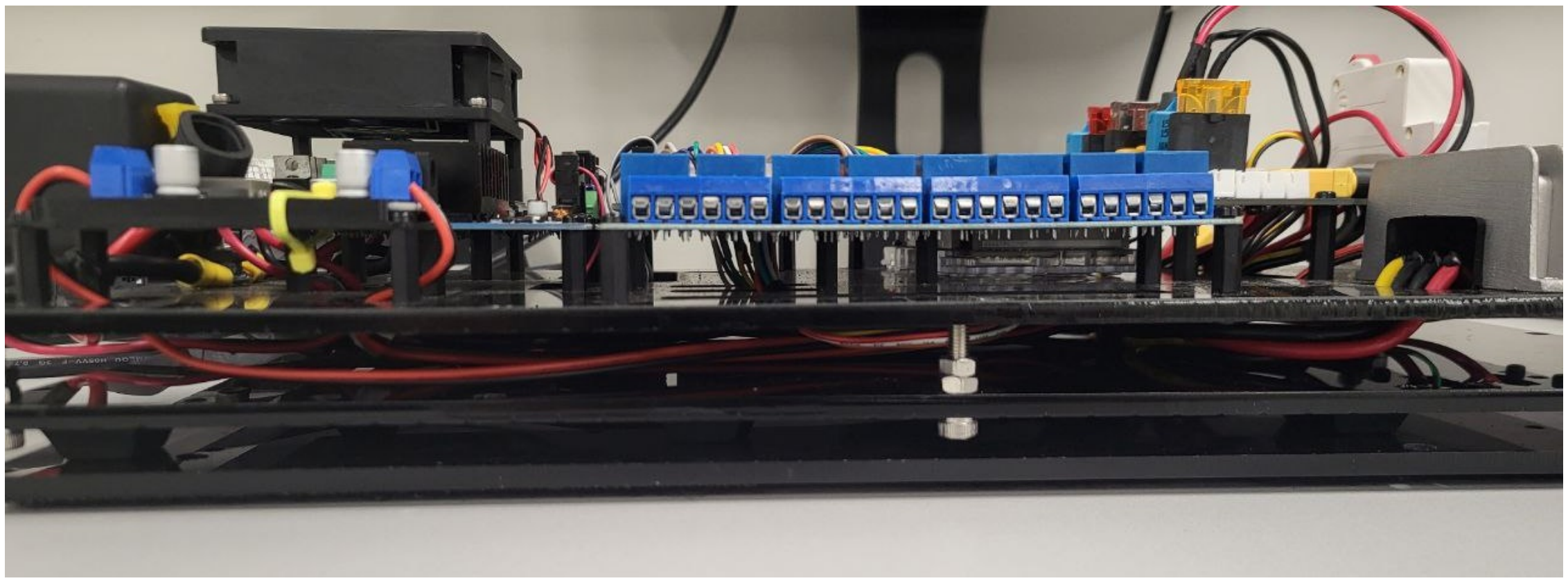

Figure 18: Mounting board, formed from three stacked Acrylic Boards

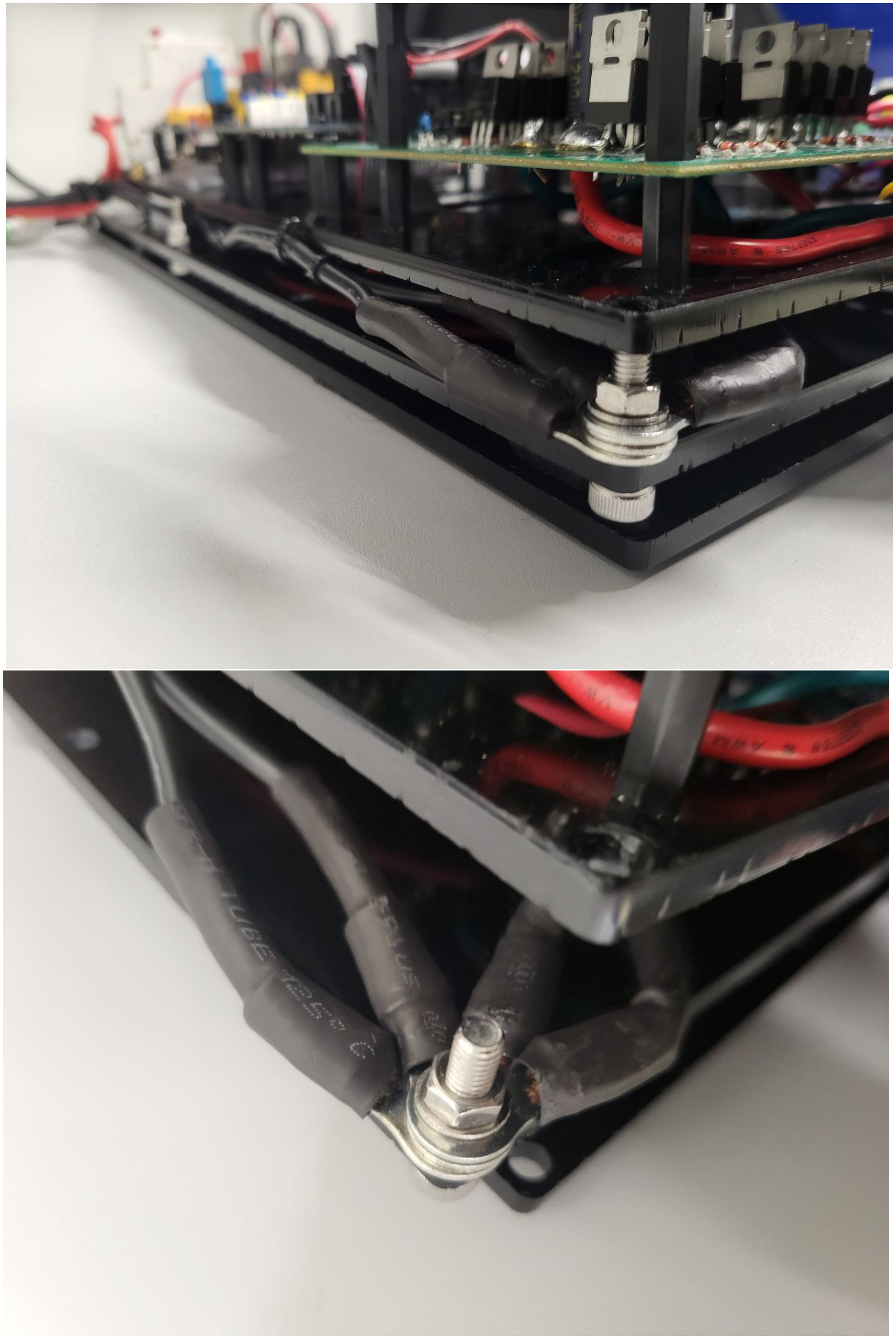

Figure 19: Forming the GND distributor from one corner's spacers.

## Electrical and Electronics

### Circuit breaker

Mount the Circuit Breaker on the Mounting Board using the 10mm DIN rail bolted to the Mounting Board.

Ensure that the Circuit Breaker is in the off position.

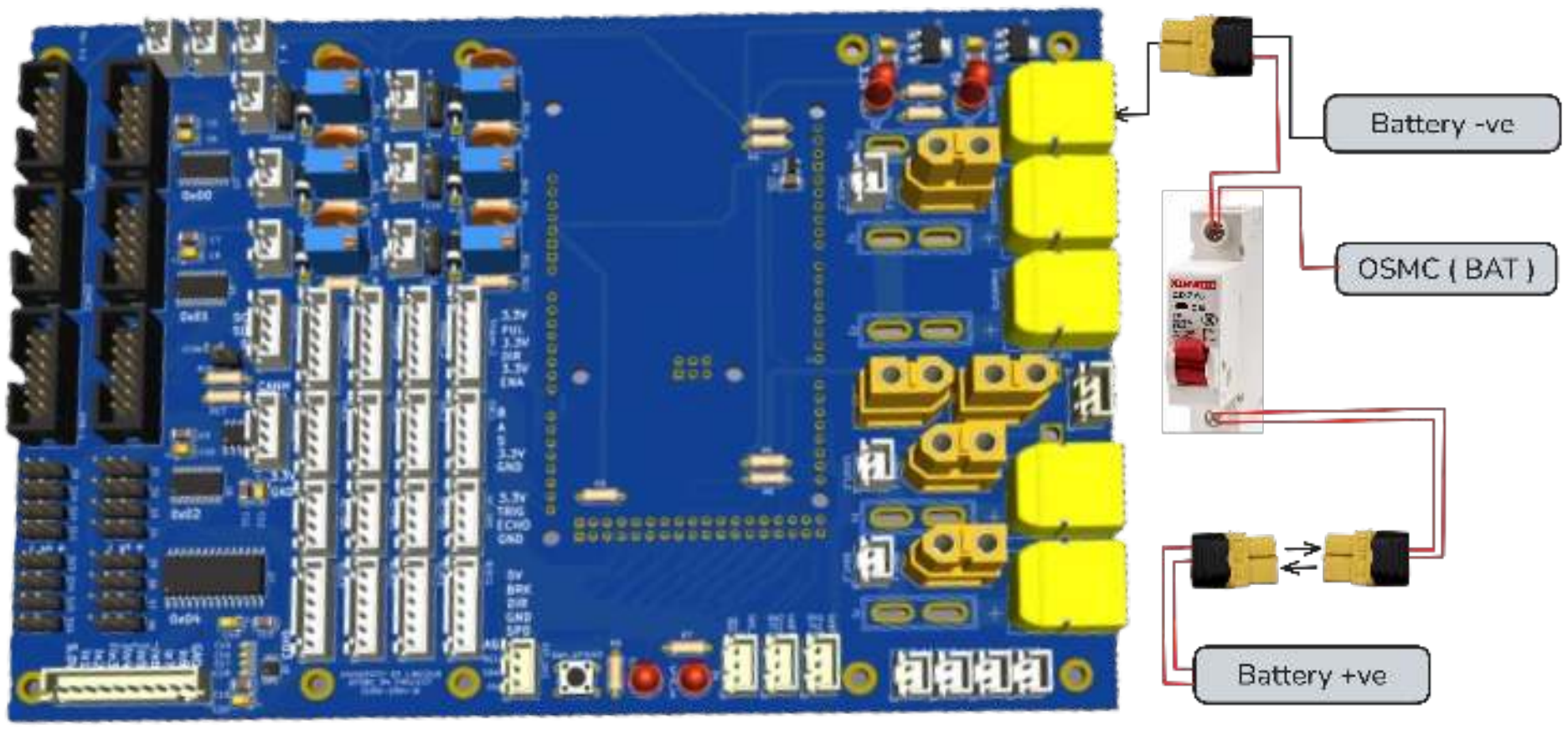

Figure 20: Circuit Breaker connections

### R4

Bench-test the R4 according to the R4 documentation. Configure the R4 firmware to use the network name and password as the router, using the R4 documentation.

Fit the 5-15A (inclusive) fuses in the fuse holders mounted on the R4 as shown in table 1 and Figure 21. (Further 20A and 100A fuses will be added elsewhere in OpenPodcar2, later.)

Table 1: Connection table for R4 fuses

| Fuse | Destination | Notes |
|---|---|---|
| Fuse 2A | R4: F3 | Sensor & auxiliary protection |
| Fuse 5A | R4: F5 | Protects low-current logic circuits |
| Fuse 10A | R4: F2 | Moderate load circuits |
| Fuse 15A | R4: F1 | Medium-current subsystems |
| Fuse 15A | R4: F4 | Medium-current subsystems |

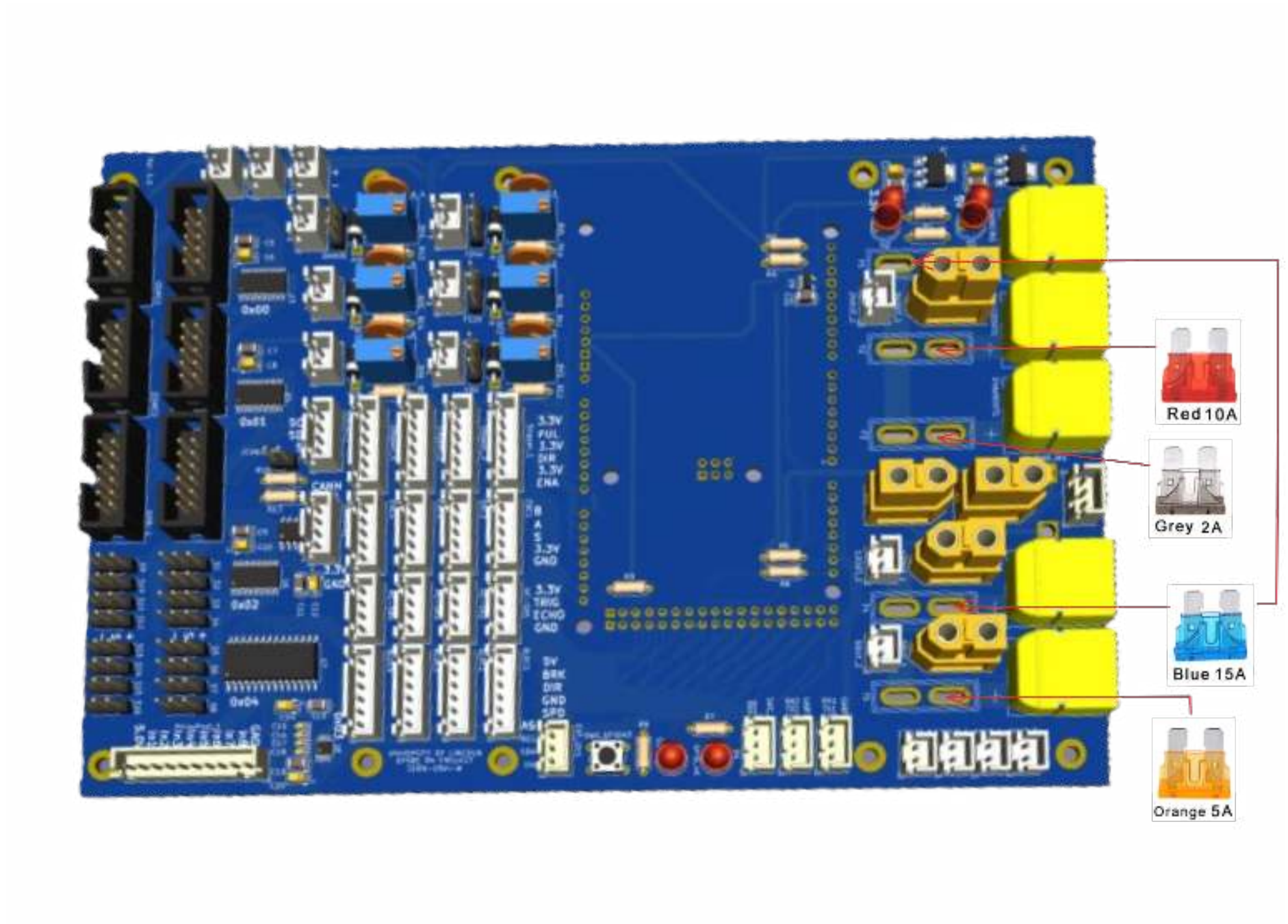

Figure 21: Required fuses for OpenPodcar2 mounted on R4

Mount the R4 on the mounting board, using standoffs as shown in Figure 18.

Connect the R4:24VDCIN1:1 to Circuit Breaker:1 and R4:24VDCIN1:2 to GND, using an XT60H connector and wires.

### Dead Man Handle (DMH)

Wire the DMH as a two-core input to the R4 safety input channel DMH_IN1.

Table 2: Connection table for dead man's handle (DMH)

| DMH Pin | Destination | Notes |
|---|---|---|
| DMH:IN | R4:DMH_IN1:1 | XT60H-M |
| DMH:OUT | R4:DMH_IN1:2 | XT60H-M |

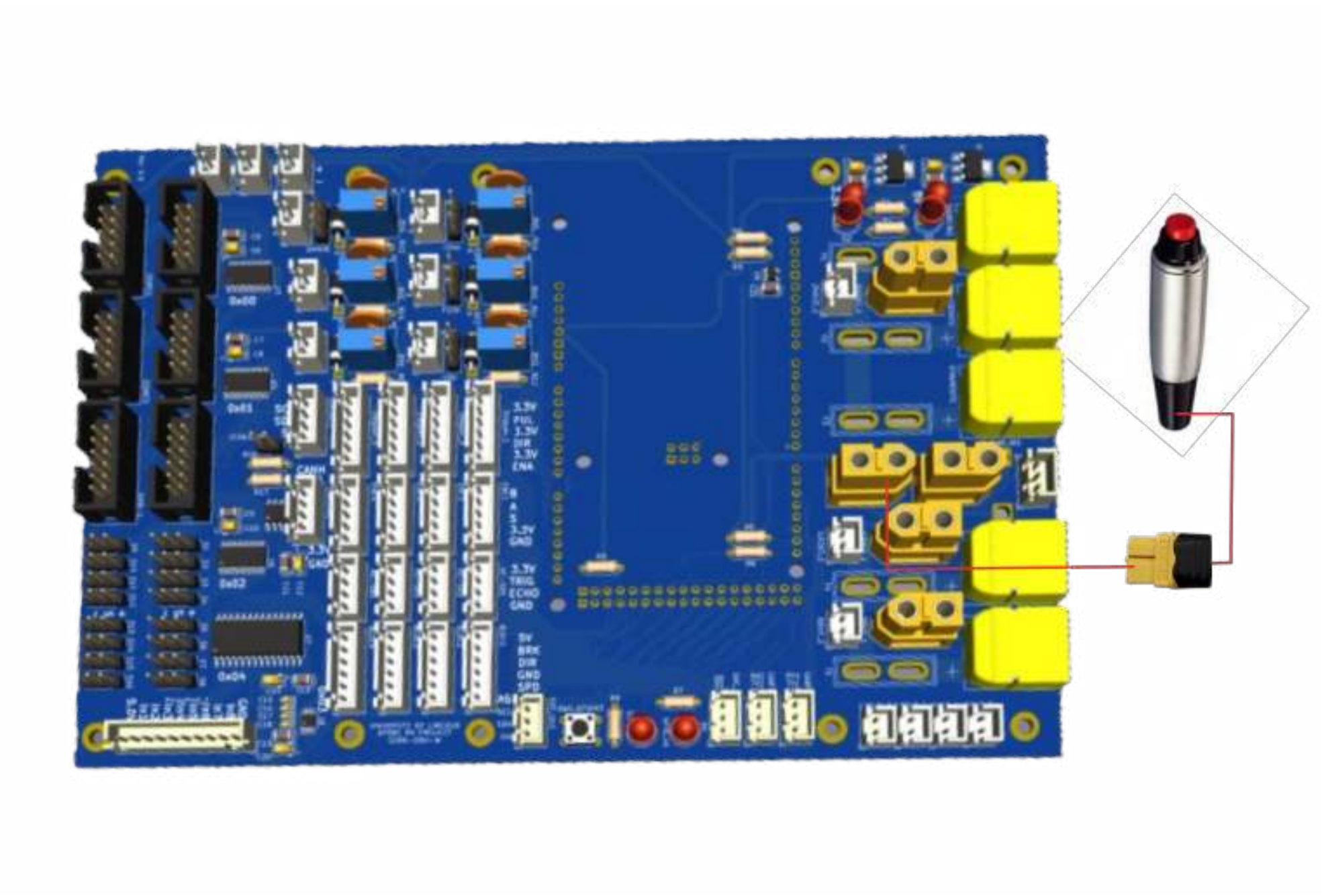

Figure 22: Dead man handle (DMH)

Mount the R4 DHM Relay on the mounting board using standoffs.

Connect the R4 DHM Relay to the R4 relay interface using a 3 pin JST harness carrying input, 5V supply and ground, as shown in table 3 and Figure 23.

Table 3: Connection table for R4 DHM Relay

| **R4 DHM Relay Pin** | **Destination** | **Notes** |
|---|---|---|
| NO | R4: J1_CAR_IN1 | Single red wire (amp) (norm. open) |
| COM | R4: J2_DMH_OUT1 | Single red wire (amp) |
| NC | No connection | – |
| DC- | R4: DMH_RLY1:1 | 3 Pin JST |
| DC+ | R4: DMH_RLY1:2 | 3 Pin JST |
| IN | R4: DMH_RLY1:3 | 3 Pin JST |

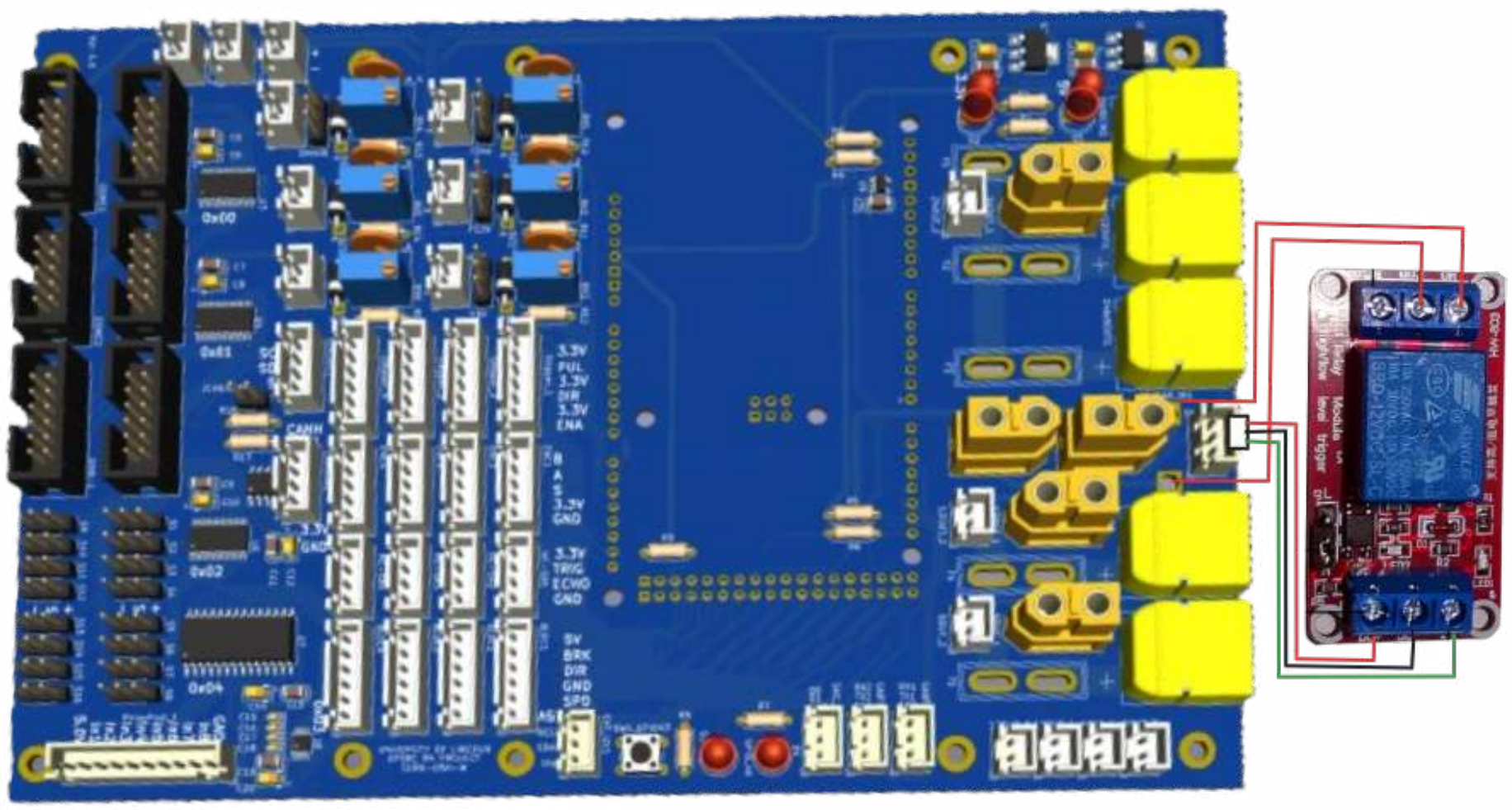

Figure 23: R4 DHM Relay

### DC/DC 24→5V converter

Mount the module on the Mounting Board using standoffs.

(Note: XT60H is a standard connector type. These yellow plastic connectors have a curved side hosting pin 1 for +ve and a square side hosting pin 2 for -ve; and that -M an -F denote male and female connectors. They can be attached to wires by soldering and heat shrink.)

Connect the converter to the R4 power distribution at 5IN1 using an XT60H-F two-way connector. Route the 5V output to R4:24Vto5OUT1. Connections are shown in Table 4 and Figure 24. Keep cable runs short.

Table 4: Connection table for DC/DC (24→5V)

| DC/DC 24→5V Pin | Destination | Notes |
|---|---|---|
| IN+ | R4: 24Vto5OUT1:1 | XT60H-M |
| IN- | R4: 24Vto5OUT1:2 | XT60H-M |
| OUT+ | R4: 5IN1:1 | XT60H-F |
| OUT- | R4: 5IN1:2 | XT60H-F |

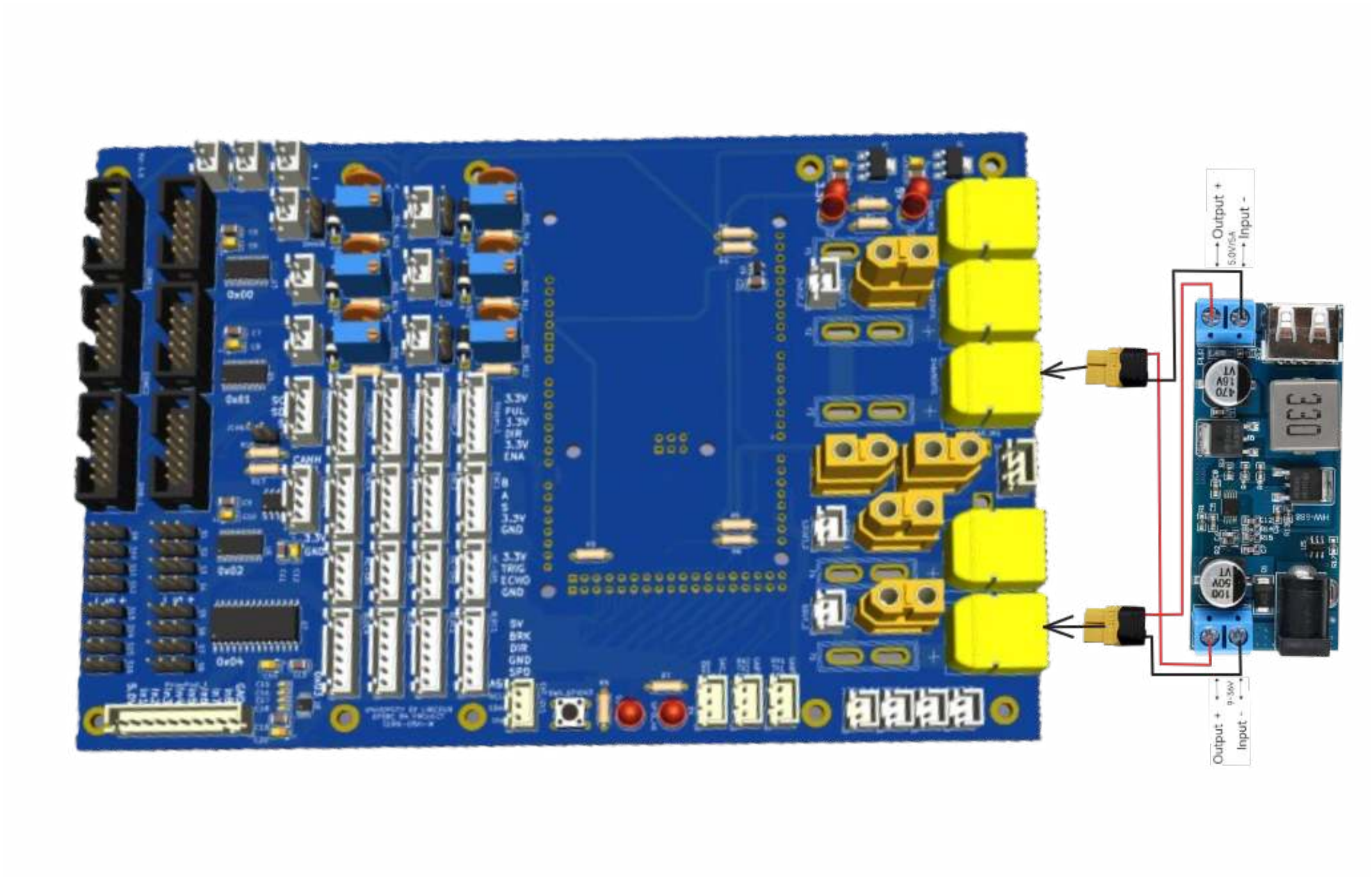

Figure 24: DC/DC 24→5V converter

### DC/DC 24→12V converter

Mount the converter on the Mounting Board using bolts.

Connect the converter both to and from R4 as in table 5 and Figure 25.

Table 5: Connection table for DC/DC 24→12V converter

| DCDC Pin | Destination | Notes |
|---|---|---|
| Heat Sink (Red) | R4 : 24to12out1:1 | XT60H-M connector |
| Heat Sink (Black) | R4 : 24to12out1:2 | XT60H-M connector |
| Heat Sink (Yellow) | R4 : 12IN1:1 | XT60H-M connector |
| Heat Sink (Black) | R4 : 12IN1:2 | XT60H-M connector |

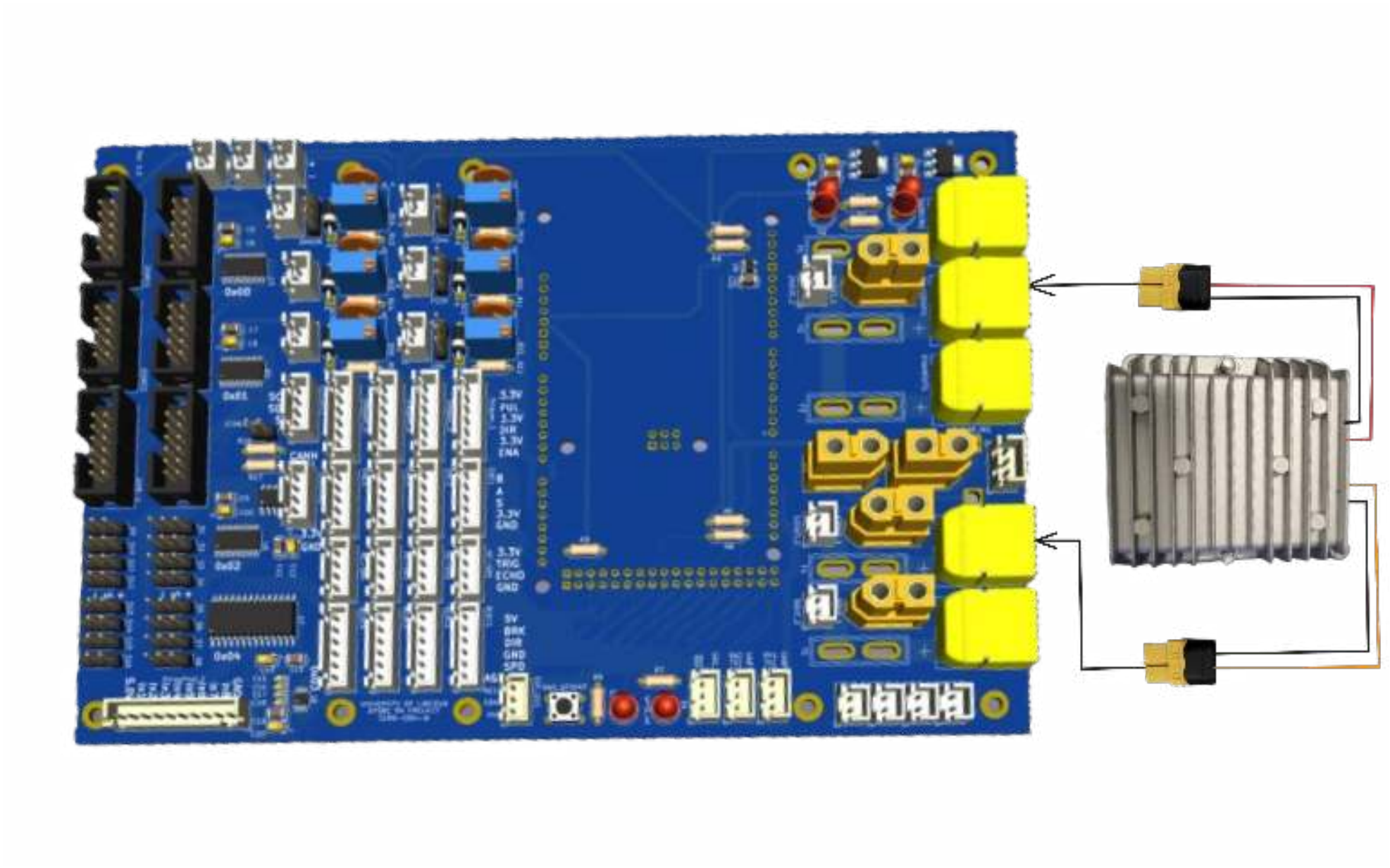

Figure 25: DC/DC (24→12V)

### DC/DC 24→19V converter

Mount the module on the mounting board using bolts.

Take input power from the R4 high current 24OUT_1 connector using the XT60H M red and black pair, and route the converter output to the target device using the yellow and black XT60H M pair, as shown in table 6 and Figure **??**. Keep cabling short and strain relieved.

Table 6: Connection Table for DC/DC 24→19V converter

| **Heat Sink Pin** | **Destination** | **Notes** |
|---|---|---|
| Red | R4: 24OUT_1:1 | XT60H-M |
| Black | R4: 24OUT_1:2 | XT60H-M |
| Yellow | Laptop: DC in:+19V | Laptop's DC connector type |
| Black | Laptop: DC in:GND | Laptop's DC connector type |

### DHB12 motor driver

(Note: DHB12 is a dual channel driver, which can drive motors A and B, but only A is used in OpenPodcar2).

Mount the DHB12 on the mounting board using standoffs.

Connect to the R4 DHB12 1 socket using a 12-pin IDC ribbon cable to the DHB12 logic header, with careful pin one alignment at the driver end.

Supply 12V power to the DHB12 from R4:12OUT1.

Connect the actuator to the DHB12 output terminals as a two wire pair using its channel A pair, using thick wire.

Connections are summarized in table **??** and Figure 27.

Table 7: Connection table for DHB12 steering motor driver

| DHB12 Pin | Destination | Notes |
|---|---|---|
| Logic header (2x6) | R4: DHB12-1 | IDC12 pin |
| A+ | Linear Actuator: +ve | use thick wire |
| A- | Linear Actuator: -ve | use thick wire |
| PWR+ | R4: 12OUT1 | XT60H |
| PWR- | R4: 12OUT1 | XT60H |

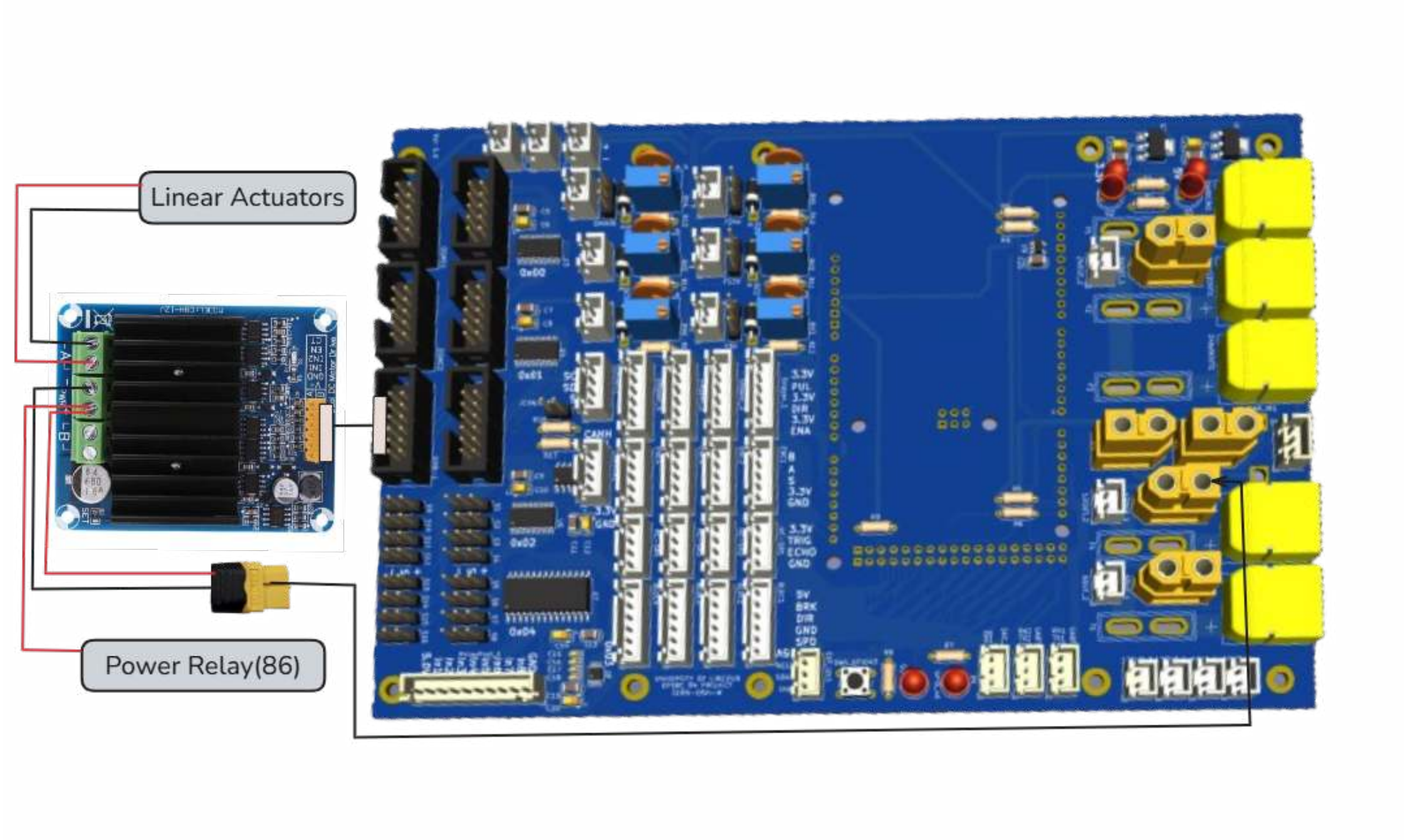

Figure 26: DHB12 steering motor driver connection with R4

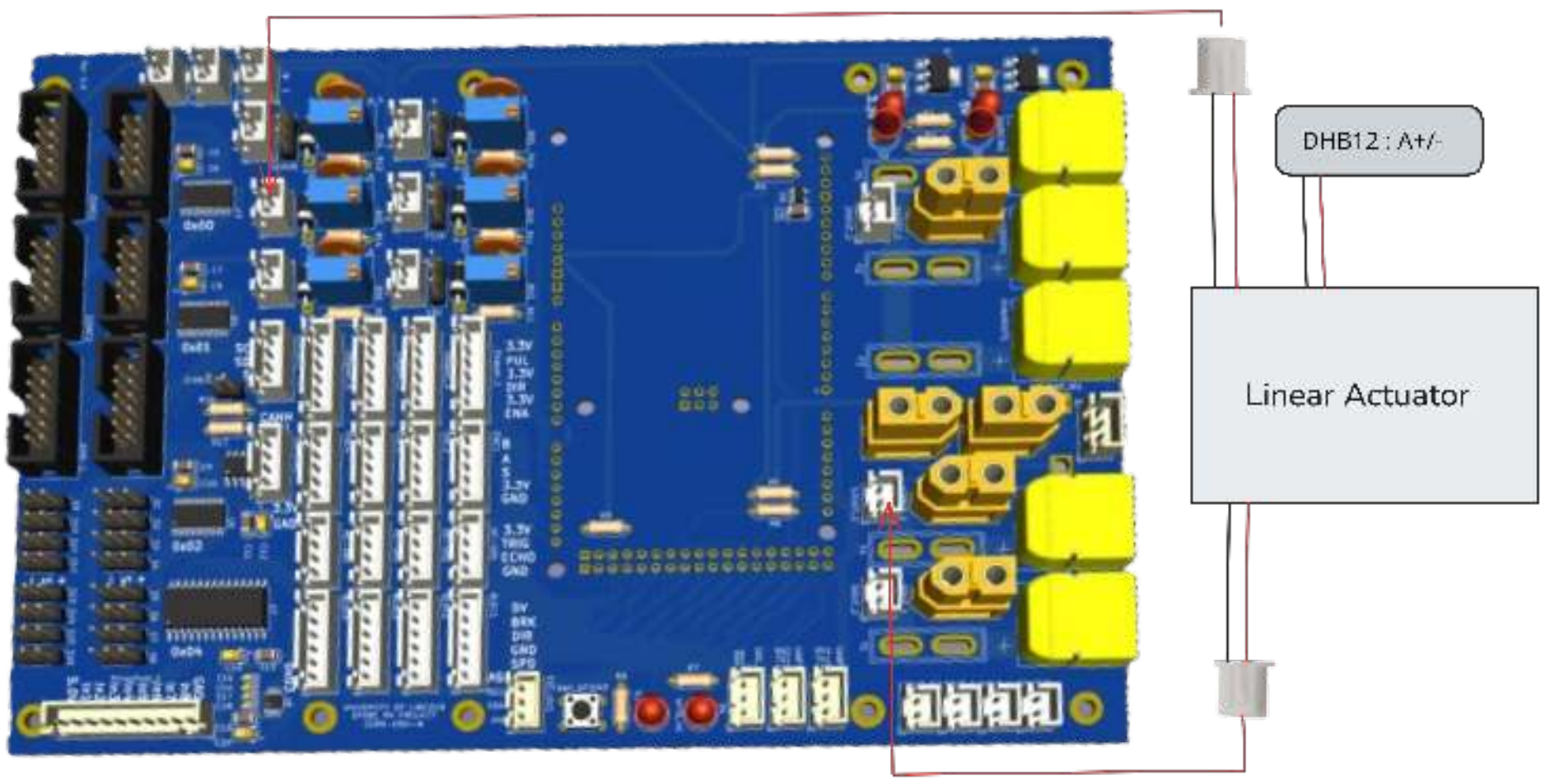

Figure 27: Linear Actuator connection with R4

### Linear Actuator connection

Connect the Linear Actuator's motor pair (+ve, -ve) wires to the DHB12 output terminals on channel A, using the thicker conductors for high current drive.

Power the Linear Actoator's 12V encoder from the R4:12OUT2 pair. Connect the encoder signal to R4:AIN5.

Table 8: Connection Table for Linear Actuator

| **Linear Actuator Pin** | **Destination** | **Notes** |
|---|---|---|
| +ve | DHB12: A+ | thick wire |
| -ve | DHB12: A- | thick wire |
| Encoder Power | R4: 12OUT_2:1 | 2-pin JST connector |
| Encoder GND | R4: 12OUT_2:2 | 2-pin JST connector |
| Encoder Output Channel A | R4: AIN5:1 | 2-pin JST connector |
| Encoder Output Channel B | R4: AIN5:2 | 2-pin JST connector |

### OSMC motor driver

(Note: The wires from the Motor were previously disconnected from the original Donor Vehicle's controller and driver when they were removed, so are available in the compartment under the seat and do not require further access via the underside of the vehicle. You can use the original connector pair from the Donor Vehicle to make connections to them.)

Mount the OSMC on the mounting board using standoffs.

Mount the Power Relay on the mounting board using bolts.

To power the OSMC: Connect OSMC:BAT to Circuit Breaker:1 and OSMC:GND to GND (i.e. the Mounting Board ring terminal distributor).

Connect control signals from the R4:OSMC1 interface to the OSMC:CN5 10 pin header, using a keyed ribbon data cable, and maintaining correct pin one orientation.

To send power to the motor: Connect OSMC:MOT- to Motor:-ve. Connect OSMC:MOT+ to Power Relay:30, and connect Power Relay:87 to Motor+ve via a 20A Fuse mounted in a blade fuse holder.

To control the Power Relay: connect Power Relay: 86 to DHB12:PWR+ (this is just a convenient place to tap the 12V power supply, it not a control signal); and connect Power Relay:85 to R4:CAR_IN1 (which does the switching on DMH and eStop).

These connections are shown in tables 9 and 10 and Figures 28 and 29.

Table 9: Connection table for OSMC

| **OSMC Pin** | **Destination** | **Notes** |
|---|---|---|
| OSMC:CN5 | R4: OSMC1 | Ribbon wire for control signals |
| OSMC:BAT | Circuit Breaker: 1 | Soldered underneath of board |
| OSMC:GND | GND | Ring terminal on Mounting Board |
| OSMC:MOT+ | Power Relay:30 | |
| OSMC:MOT- | Motor:-ve | |

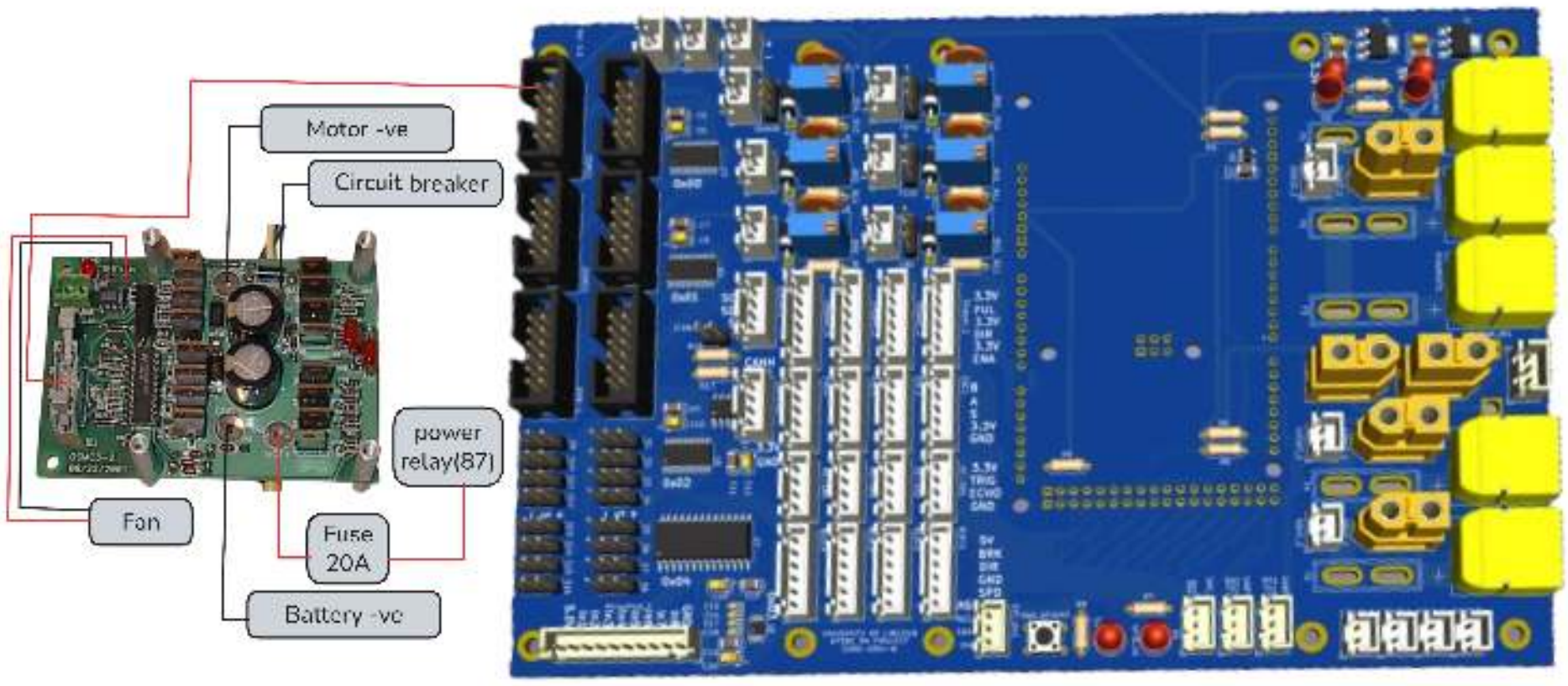

Figure 28: OSMC connections

Table 10: Connection table for Power Relay

| Power Relay Pin | Destination | Notes |
|---|---|---|
| 86 | DHB12:PWR+ | +12V supply (not control) to signal |
| 85 | R4: CAR_IN1 | signal ground switched on R4+DMH |
| 30 | OSMC: MOT+ | thick wire |
| 87 | Motor:+ve:20A fuse | thick wire |

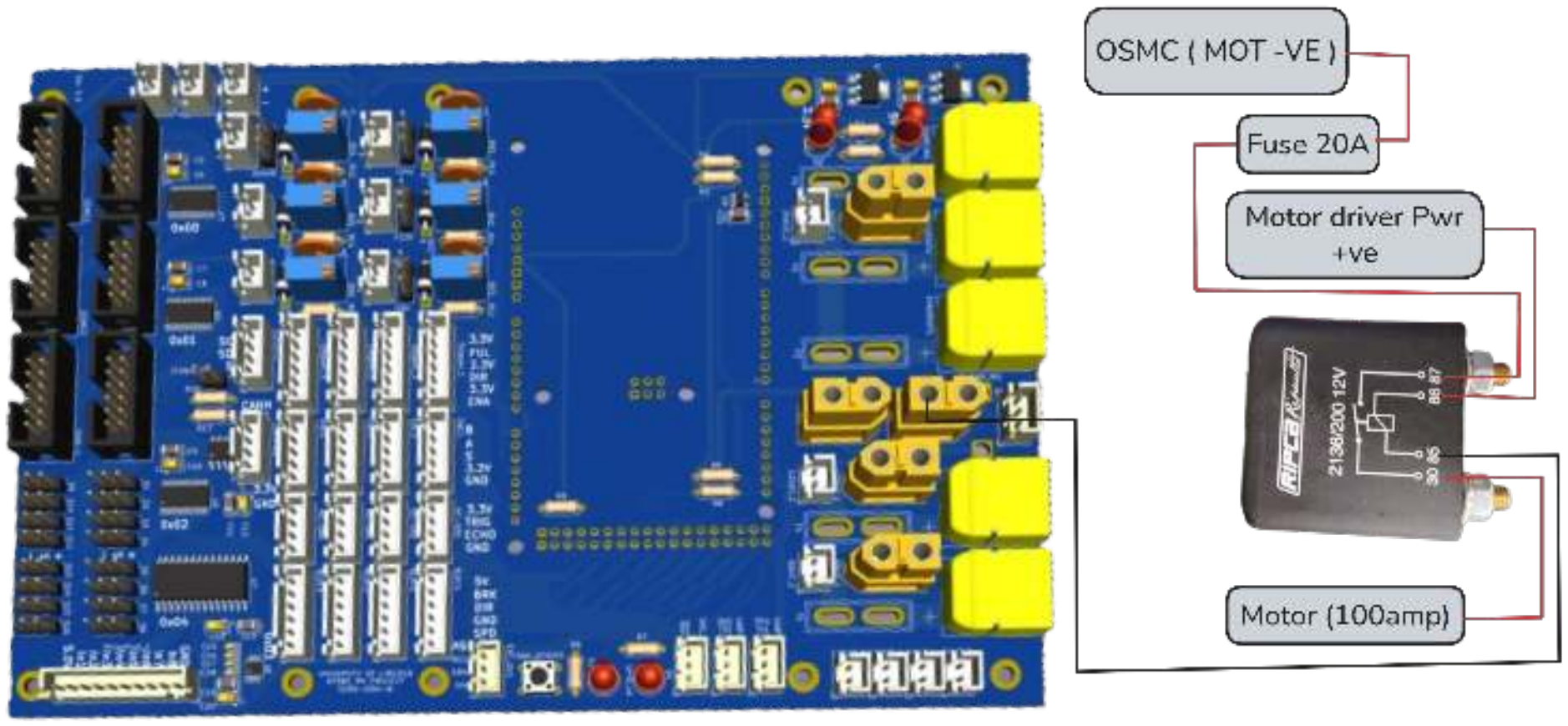

Figure 29: Power Relay

## OSMC Fan

Mount the OSMC Fan on top of the OSMC with standoffs.

Power the fan from OSMC's aux pins in Table 11 and Figure 30.

Table 11: Connection table for OSMC Fan

| Fan Pin | Destination | Notes |
|---|---|---|
| Fan:+12V | OSMC: Auxiliary +12V Supply | Built-in fan wire connection |
| Fan:GND | OSMC: Auxiliary Ground | Built-in fan wire connection |

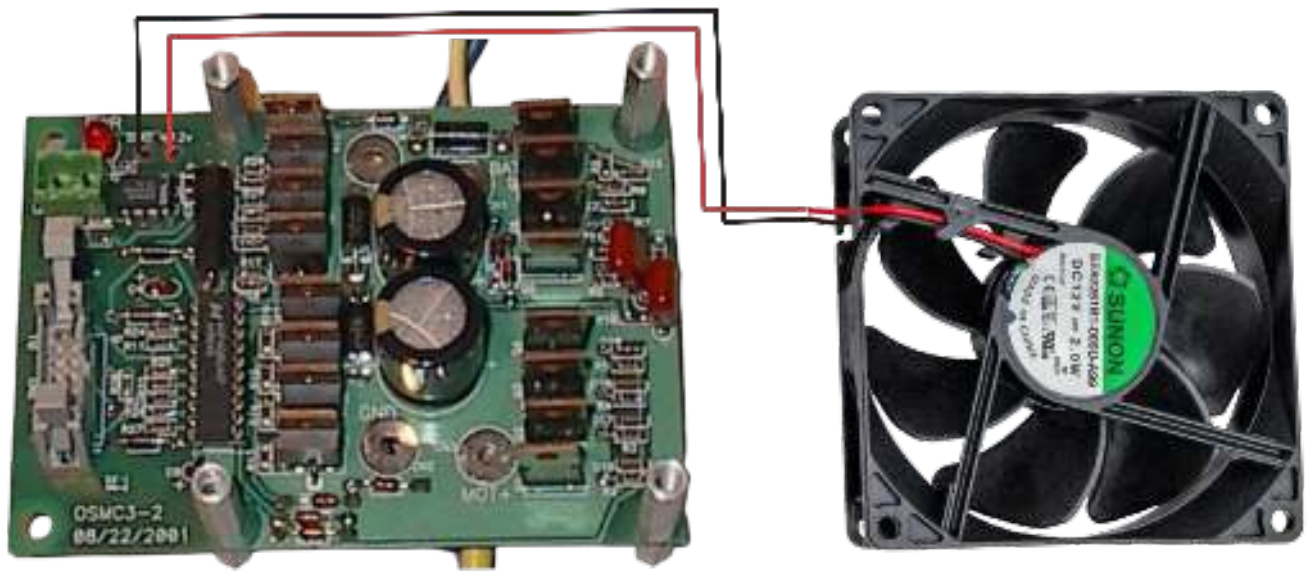

Figure 30: OSMC Fan connection with OSMC

**8-Way Relay Bank**

Mount the Relay Bank on the mounting board using standoffs.

Connect the relay bank logic and coil supply to the R4 eight channel relay interface on RelayPort_1, providing the individual relay control signals and common supply reference. Terminate load wiring at the relay screw terminals and routed to the target circuits according to whether normally open or normally closed behaviour is required, as shown in table 12 and Figure 31. Keep wires short.

Table 12: Connection table for 8-Way Relay Bank

| Relay Bank Pin | Destination | Notes |
|---|---|---|
| Header block | R4: RelayPort_1 | 8-pin connector (INx pins + GND + VCC) |

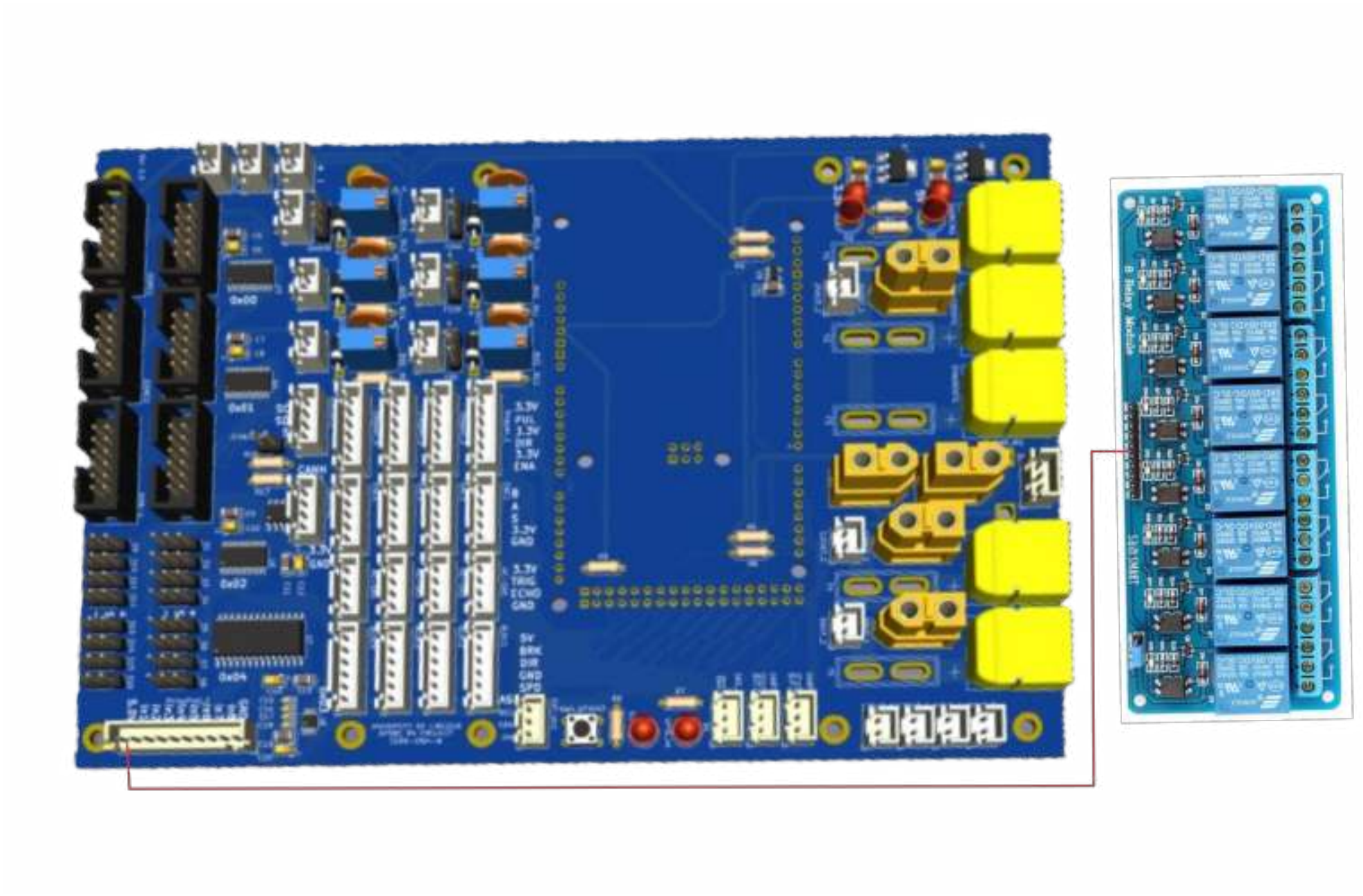

Figure 31: 8–Way relay bank

### Mounting Board Assembly installation into Donor Vehicle

Install the completed Mounting Board Assembly under the seat area of the Donor Vehicle, positioned above the inbuilt battery and secure using Velcro to prevent movement while allowing rapid removal for inspection and servicing.

The final placement, cable routing entry points, and access space around connectors for maintenance, are shown in Figure 32 and Figure 33.

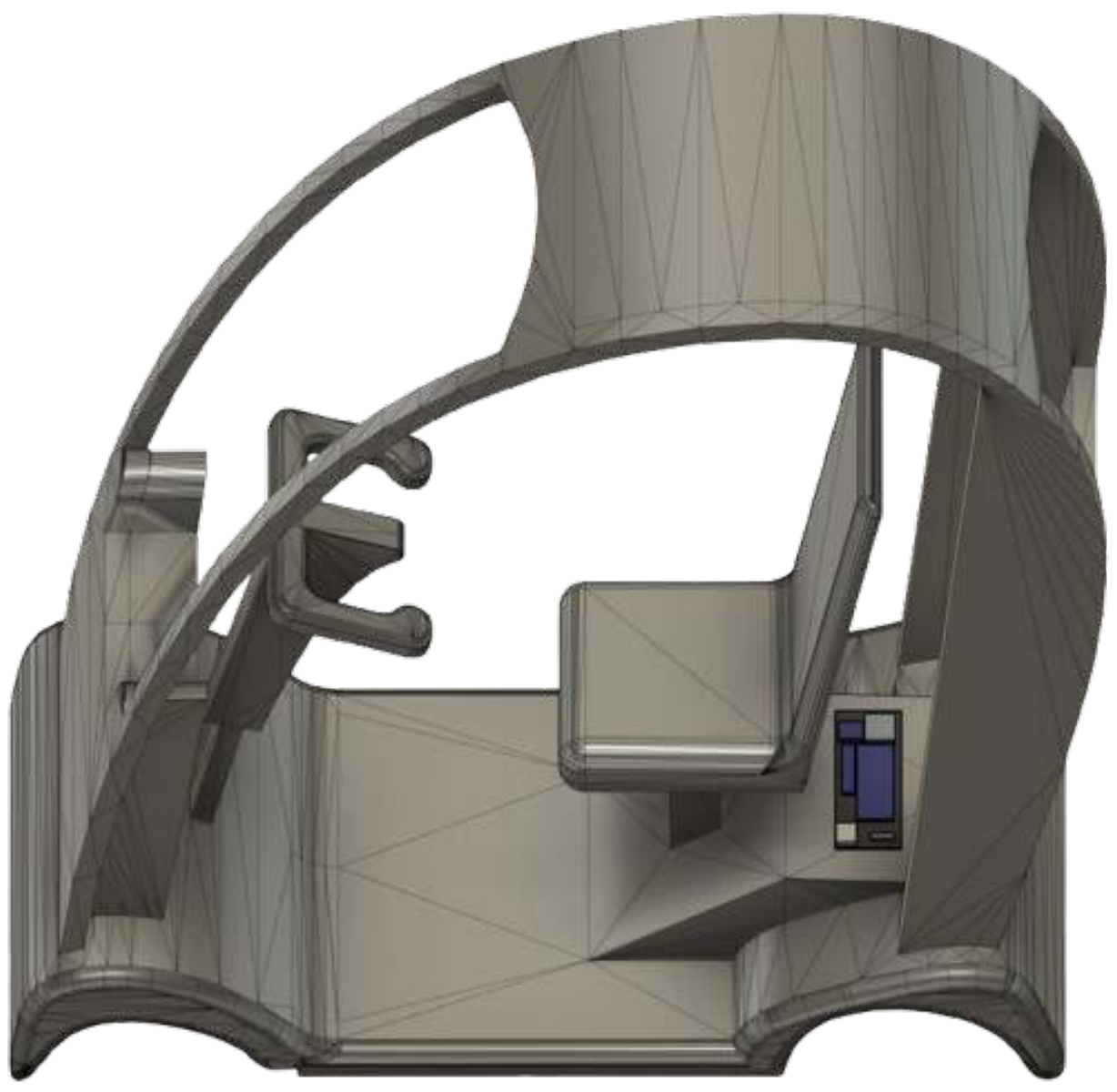

Figure 32: Side view

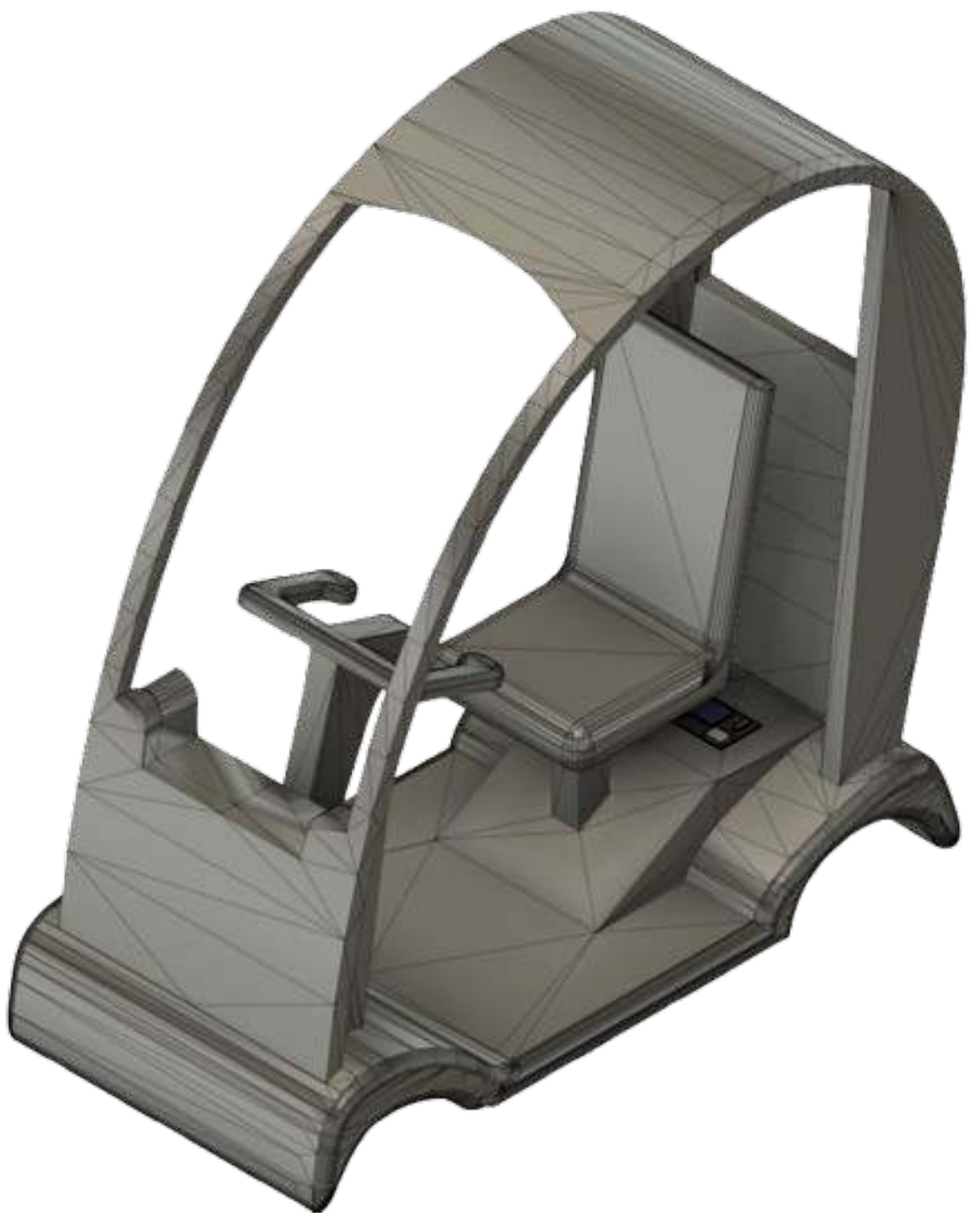

Figure 33: Corner side view

### Device connections

Connect the Wifi Router to 12V power from the DC/DC 24→12, using a suitable connector for the specific Wifi Router.

Place Laptop on Laptop Stand.

Connect the laptop to the DC/DC 24→19.

Use USB 3.0 to connect the Depth Camera to the laptop. Secure the cable to the vehicle using sticky-back Velco.

### Battery connection

Connect the GND distribution point to the battery's negative terminal (as in shown in table 13 and Figure 46).

Wire the (100A) fused battery positive to the input side of the main circuit breaker.

(You can use the original connector pair from the Donor Vehicle battery for these connections.)

Table 13: Connection table for Battery

| Battery Terminal Pin | Destination | Notes |
|---|---|---|
| Battery: +ve: 100A fused | Circuit Breaker: 2 | thick wire |
| Battery: -ve | GND | thick wire |

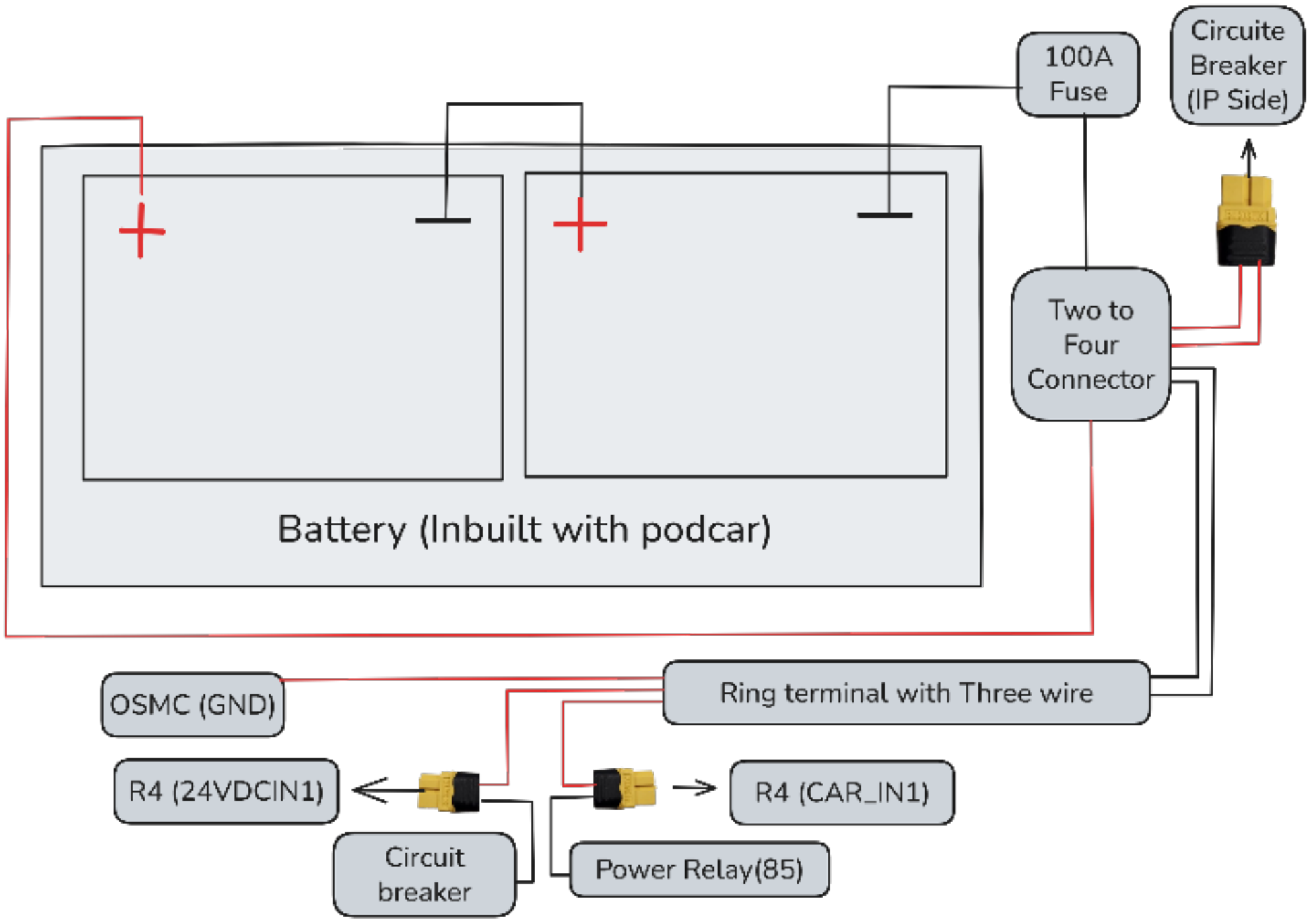

Figure 34: Battery connections

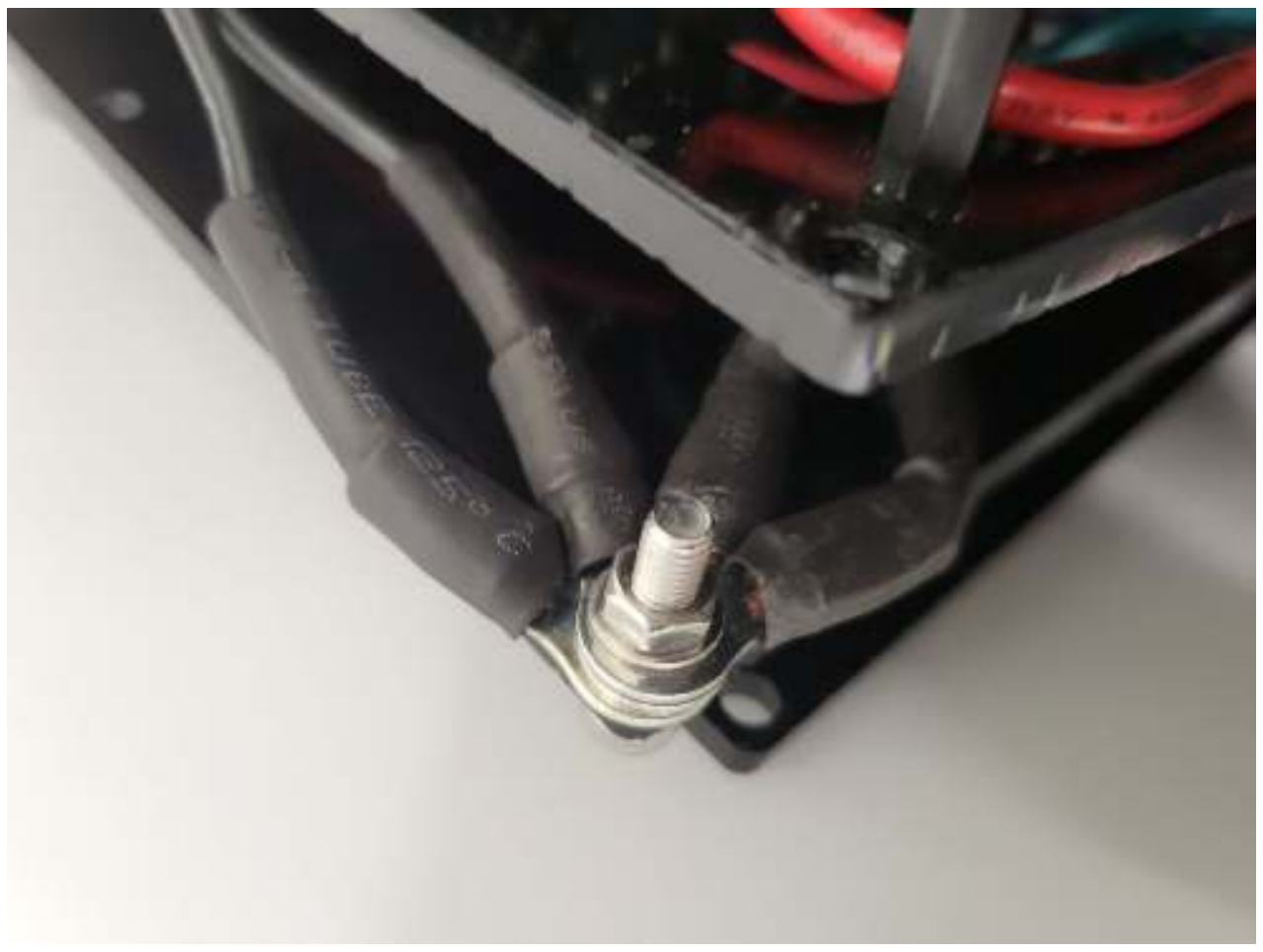

(a) Ring terminal with three pin corner for battery negative

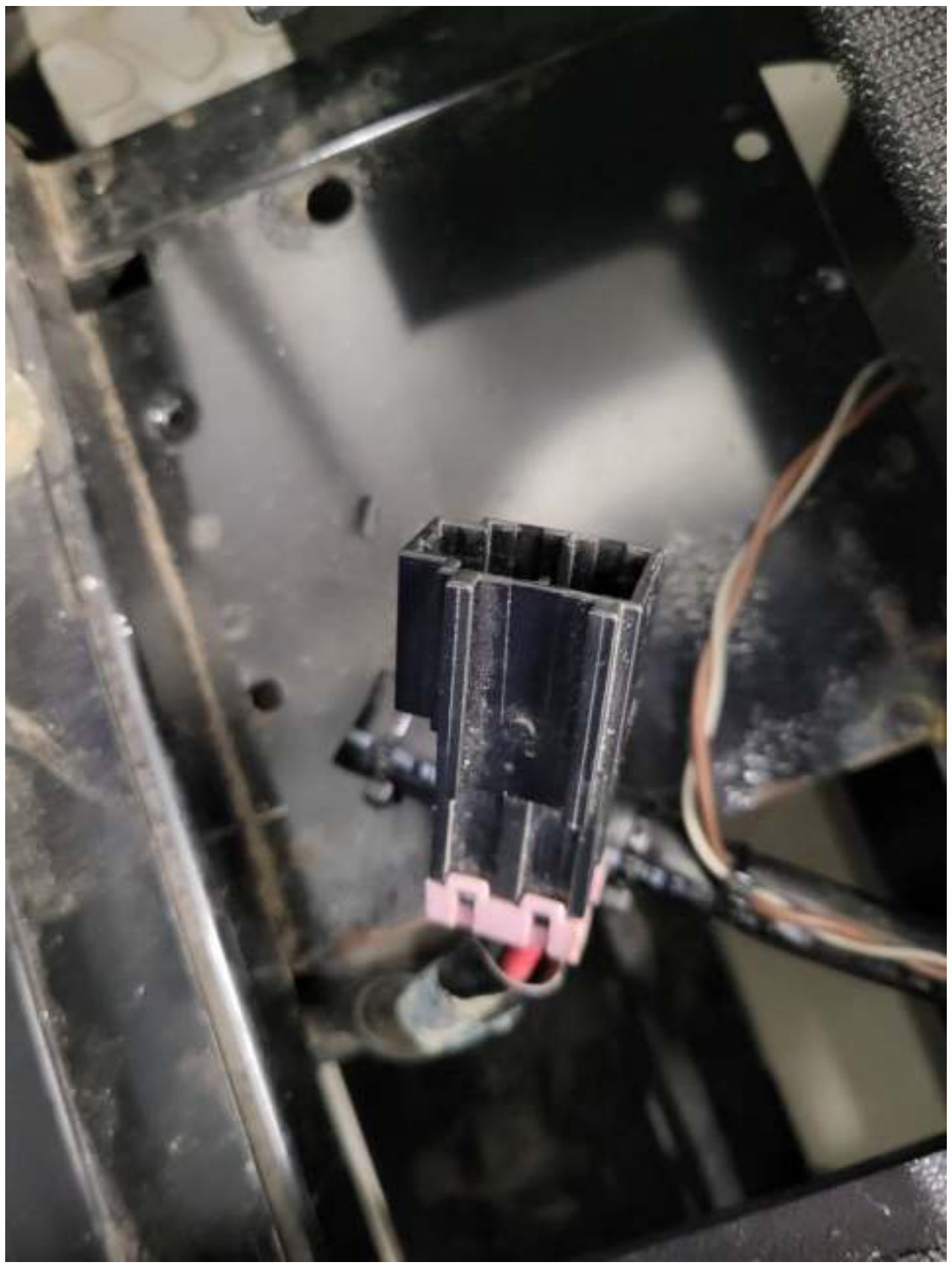

(b) Two to four wire converter for battery wire

### Software

The OpenPodcar2 software is a complete migration of OpenPodcar from ROS1 to ROS2, based on the ROS2 Humble distribution, with simulation support implemented in Gazebo. The ROS2 `OpenPodcar_V2` package consists of sub-packages namely; `pod2_description`, `pod2_bringup`, `pod2_navigation`, `pod2_sensor_tools`, `pod2_msgs`.

### System Requirements

1. Ubuntu 22.04 clean install: https://ubuntu.com/tutorials/install-ubuntu-desktop
2. ROS2 Humble full desktop install: https://docs.ros.org/en/humble/Installation/Ubuntu-Install-Debians.html
3. Gazebo fortress install: https://Gazebosim.org/docs/fortress/install
4. `ros_gz` package could be installed as a binary package. Follow this guide step by step: https://index.ros.org/r/ros_gz/. This enables communication between ROS2 Humble and Gazebo fortress.
5. Install and configure R4 firmware: https://gitlab.com/charles.fox/r4pcb
6. Install depthcam drivers. Follow the guide for ROS2 link for Intel wrapper – https://github.com/IntelRealSense/realsense-ros#installation-instructions. Choose option 1 and click on Linux Debian installation guide and run commands mentioned line by line.
7. Install ROS2 integration for depthcam: In the same repo: Under installation and step 2 : install latest Intel RealSense SDK 2.0. Select option 2: Install `librealsense2` package from ROS servers: `sudo apt install ros-humble-librealsense2*`.

Then, install ROS2 Wrapper: option 1 → `sudo apt install ros-humble-realsense2-*`

Realsense ROS2 wrapper to visualize pointclouds on Pointcloud2 message type[1]: `ros2 launch realsense2_camera rs_launch.py align_depth.enable:=true pointcloud.enable:=true`

### Testing Installations

1. To test that ROS2 is installed properly.

- Add the following to your `~/.bashrc`: `source /opt/ros/humble/setup.bash`.
- Open two terminals, in first run: `ros2 run demo_nodes_cpp talker`, you should see `hello` in the console.
- In other terminal, run: `ros2 run demo_nodes_py listener`, you should see `I Heard`.
- In order to keep the nodes communication robust, set the `ROS_DOMAIN_ID` in your `~/.bashrc`. For example: `export ROS_DOMAIN_ID=0`

2. To test the Gazebo fortress is installed on the system, in the terminal run: `ign gazebo`.If it launches, you'll see the simulation software window.
   - To test the `ros_gz package`, run: `ros2 run ros_gz_bridge parameter_bridge chatter@std_msgs/msg String@gz.msgs.StringMsg`
   - View the topic in other terminal using: `ros2 topic list -t`

### Installation for OpenPodcar2

To use this package for testing and running simulations using Gazebo and ROS2 follow the below instructions:

1. If using Gazebo fortress, clone the repository following below commands.

- Make the new workspace, with src directory. `mkdir -p ros2_gz_ws/src`.
- Clone the repository using: `git clone --recurse-submodules -b Fortress https://github.com/Rak-r/OpenPodcar2.git`

2. After cloning the repository, check that you have `pod2_description`, `pod2_bringup`, `pod2_navigation`, `pod2_sensor_tools` in your `src` directory.
3. Now, build the packages from the root of the workspace directory using ROS2 package building tool colcon.

- Assuming you are in `src` directory, run: `cd ..`
- `colcon build --symlink-install`. This will build the packages and the `--symlink-install` is used to make changes in the packages in src directory and also changes in the install dircetory without re-building the package.

[1](For more details see: https://github.com/IntelRealSense/realsense-ros/issues/2295)

4. If everything works well, you will have three directories along with `src` named `install`, `build` and `log` . If colcon build fails to build any packages and shows `stderr` in terminal console, make sure all the dependencies are installed correctly.

5. In case of Step 4, try running: `rosdep install --from-paths src -y --ignore-src`.

## Calibration

### Depthcam calibration

The Depthcam needs physical calibration in order to achieve reliable SLAM and mapping operation. For this, place an object 10m away at the same height as the camera from the ground. Adjust camera to ensure that the object appears in the center of the camera image (same height measurement at different distances from the camera). This calibration is essential to verify that the camera is mounted parallel to the ground to avoid irregularities in the mapping which may consider floor as an obstacle. Figure 35 shows the RGBD sensor mounted and checked with a spirit level to be parallel to the ground to ensure correct mapping.

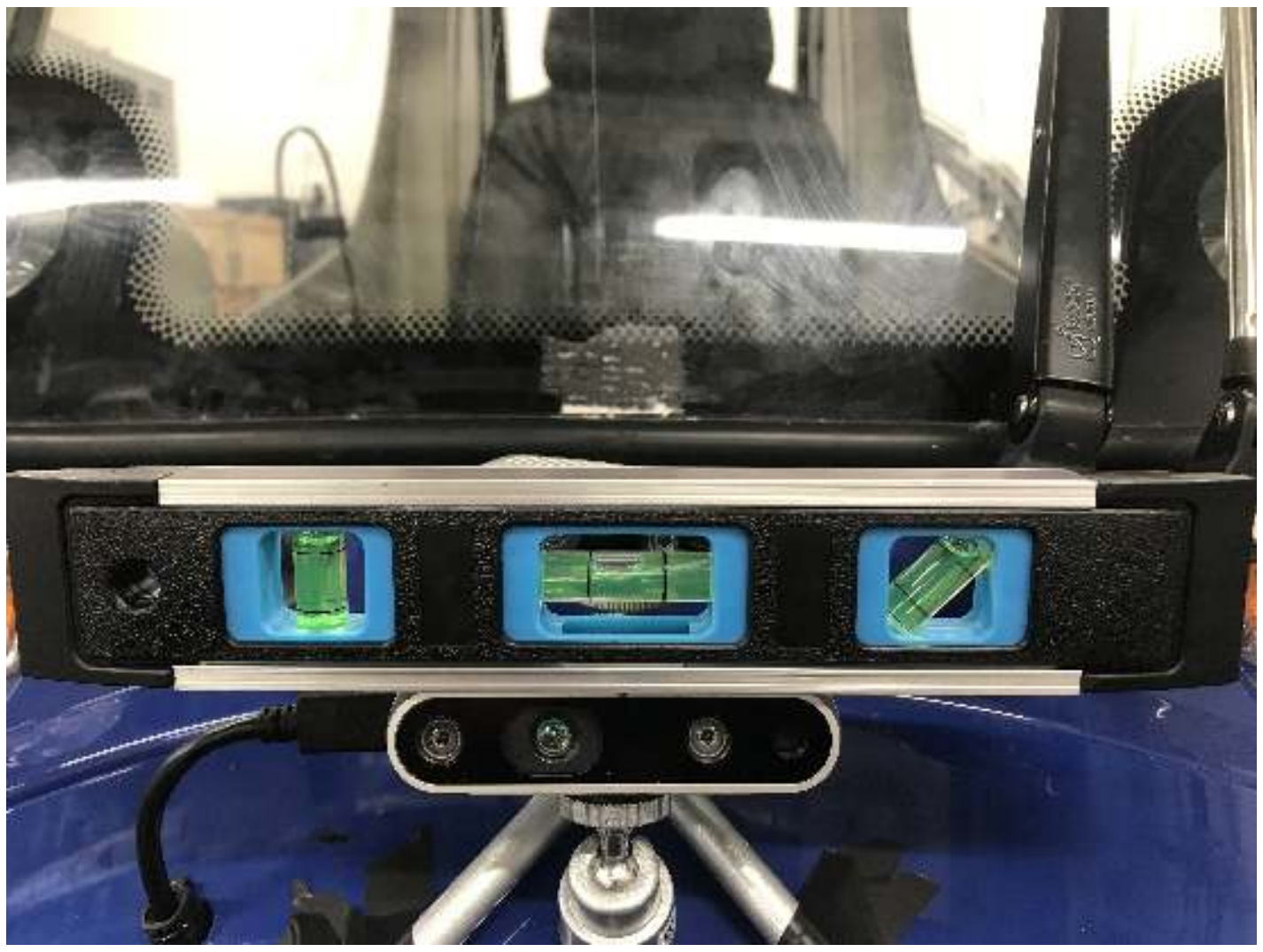

Figure 35: Depthcam level checked with spirit level

### Steering calibration

To calibrate the Ackermann steering angles on both left and right turnings various voltages are sent to the linear actuator and both inner and outer wheel angles are measured. The output result shown that the steering mechanism can be approximated as linear with some fluctuations while small turning angles. The mapping between steering angle and desired linear actuator voltage then could be computed using linear regression fit. Figure 36 represents the readings at different actuator voltages as well the angle values.

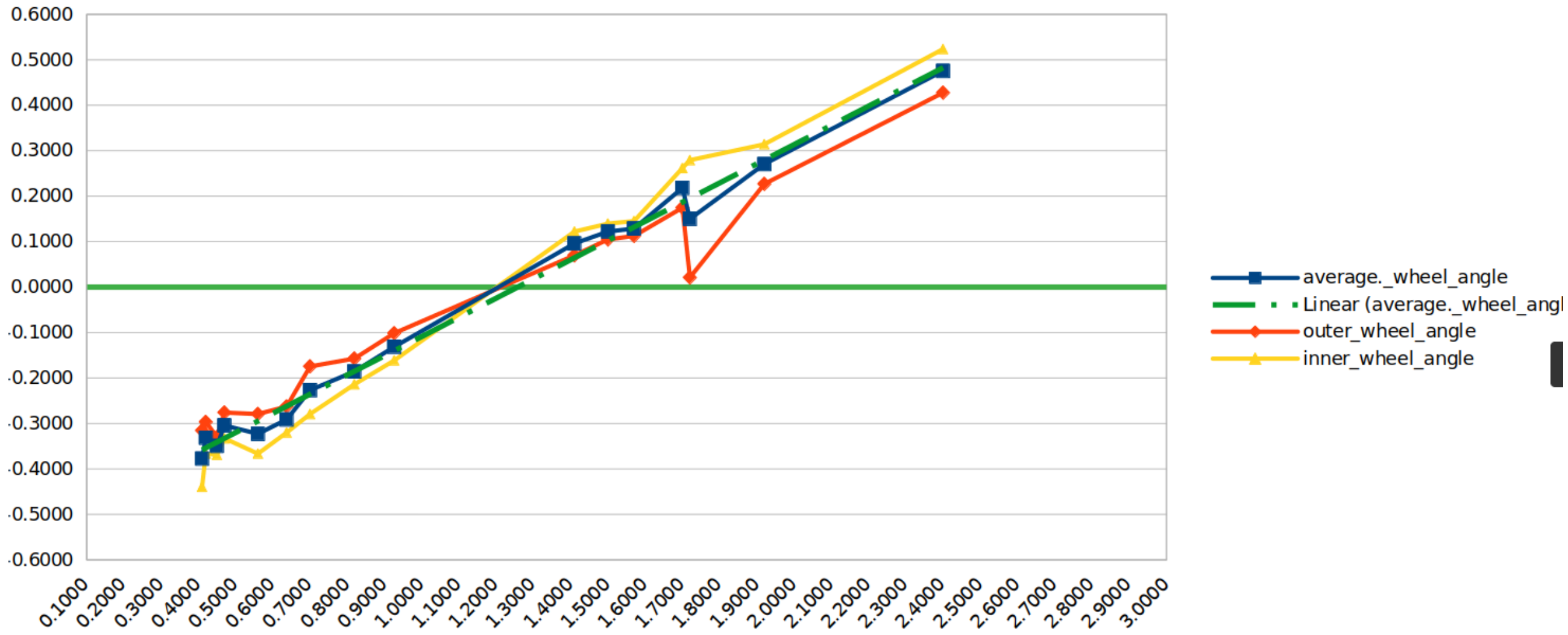

Figure 36: Steering angle readings vs Linear actuator voltages across full left and right turns

(Asymetries are caused by the linear actuator being not mounted exactly at the centre of the wheel track of the vehicle due to the mechanical mounting limits. The calibration controls for this.)

# Appendix C

# OpenPodcar2 User guide

## Teleop simulation

To test OpenPodcar2 in simulation with teleoperation, the `pod2_description` and `pod2_bringup package` are used. The description package consists the robot's URDF files in the `xacro` directory, meshes of the robot model, sensors in the `meshes` and the `launch` directory contains the `pod2_description.launch.py` file which launches the robot model's URDF equipped with RGBD sensor and the world file in the Gazebo fortress with a condition to start along the `rviz2` node. The launch file also consists the `ros_gz_bridge` package which is used to establish communication between Gazebo and ROS2. The parameter bridge is created for `/model/podcar/cmd_vel` topic from ROS to GZ, on this topic the Ackermann system plugin publishes the Twist messages.

1. To launch OpenPodcar2 with depth camera enabled, run the below launch file:

   - Launch with Rviz2 :

     ```
     ros2 launch pod2_decsription pod2_description.launch.py scan_node:=false RGBD_node
        :=true rviz:=true}
     ```

   This launch will launch the simulation in Gazebo. Turn on the play pause button to run the simulation. To view the active topics in ROS2, use `ros2 topic list -t` in the terminal window. To view active topics in Gazebo, use `gz topic -l` in the terminal window.

   This launch will launch the simulation in Gazebo. Remember to turn on the play pause button to run the simulation. To view the active topics in ROS2, use `ros2 topic list -t` in the terminal window. To view active topics in Gazebo, use `gz topic -l` in the terminal window.

### ROS2 simulation nodes

The `pod2_description` package consists of intermediate nodes which are used to convert the incoming messages over the topics `/model/podcar/odometry`, `/depth`, `camera/color/image_raw`, `camera/color/camera_info`, `/points` to publish on wall time and `/tf` is broad-casted from the wall time odometry message. This approach is employed to avoid the time related issues, tf errors and with an assumption that the simulation and the physical vehicle should work on same time.

The `pod2_bringup` package utilizes the ROS2 `teleop-twist-joy` package, in order to control OpenPodcar2 using the gamepad controller in the simulation. Different gamepads have been tested: XBox, Logitech Extreme3dPro, PS2, and an unbranded generic Linux USB gamepad. To test that your gamepad is connected to the system run `ls /dev/input`. In order to use other gamepad types you might have to create a `.yaml` config file (see https://github.com/ros2/teleop_twist_joy/tree/humble/config) and The `launch` directory contains two launch files for teleoperation. The `joy.launch.py` file is used for real-world teleoperation on the OpenPodCar2, where the vehicle speed is constrained in accordance with health and safety regulations. In contrast, simulation uses the `generic_gamepad.launch` file, which launches a generic gamepad interface and allows users to test the setup at different speed levels. (If using a non-XBox gamepad, you might need to check which physical buttons and axis are mapped to which logical ones. We recommend using https://flathub.org/apps/io.gitlab.jstest_gtk.jstest_gtk. This tool also provides calibration for the gamepad which might be helpful if deploying on the physical vehicle for tele-operation.) After launching the simulation, run the launch file:

- Launch the generic game-pad for simulation test and runs :

  ```
  ros2 launch pod2_bringup generic_gamepad.launch.py}
  ```

For safety operation, an enable button is set without which the commands cannot be transferred to the robot in both simulation and real-physical OpenPodcar2 on the gamepad as a DMH. These buttons can be easily adjusted from the configuration file present in the same package. The vehicle is controlled using the right joystick on an XBox gamepad. Vertical axes sets the speed, horizontal axes are used to steer the wheels.

### Autonomous simulation

To operate OpenPodcar2 with ROS2 SLAM and Navigation, repeat the above steps up to controlling the robot with game-pad and visualize in `rviz2`, then run the following launch files from the mentioned ROS2 packages:

#### `pod2_rtabmap`

The RTAB-Map package is used, to provide the visual odometry and to perform SLAM on custom gazebo worlds. RTAB-Map allows mapping using depth data, `PointCloud2` data and `LaserScan` data from any RGBD sensor. These are configurable according to the usage and test. The current setup has been tested with both `LaserScan` data and depth data for mapping.

To use LaserScan messages from RGBD camera, `pointcloud_to_laserscan` is utilized.

#### `pod2_navigation`

The package consists of the `launch`, `rviz2`, `maps`, `config` directories.

1. Config directory:

   This includes the parameter file which includes the parametrers for AMCL, `BT_Navigator`, Controller server, Planner server, Global and Local Costmaps, Behaviour servers, Map server.

2. Launch Directory:

   The launch directory consists of `OpenPodCar_V2_NAV2.launch.py` which is modified from the default `nav2_bringup` package for launching the specified nodes and takes the `nav2_dwb_smac` from the config directory which uses dwb as controller server and SmacPlannerHybrid as global planner server.

3. Maps Directory:

   The maps directory consists of maps created to test the NAV2 with prebuilt maps.

### Launching instructions

To run the Navigation with OpenPodcar2 while mapping launch RTAB-Map to start the visual odometry (VO) and Mapping. Following this start the NAV2 stack as described below. Moreover, for pure localization with prior map setup, the default localization method is Adaptive Monte Carlo Localization (AMCL) but RTAB-Map can also be configured to localization mode to perform further analysis.

1. Navigate to root of the ROS2 workspace:

   ```
   cd <workspace>
   ```

2. Source the ROS2 workspace:

   ```
   source install/setup.bash
   ```

3. Start RTAB-Map:

   `ros2 launch pod2_rtabmap rtabmap_sim.launch.py`

4. NAV2 Stack:

   `ros2 launch pod2_navigation OpenPodCar_NAV2.launch.py slam:=false rviz:=true amcl:=false`

5. If want to build the custom map using `slam_toolbox`, then change the mode to mapping in `mapper_params_online_async.yaml`

6. RTAB-Map saves the information in a database (.db) file from which 3D map, 2D occupancy grid map can be generated. RTAB-Map can be started in localization mode by passing the argument localization to true while launching the algorithm.

7. The map can be saved either by command-line: `ros2 run nav2_map_server map_saver_cli -f <map>`

### Docker support

The docker version is supported for ROS2 humble and gazebo Fortress due to LTS version of gazebo at the time project development. In future more version suppport will be added. Follow the below instructions for using docker version of OpenPodcar2 with simulation.

1. After cloning the repository from same above instructions, make sure docker is installed correctly.

2. Install rocker in the local machine to run GUI applications without any hastle inside the container. `pip3 install rocker`

3. Build the image: `docker build -t openpodcar2_docker`

4. Run the container with rocker: `rocker --x11 openpodcar2_docker`

5. If want to build the perception stack as well, then build the image with argument in below command.

   `docker build --build-arg INCLUDE_YOLO=true -t openpodcar2_docker`

6. Source ros2 and the workspace before running any ros2 nodes or launch files as done in normal local machine setup.

7. Test the setup by running the below.

   `ros2 launch pod2_description pod2_description.launch.py scan_node:=false rgbd_node:=true`

## Teleop physical vehicle

To launch the physical OpenPodcar2 with teleoperation mode, the higher-level incoming game-pad commands as Twist message `linear.x, angular.z` are converted to R4 protocol message which controls the main driver motor for forward and backward movement and linear actuator for controlling the steering for OpenPodcar2. User Datagram Protocol (UDP) is adopted to establish the communication between R4 and higher-level ROS2 stack.

In the `pod2_bringup`, another launch file is provided which starts the UDP server for establishing the communication with R4 and ROS2 interface. To control the robot, two nodes are created which subscribes to `linear.x` message from game-pad and publishes R4 protocol messages for main driver motor. The `Podcar_Steer_node.py` is a closed loop steering controller which subscribes to `angular.z` field of `Twist` message, the lower level `/R4_AINSTEER` message consist of the voltage information which then used to determine the desired voltage and to control the direction by calculating the feedback error from the desired feedback voltage and the subscribed `/R4_AINSTEER` showing actual voltage. The node publishes the `/R4_Command` messages to control the linear actuator for steering the robot. Moreover, in order to monitor the desired steering angle, the standard ROS2 `ackermann_msgs` are also published which also consists speed field. This `ackermann_msgs` is subscribed by main driver motor node named `Podcar_Motor_driver.py` and publishes to `/R4_OSMC1` topic.

1. Turn on the R4:

   ```
   The R4_Board feature a MCB for ease to switching on and off.
   ```

2. Turn on the wifi router:

   Connect the laptop/PC with the visible router wifi network name. Refresh the wifi search in the PC to see the available networks.

3. Navigate to root of the workspace directory:

   ```
   cd <workspace>
   ```

4. Source the ROS2 workspace:

   ```
   source install/setup.bash
   ```

5. With `rviz2`:

   ```
   ros2 launch pod2_description OpenPodCar.launch.py rviz:=true
   ```

6. Launch the R4-ROS2 `R4_Websockets`, `R4_Reciever`, `R4_Publisher` and `teleop_twist_joy` node:

   ```
   ros2 launch pod2_bringup R4_ROS2.launch.py teleop_node:=true
   ```

To drive the vehicle, hold down the physical DMH button on the gamepad (named as enable button in the configuration directory, config) then use its joycon to steer left and right and drive forward and backward.

### Autonomous physical vehicle

To operate the physical OpenPodcar2 with autonomous diving mode, first follow the same structure as mentioned in the teleoperation mode for the physical robot above, then perform the following additional steps. (The manual gamepad tele-operation is still available during autonomous driving).

The physical OpenPodcar2 is equipped with only single RGBD sensor, therefore the instructions involve related to RGBD sensor setup. The setup is tested with both depth data input as well as PointCloud2 data.

#### Starting the camera sensor:

After following the calibration guideline discussed in previous section, to launch the camera sensor;

1. Launch the depthcam Node:

   ```
   ros2 launch pod2_sensor_tools realsense_launch.py
   ```

2. Verify proper camera launch: Look for `/camera_color/_image_raw`, `/depth`, `/depth_camera_info` and `/cloud_in` topics using;

   ```
   ros2 topic list -t
   ```

3. Start the `pointcloud_to_laserscan` node:

   ```
   ros2 launch pod2_sensor_tools point_to_scan.launch.py
   ```

   This node will convert the incoming `PointCloud2` message into `LaserScan` message which is used by SLAM.

**Starting the robot model and rest of the stack:**

1. Launch robot URDF model: `ros2 launch pod2_description OpenPodCar.launch.py rviz:=true`
2. Launch RTAB-Map visual odometry (VO) and SLAM: `ros2 launch pod2_rtabamap rtabmap.launch.py`
3. Launch NAV2: The same command from the autonomous simulation instruction can be used to launch the physical vehicle `nav2` stack.

## Parameter tuning

To achieve reasonable results, parameters have been tuned to match the developed mobile platform, with the final parameters included in the release code. Where new builds deviate from the original design it is likely that local retuning will be needed.

Initially with default parameters setup, the DWB controller failed to make any progress with entering into recoveries and for straight line forward goals huge oscillations and jerks were observed. After fine-tuning the acceleration and deceleration parameters the jerks were reduced to minimal level. The usage of rotate to goal critic was removed due to Ackermann kinematics. Following this, weights for PathAlign and GoalAlign critics have increased to make the vehicle stay on the global plan. Moreover, the oscillation related parameters are slightly tuned.)

The robot_footprint parameter was carefully tested and set slightly larger than the actual vehicle dimensions to minimize close obstacle contacts. The obstacle avoidance behavior was evaluated indoors by setting close obstacle goals and adjusting the inflation_radius for the local costmap. The system effectively maintained close driving behavior without collisions.

Due to the frequent plan update, it was sometimes observed that the generated plan was not able to be followed by the controller server. The parameter which plays a crucial role for Ackermann model is `min_turning_radius` which needs to match the measurement from the real-physical vehicle. The term has been manually measured for OpenPodcar2 and was approximately resulted to 2.05m. For longer goals, it is observed due to less range of RGBD sensor, the planner server struggles to generate large plans. To improve planning for large goals, the dimension of the global costmap was increased to make the planner less constraint and it was observed that longer goals of more than 20m are generated. Tuning was required for the penalty weights to minimise heavily curve paths which might be difficult for the controller server to follow.

# Appendix D

## OpenPodcar2 ROS2 software stack

The build instructions are user guide suffice to build and use the system, including the ROS2 software. The following is provided as a description of the software architecture for users wishing to modify or debug the software or otherwise learn about how it works internally.

### Vehicle state

ROS2's `robot_state_publisher` and `joint_state_publisher` packages play distinct roles in broadcasting information about a robot's spatial configuration. The `robot_state_publisher` is responsible for computing the positions and orientations of all the robot's links based on the joint states, typically provided by a URDF (Unified Robot Description Format) file. The derived information is then published as transform over topics `/tf` and `/tf_static`. The `joint_state_publisher` package specializes in broadcasting information related to the positions, velocities, and efforts of all the joints within a robot.

### Kinematic control

OpenPodcar2 is as an Ackermann-steered vehicle. The ROS2 community uses `Twist` as the standard for sending mobile robot motion commands (ROS REP-119), so OpenPodcar2 takes `cmd_vel:Twist` commands as inputs. ROS2 also provides an Ackermann drive message (`ackermann_msgs/msg/AckermannDrive`) which contains fields for speed and steering angle.

ROS2 `Twist` messages can map exactly onto differential drive vehicle control, but cannot map exactly onto Ackermann control. This is because a twist is defined as a linear velocity and a rotational velocity. When the linear velocity is zero, a differential drive vehicle can still perform the rotational velocity by turning on the spot, but this is impossible for an Ackermann vehicle. An Ackermann vehicle can however dry turn its front steering wheels while it is stationary, which is not describable as a Twist. To map between this, OpenPodcar2 re-interprets turn-on-spot Twist commands as dry steering commands as follows:

Ackermann steering angle represents the angle of a single (tricycle-like) front virtual wheel which can be computed using the physical vehicle's wheelbase, the turning radius. The turning radius is defined by the distance between the centre of curvature and each wheel. The curve radii for the wheels can take positive and negative values, depending on the chosen coordinate system. Typically, a Cartesian coordinate system is established with the $x$-axis aligned with the facing direction of the robot and the $y$-axis to its left. In this setup, a left turn corresponds to a positive radius, while a right turn has a negative radius. When driving straight forward, the curve radius, and consequently the Instantaneous Center of Curvature (ICC), approaches plus or minus infinity. When the ICC is at plus or minus infinity, the perpendicular lines on all wheels become parallel. For simplicity, only one wheel is considered. Following the approach, the two angles namely; inner wheel and outer wheel angle can be computed using mathematical relation.

In the conversion, the `Linear.x` field of Twist messages represents the velocity which is assumed to be same for the Ackermann vehicles. To calculate the steering angle, the physical steering limit of real-robot should be known using this the steering angle could be determined.

### Manual teleoperation

An Xbox controller is used as the manual controller. `joy` is a standard ROS2 package which consists of node to interface this controller (or many others) to send messages of type `joy` on the topic `/joy`.

telelop_twist_joy is a standard ROS2 node which converts joy messages to Twist type cmd_vel messages. It includes an additional DMH configured on a button on the controller used via standard ROS2 YAML parameter file (RL for XBox). This DMH button must be held down on the controller in order for any other controls to have an effect.

The package provides scales for managing the linear and angular velocities of the twist message. To make sure that the speed scales are under safety threshold for OpenPodcar2's motor driver, which is currently set to the limit of 400, the maximum speed according to the odometry measurement is set to 0.2 m/s. To ensure this limit is followed by the teleop_twist_joy package, the scale in $x$ is set to 0.2 which on the low level motor driver reads 400. To test the if values are set at the high level ROS2 package, test the framework without pressing the DMH, and echo /R4_OSMC1 topic to visualize the max range (this should read 400). Otherwise it may damage the motor driver.

## Localization and Mapping

The ROS2 ecosystem provides several alternative tools for SLAM. OpenPodcar2 is equipped with Intel RealSense D435 RGBD camera which outputs RGB and depth image. The sensor also outputs the PointCloud2[2] ROS2 message which is the standard message used by stereo and RGBD sensors. The PointCloud message has been deprecated after ROS2 Foxy version. Therefore to comply with the sensor setup and input data, we use Real-Time Appearance based Mapping (RTAB-Map) [21], a graph based SLAM framework that supports RGB-D cameras by leveraging both visual appearance and depth information for 3D mapping and loop-closure detection.[3]

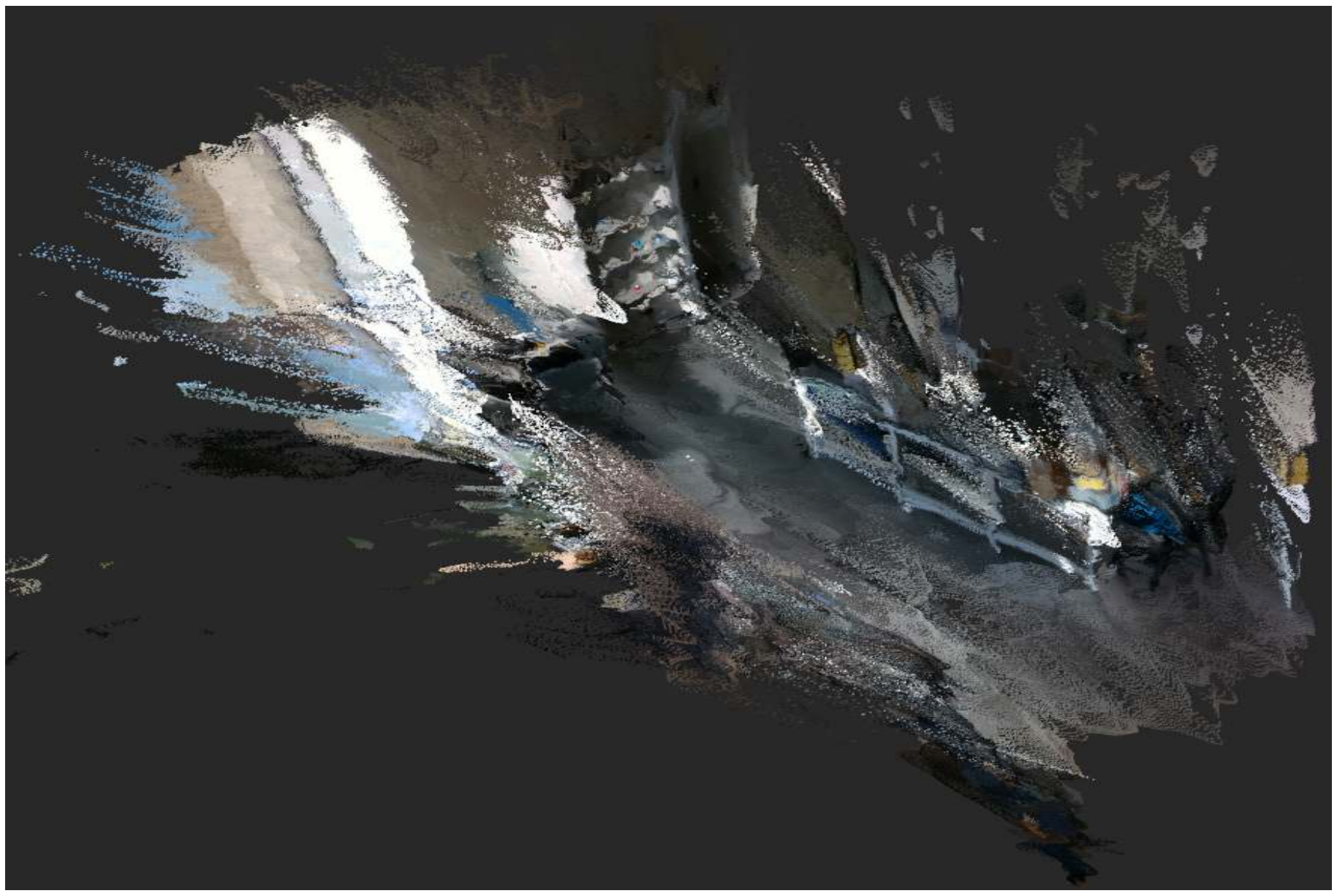

Figure 37: RTAB 3D Indoor Map in a tight lab space

The RTAB-Map package also provides the visual odometry named as RGBD odometry which is deployed in the vehicle to provide the pose information and odom to base_link transform. This has shown reliable results for indoor operations. Moreover, the RTAB-Map SLAM package is used to perform the the mapping and localization which takes the odom to base_link transform as input and corrects the pose information of OpenPodcar2 in the map frame. To ensure that the 2D occupancy grid map space is marked as free space from the RGBD camera mounted in-front of the vehicle, there is need to tweak the Grid/RangeMax, Grid/RangeMin and Grid/RayTracing according to the range of the RGBD camera. The Intel Realsense D435 mounted on the vehicle has range of 10m but offers good accuracy for 5m range. The system checked with both ranges and default is set to 5m. The outputs are the `map` to `base_link` transform, and the corrected pose information of the robot over the topic `/localization_pose` which can be used for further applications.

[2] https://docs.ros.org/en/ros2_packages/rolling/api/sensor_msgs/interfaces/msg/PointCloud2.html

[3] RTAB-Map was selected over LiDAR oriented alternatives such as 2D SlamToolbox [22], Hector Slam [20], Cartographer [16]. Also nav2 (which is primarily a planning framework) includes an AMCL localizer-only (no mapping) similar to ROS1's AMCL.

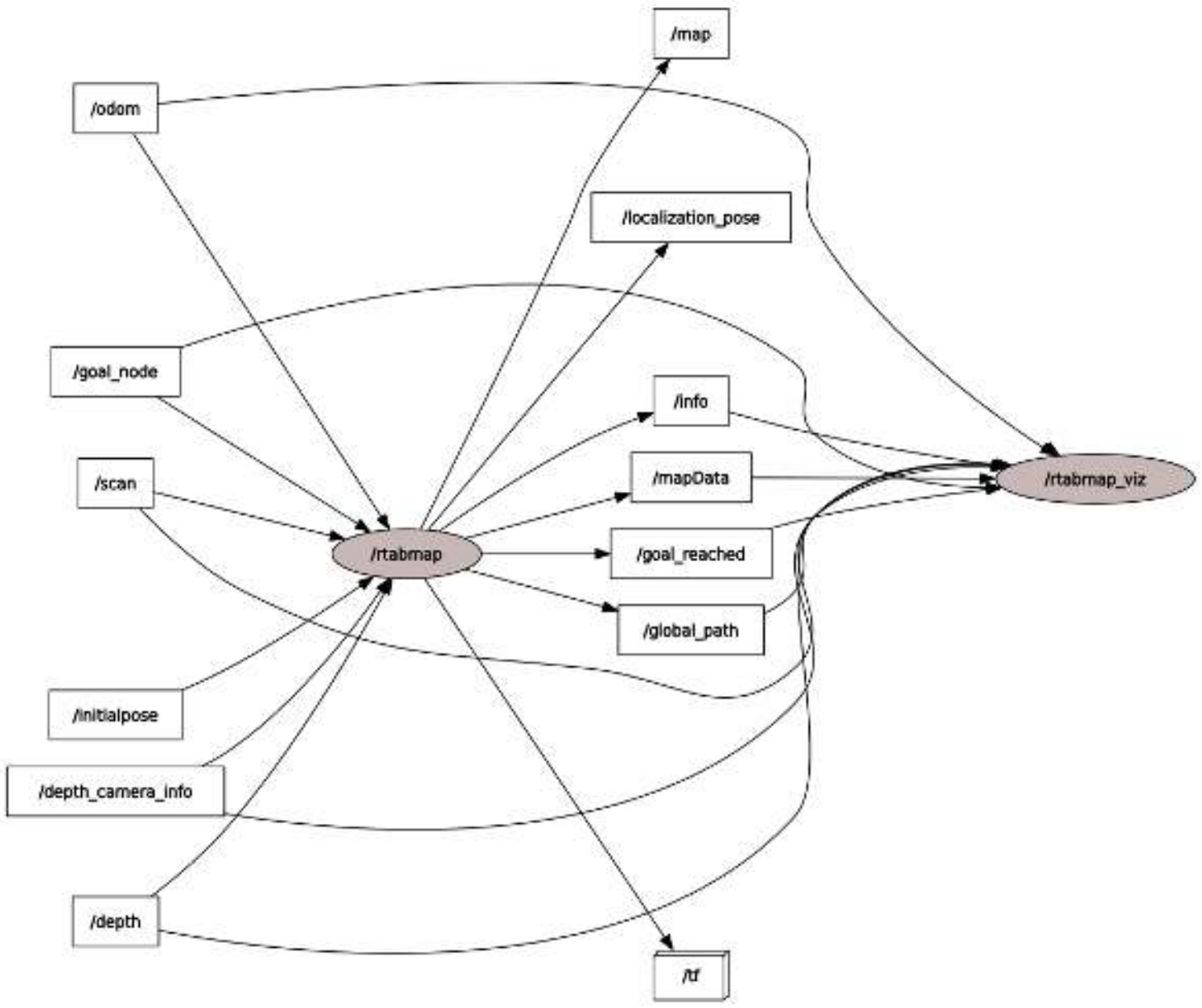

Figure 38: ROS2 RTAB nodes and messages

The pose only gets updated when the robot explores the environment, however to achieve continuous pose updates we can enable the RTAB-Map SLAM `map_always_update` parameter. Figure 37 shows the 3D map of indoor lab space using RTAB-Map while Figure 38 highlights the topics subcribed by RTAB-Map node via ros2 node graph

For more accurate odometry estimation, more RGBD sensors can be used for wider field-of-view (FOV), IMU sensors for orientation estimation can be fused together using the ros2 `robot_localization` package.

## Navigation

The ROS2 ecosystem includes a navigation stack, Navigation2 (`nav2`) [23], designed to facilitate the smooth navigation of a vehicle from one pose to another while avoiding obstacles along the path. Navigation2 comprises nodes which host plugins providing navigation capabilities, including cost map generators, planners, controllers, and recovery behaviors. OpenPodcar2 integrates these packages within its system architecture to provide onboard autonomous navigation:

- *nav_costmap_2d:* This `nav2` subpackage reads the map and publishes the global costmap. It also reads the sensor and publishes a local costmap (which does not use any map data). It inflates the obstacles to create costs close to them in both costmaps.
- *SMAC Planner:* Planners calculate a path (a.k.a. route) from a source point to a destination point. We use `nav2`'s SMAC planner, which is designed for Ackermann steering kinematics. It implements a hybrid algorithm, smac-planner-hybrid, which combines local Reeds-Shepp/Dubin paths in open spaces with global A* around obstacles. (It also has options to use two other algorithms). The Reeds-Sheep model is used as default for OpenPodcar2 it allows for the reversing behavior.

- *Dynamic Window Band controller (DWB):* Controllers[4], transform a spatial path into a temporal trajectory, using the local costmap. We are using DWB, which is a modified version of DWA. The DWB controller due to its collision avoidance ability is preferred as default controller for the podcar at the time of the development of the automation stack.
- *Behaviour Tree Navigator:* A Behavior Tree action server (not messages or services) is used for communication between the planner and controller, and user goals. User goals can include moving to a pose, but also others such as area coverage or waypoint list visiting. It implements the NavigateToPose and NavigateThroughPoses task interfaces.
- *Behaviour/recovery Servers:* For special cases of recovery from failed controls, behaviour servers add abilities to reverse and recover to a previous state. `nav2_behaviour_server` features various kind of actions: spin, backup, wait, Assisted teleop, and Drive on-heading.

For safety, navigation requires the robot's `base_link` (the center of the robot, not the the front bumper) should be within the bounds of the map. The planner will do nothing if either the origin or destination are unknown in the map. Moreover, due to less FOV of the RGBD sensor, the global costmap warning 'Robot is out of bounds' is triggered. A short manual drive around the operating area should thus be performed to build an initial map before engaging navigation. Figure 39 illustrates the ROS2 node graph and topics associated with the `nav2` stack. This figure provides a visual representation of the various ROS 2 nodes and their interconnections within the Nav2 framework.

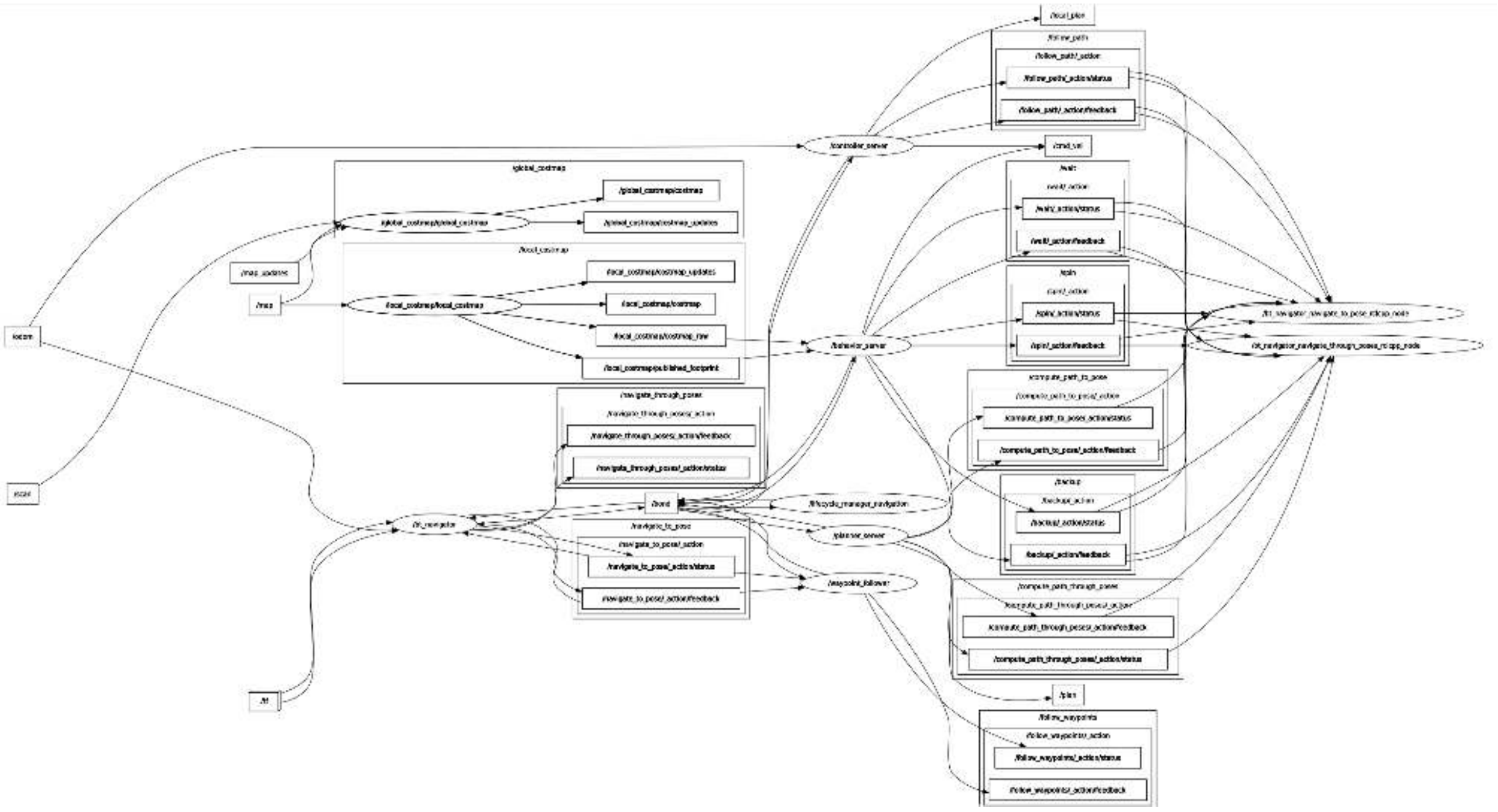

Figure 39: ROS2 nav2 nodes and messages

## Pedestrian/object detection and tracking

The RGBD camera is used for pedestrian detection and tracking as well as for SLAM. An off-the-shelf ROS2-wrapped YOLOv8[5] is linked to the camera to perform and report pedestrian and vehicle detection and tracking

[4] known as local planners in ROS1

[5] https://github.com/Rak-r/yolov8_ros2_OpenPodCarV2.git

in 3D space. YOLOv8 2D detection reports bounding box co-ordinates as ($x$ center, $y$ center, width of the box, height of the box); class name, class ID, confidence score are also extracted. YOLOv8 includes 2D tracking on these detection with BotSORT [4] and ByteTrack [34] which provides acceptable real-time performance but suffers from false re-id errors. ByteTrack is used as default here. An Intel depth camera ROS2 wrapper is provided by the Intel Realsense SDK[6] which publishes the data over `\camera_image_raw` for RGB image, `\depth` for the depth image, and `\depth_camera_info` reporting the intrinsic cameras parameters.

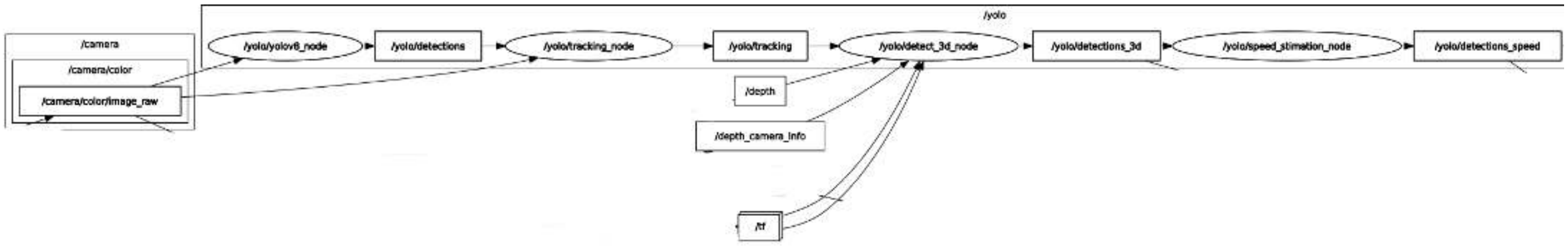

Figure 40: ROS2 nodes for pedestrian detection and tracking

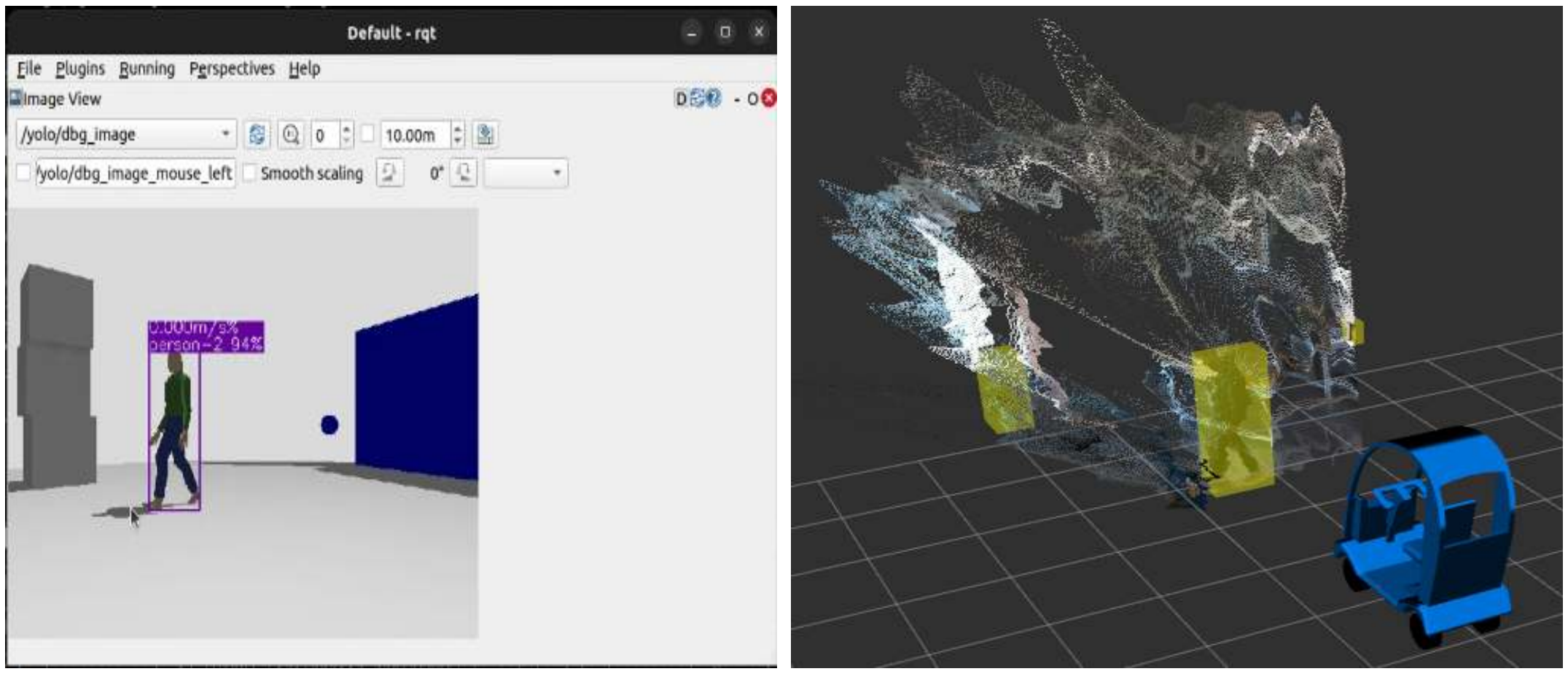

Figure 41: Pedestrian detection results in simulation (left) and real-world deployment on the physical OpenPodcar2 platform (right).

All messages are published under the namespace of the `/yolo`. It is used with similar structure to ROS2's standard `vison_msgs` message types. The 2D detection node subscribes to camera RGB messages and publishes `/Detection` messages. The ROS2 YOLOv8's tracker node subscribes to these Detection2D messages, and keeps track of the objects over each frame, publishing `DetectionArray` messages showing the tracks. Figure 40 shows the detection and tracking nodes in ROS2 followed by Figure 41 illustrating pedestrian detection outputs from the unified perception stack in simulation (left) and during execution on the physical OpenPodcar2 platform (right), confirming functional equivalence between simulated and real-world operation.

The ROS2 YOLOv8 wrapper's 3D projection then takes as input these 2D tracks and the $z$-coordinate from the original RGBD image and outputs 3D locations. It publishes the 3D bounding boxes in real-world coordinates ($x$, $y$ and $z$) along with the dimensions (length, width and height) in meters as `BoundingBox3D` messages.

The 3D bounding box messages reporting the center position of pedestrians are fed to a Kalman filter as noisy observations. This creates a new message containing both smoothed location estimates and speed estimates. Figure 42 illustrates the full pedestrian detection and tracking stack deployed on the physical OpenPodcar2 platform. The left panel shows the modified depth image (masked_depth), in which regions corresponding to YOLOv8 2D pedestrian detections are removed and marked as "no data". This masked depth image is generated by feeding the detected bounding boxes back to the perception wrapper. The same masked depth stream is provided to the localisation, mapping, and navigation components. As shown in the right panel of Figure 42, the vehicle operates in mapping mode while the pedestrian remains detected via 3D markers, but is excluded from the depth data used for feature extraction and costmap updates. This prevents dynamic pedestrian motion from perturbing localisation or triggering unnecessary path replanning.

[6] https://www.intelrealsense.com/sdk-2/

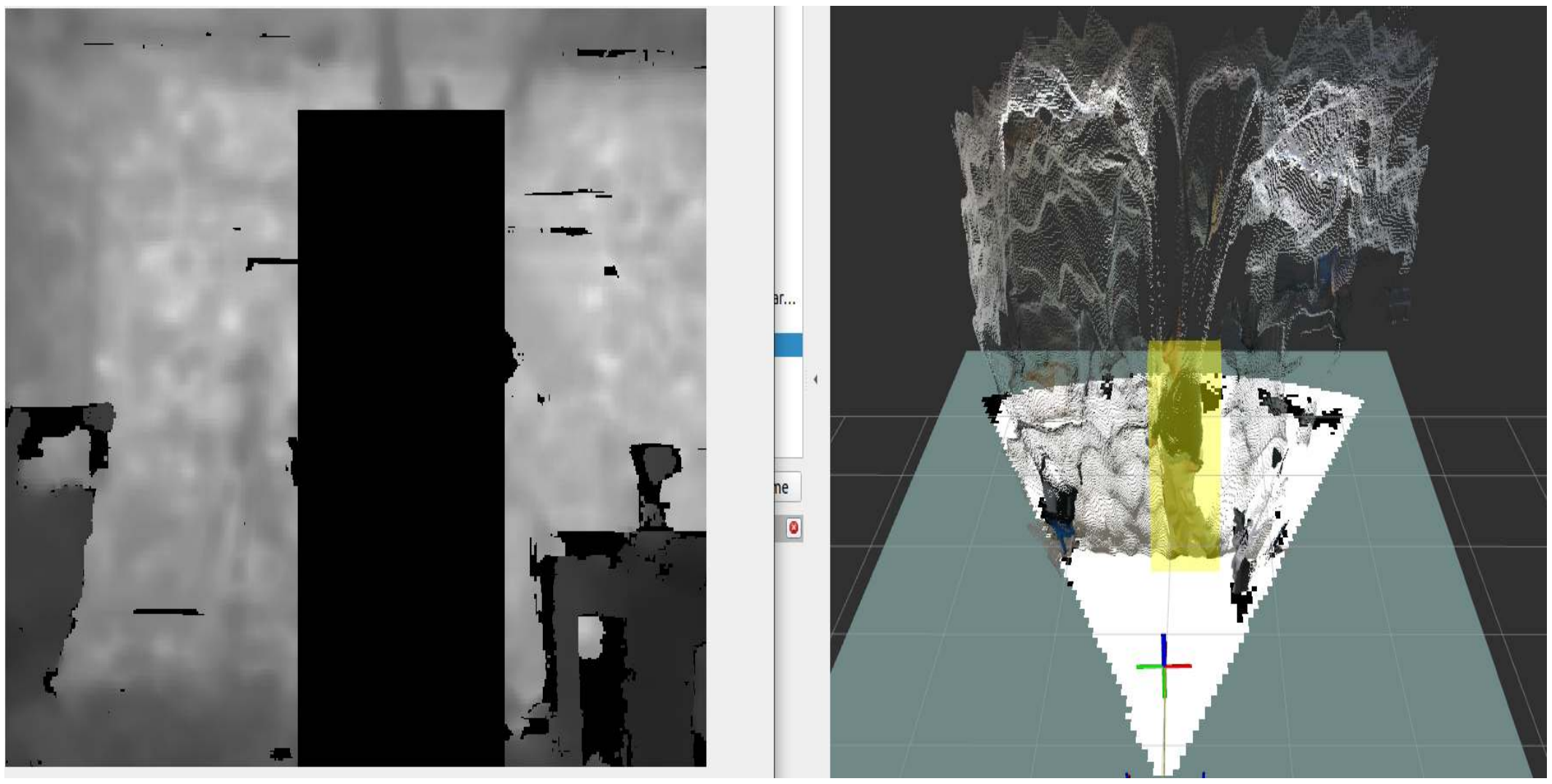

Figure 42: Pedestrian masking and navigation during real-world deployment of OpenPodcar2. (Left) masked depth image generated from YOLOv8 2D detections. (Right) real-time mapping mode using the masked depth stream for localisation and navigation, while the pedestrian remains tracked via a 3D bounding box but excluded from environmental mapping.

## Simulation

A ROS2 simulation of OpenPodcar2, is provided, using the open source igntion Gazebo simulator. To handle time synchronization issues at the software level, it is necessary that both the ROS2 stack and Gazebo simulation should work on the same time. Although ROS2 provides the configurable parameter for the nodes named `use_sim_time` which could be set to a boolean value of either true or false, there are still some issues faced: lookup transforms, message filter dropping when setting the the whole NAV2 stack with Gazebo which generates lags in the system and ultimately failures. Nevertheless, Navigation2 must be implemented on the real/physical OpenPodcar2, so it makes sense to set the stack with working on wall time/system time. To achieve this condition, the Gazebo `ackermann` steering plugin is used in our stack which deals the kinematic control for the robot, lidar sensor system plugin to receive the `LaserScan` data in case of using lidar, `rgbd` sensor system plugin for simulating the depth image and odometry publisher plugin to get the ground-truth odometry data out from Gazebo needs to publish the data on wall time and such condition in Gazebo (making use wall time) is not a very discussed topic and in the ignition Gazebo fortress makes it more of a highly debugging task. To handle this condition, custom ROS2 nodes are created which subscribes to the gazebo output topics and publishes the topic data on wall/system time.

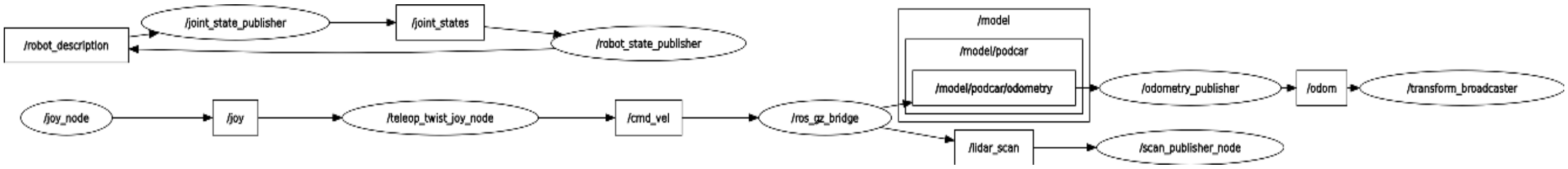

Figure 43: ROS2 GZ Bridge node graph with custom nodes setup.

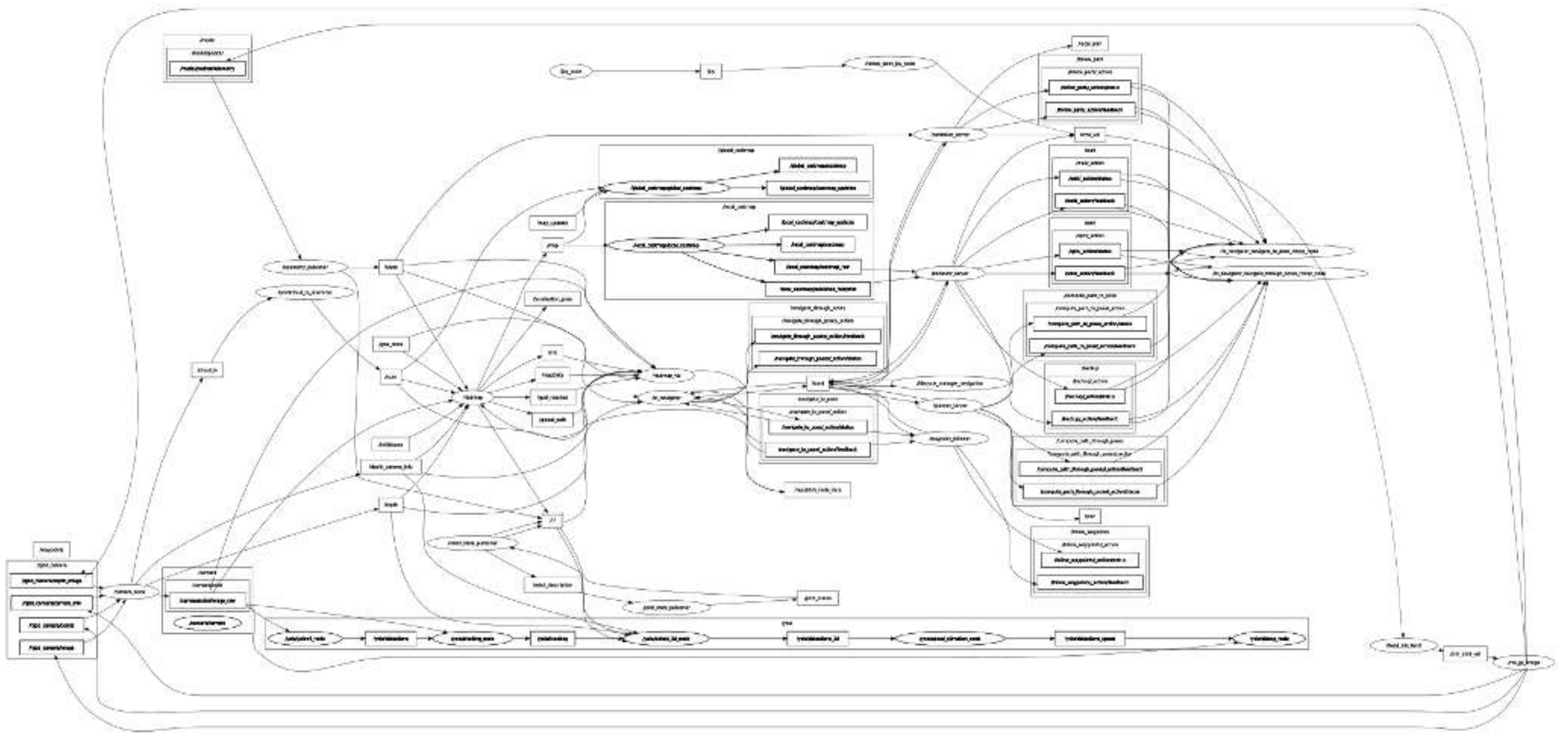

Figure 44: ROS2 NAV2 full stack setup with Gazebo sim.

To integrate ROS2 and newly released version of Gazebo, the community has introduced a new package called `ros_gz_bridge` which features an interface exchange between ROS2 messages and Gazebo protobuf messages.

The package could be used to bridge incoming topics from ROS2 towards Gazebo or vice-versa. The other approach to generate the communication between ROS and Gazebo, is the direct embedding of ROS nodes into Gazebo plugin style codes which is still provided for application specific tasks. Using the above mentioned package, we created nodes which subscribes to bridged topics from Gazebo to ROS2, `/lidar_scan`, `/model/podcar /odometry`, `/rgbd_camera/image`, `/rgbd_camera/depth_image`, `/rgbd_camera/points` and `/rgbd_camera/camera_info` and the message field header which represents the timestamp is set to wall/system time. The nodes `scan_publisher_node`, (`/scan` topic), `odometry_publisher` (`/odom` topic), `RGBD_node` outputs the following topics with same data and with system timestamps. The custom `ros_gz_bridge` node with messages established for OpenPodcar2 is shown in Figure 43. In order to provide the transforms on the system time, a transform broadcaster node is created which subscribes to the `/odom` topic, to broadcast the `odom` to `base_link` transform for mapping, localization and navigation. The full set of working nodes and active topics in simulations is shown in Figure 44.

The `pod2_description` package contains the mesh files for loading the robot in `rviz` and for simulation purposes. To make the full custom stack operate on single time source, following custom nodes were created:

- `laser_sim2real.py`: In case of using a simulated lidar sensor this node subscribes the lidar scan topic from GZ laser/lidar plugin and publishes `sensor_msgs/msg/LaserScan` to ROS2 over topic `/scan` with changing the time stamp to wall time.
- `odometry_wall_time.py`: The node subscribes the ground truth odometry topic from GZ and publishes `nav_msgs/msg/Odometry` to ROS2 over topic `/odom` with changing the time stamp to wall time. The node also broadcasts a transform between `odom` and `base_link`.
- `RGBD_wall_timer.py`: The node subscribes the `rgb`, `depth_image`, `pointcloud2` and `camera info` topic from GZ RGBD camera plugin and publishes `sensor_msgs/msg/Image` to ROS2 over the topics `/depth`, `/depth_camera_info` , `/camera/color/image_raw` and `/cloud_in` with changing the time stamps to wall time. The naming conventions here are focused to be the same as what the real RGBD sensor publishes.

The `pod2_bringup` package consist of all the control nodes for simulation which utilizes the ROS2 `teleop_twist_joy`

package to control the the robot using a game pad. The system is tested with a branded XBox game pad and a compatible generic USB gamepad. The gamepads must be calibrated properly.

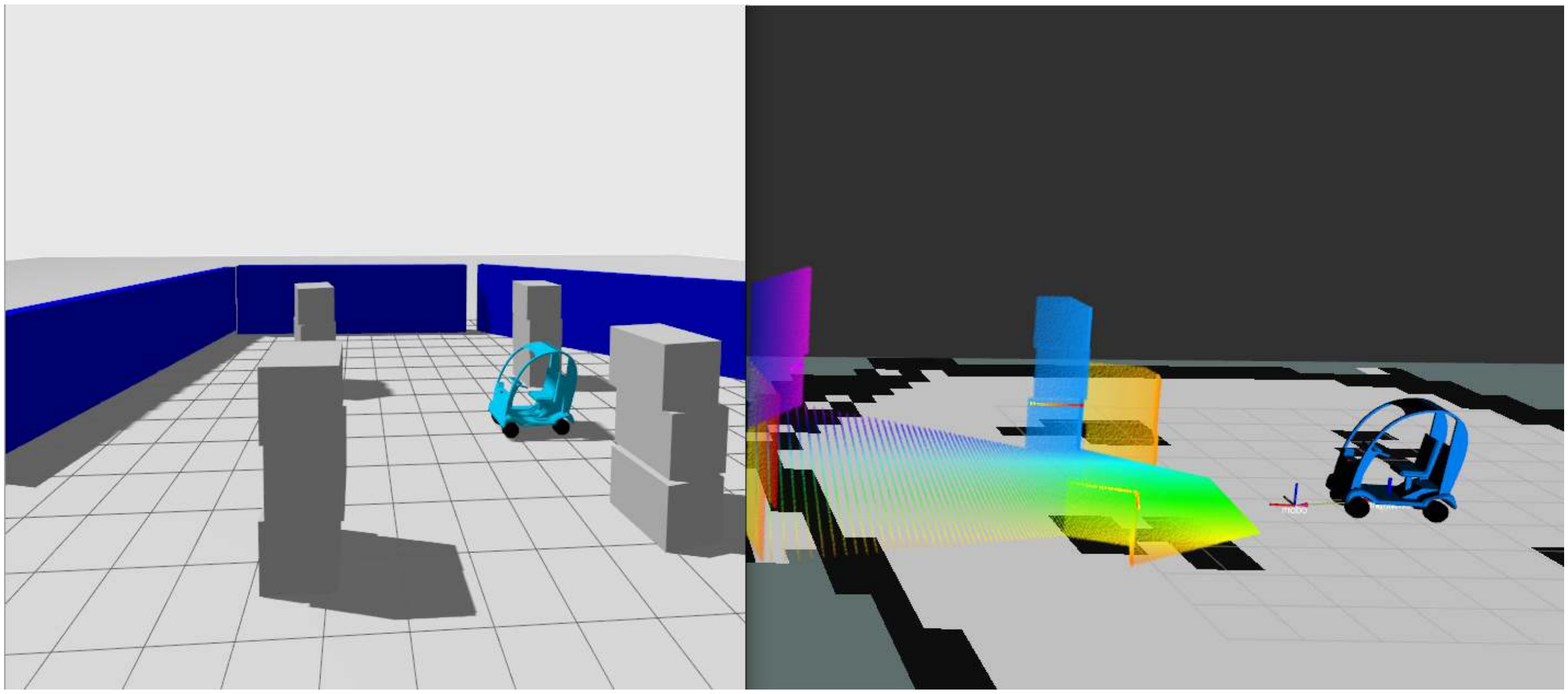

Figure 45: Simulated world with map in `rviz` along with `PointCloud2` for camera sensor

The `pod2_navigation` package consists all the scripts referenced and modified from the official release [23]. For simulation tests, we have tested the different level of the stack in simple custom world created to monitor how the podcar performs in tight environments. For this, the package also involves custom built maps. `nav2` is then launched on the prebuilt map along with SLAM in localization mode. Furthermore, the `nav2` stack is also tested without providing the any pre-built map and performing actual SLAM with autonomous exploration. Figure 45 illustrates the world in Gazebo with RGBD camera mounted in front of OpenPodcar2 and correspondingly the PointCloud2 and LaserScan topics in a map built using RTAB-Map visualized in rviz2.

## Appendix E

# OpenPodcar2 Hardware architecture

This document contains a technical description of how the hardware design works. This document is not needed to perform a build or use the system, but is intended for designers who wish to modify the hardware, service broken builds, perform a safety audit, or otherwise understand the design.

## System Architecture

OpenPodcar2 is organised around the R4 control board as the central hub for power distribution, safety interlocking, and actuator control. The inbuilt 24V battery feeds the main circuit breaker and fuse protection, then supplies the R4 24V input and the traction power branch. From the 24V bus, DC/DC converters generate regulated 12V, 5V, and 19V rails for drivers, relays, cooling, communications, and auxiliary devices. Motion control is split into propulsion and steering: the R4 drives the OSMC for the traction motor and the DHB12 for the steering actuator. A dead man's handle and motor relay sit in the enable chain so propulsion power is only present when safety conditions are satisfied.

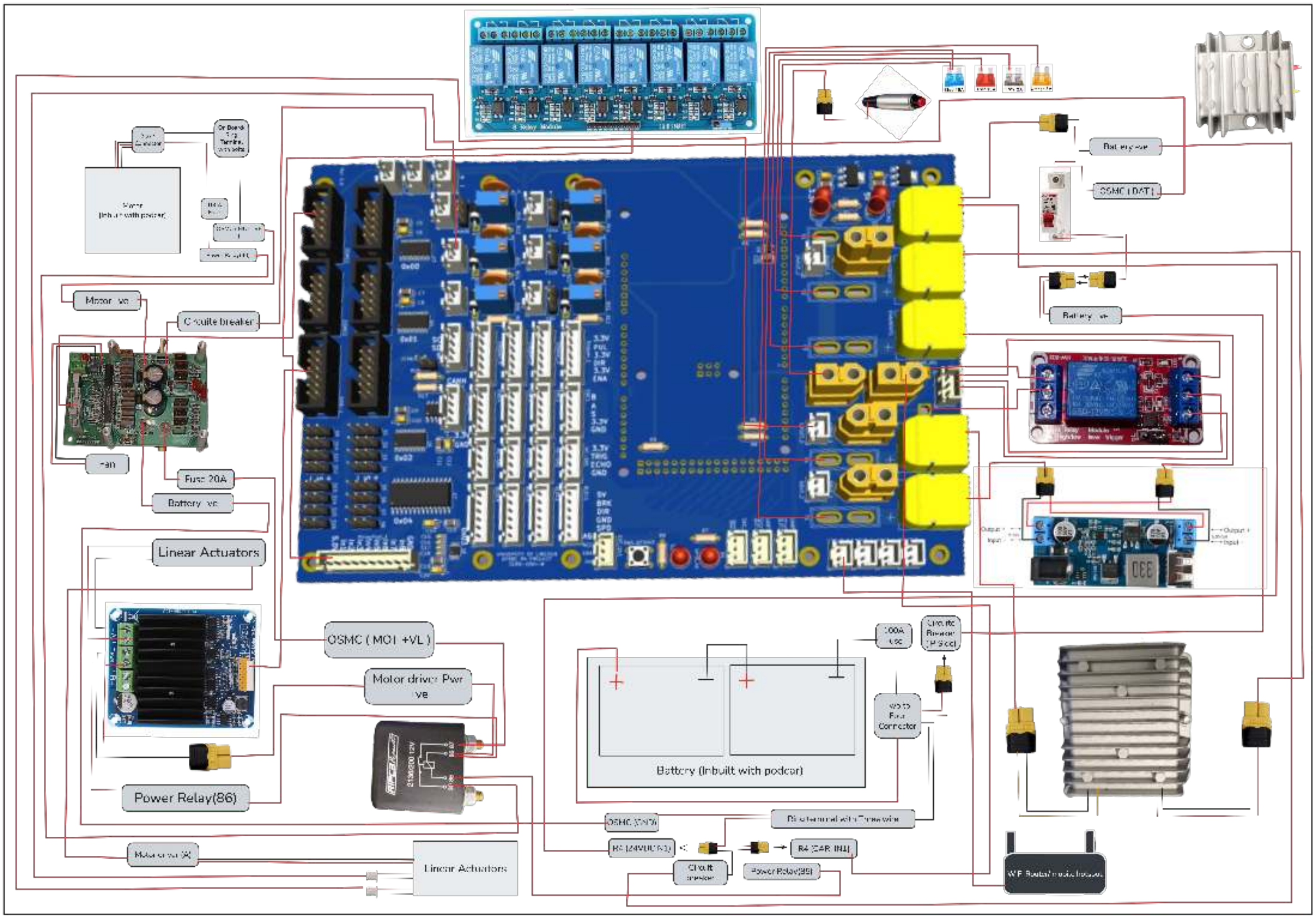

Figure 46: System Architecture

## Physical Integration

All board mounted modules are secured using standoffs for insulation, service access, airflow, and vibration resistance, with strain relief applied at power connectors and ribbon interfaces. Airflow is especially important for components which get hot such as the motor drivers OSMC and DHB12.

The mounting board's layered arrangement creates dedicated routing space between boards, allowing wiring looms to pass cleanly between layers for improved tidiness, strain relief, and maintainability.

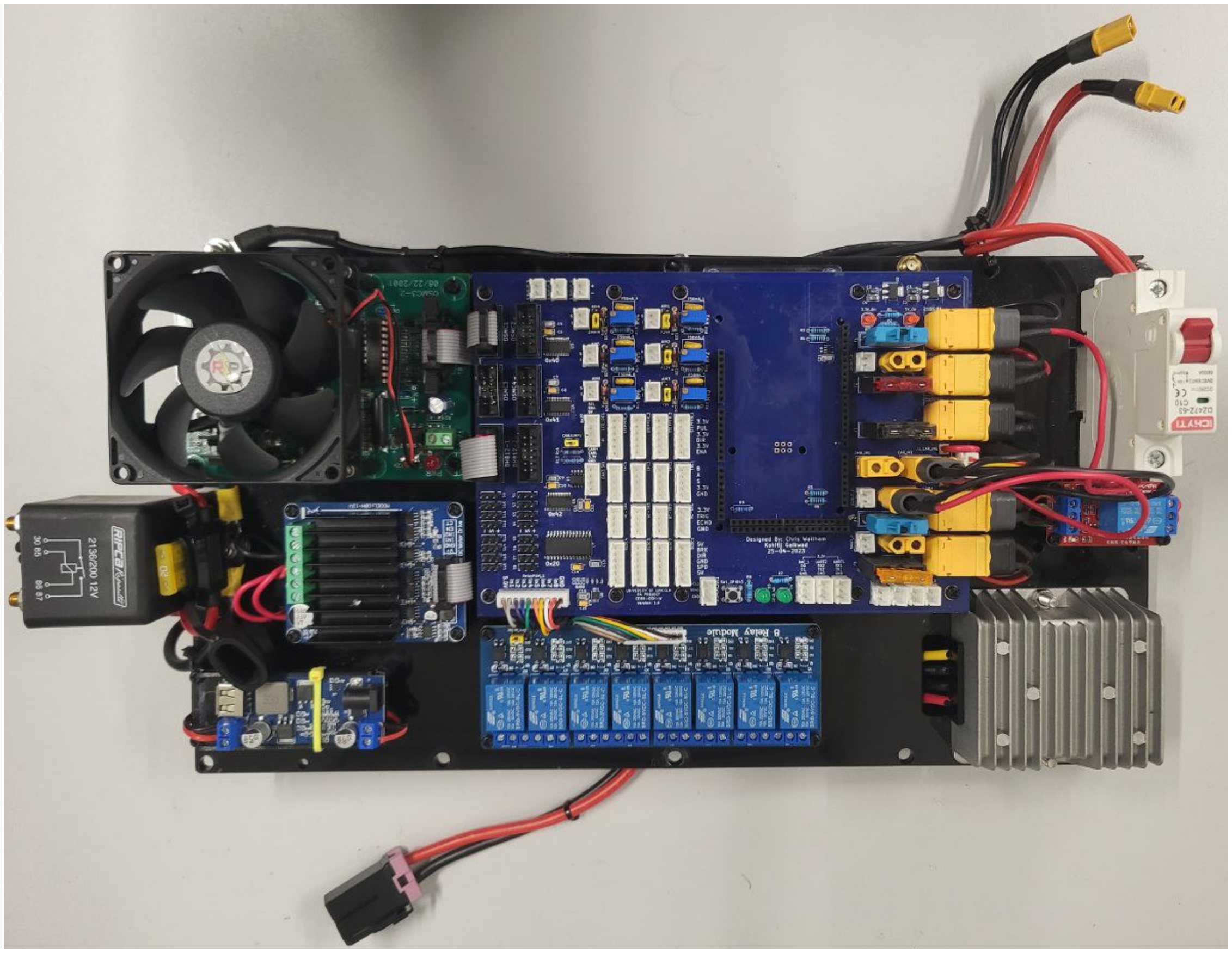

Figure 47: Top view of mounting board assembly

## Power systems

### Battery

The donor vehicle's original lead acid battery is the main 24V power source for the entire OpenPodcar2 system. It supplies energy to the propulsion motor (via the OSMC), steering (via DHB12), computation (R4, laptop) and communication (router). The battery's negative terminal forms the shared ground reference for all systems.

Compared to 12V, a 24V system reduces current draw, limits heat losses, and supports stable delivery of high-surge currents during acceleration. As the first element in the power chain, the battery also anchors the safety architecture—its output feeds directly into the circuit breaker, traction relay, and fuse network. A stable battery ensures predictable behaviour of steering, propulsion, and control electronics.

When batteries are low on change, they gradually reduce the voltage, so the 24V gradually falls to lower voltages.

This affects voltages throughout the system, so the charge level should be monitored and the vehicle should be recharged and not used when it starts to fall. R4's safety checks will be triggered if low voltages disturb normal operation and will automatically cut power to the vehicle via the R4 level DMH.

Batteries are dangerous if they are short-circuited, as very high current is then supplied causing great heat. The 100A battery fuse is used as the ultimate mitigation against this in the event of a short circuit. All connections to the battery go via this fuse.

#### Circuit breaker

The circuit breaker connects in series to the battery after the 100A fuse, as acts as an earlier line of defense, with all components connecting via it. Unlike the fuse which has to be physically replaced on blowing, the circuit breaker flips a switch which can be manually reset. Its switch also acts as the overall on/off switch for the system to end users, both to turn on and off for normal use and also to enable quick power off during testing (including bench testing) in the event of any problems.

In addition to short circuits, the circuit breaker protects against the motor drivers and other devices from trying to draw very high currents under heavy load, which can cause thermal damage, electrical fires, and catastrophic system failure.

The breaker uses thermal-magnetic protection, combining two mechanisms:

- **Magnetic trip:** activates instantly during short circuits or extreme fault currents.
- **Thermal trip:** responds to sustained overloads by heating a bimetallic strip until it bends and releases the latch.

This two-stage design allows the breaker to respond proportionally to fault severity. When tripped, the internal contacts physically separate, interrupting the high-current path and preventing further damage. Because the OpenPodCar uses a 24V battery with potentially large current reserves, fast and reliable trip behavior ensures the safety of both the hardware and operator.

The IP (input) side connects directly to the battery terminals. The switched side outputs power to the R4's 24VDCIN1 input and also to the OSMC BAT input. Large-gauge cables and ring terminals are used for secure high-current transfer.

#### Further fuses

A fuse contains a metal filament that melts when current surpasses its rated limit, instantly interrupting power flow. This stops excessive heat and prevents downstream damage. Different fuse ratings (2A, 5A, 10A, 15A, 20A, 100A) are used depending on how much current each subsystem normally draws.

Fuses protect the OpenPodcar2 from electrical faults by breaking the circuit when current exceeds safe levels. They prevent damage from short circuits, wiring failures, or overloaded components such as the OSMC motor driver, DC–DC converters, or steering actuator. By placing fuses at key points in both low- and high-voltage circuits, the system ensures that a local issue cannot propagate into a larger system failure. Each fuse rating matches the expected current load of the subsystem it protects, providing targeted isolation and improving overall safety.

Fuses are placed both on the R4 (low-current logic protection) and in the high-current traction pathway (100A fuse for the propulsion motor). Automotive blade fuses are uses as they are easily available and are color-coded for easy finding and replacement.

### DC/DC Converter 24→12V

The 24→12V DC–DC converter's role is to take the 24V supply from the main traction battery and down-convert it to a stable 12V rail that powers all medium-voltage subsystems, including the steering motor driver (DHB12), relay coils, fans, auxiliary sensors, Wi-Fi module, and other electromechanical devices. The 12V subsystem is effectively the backbone of vehicle actuation because both steering and traction-related control elements rely on it. Without a clean and reliable 12V supply, the vehicle would suffer inconsistent steering, relay chatter, and voltage drop-related faults that could cause unsafe behaviour. Thus, this converter ensures continuous regulation even under heavy and rapidly changing loads produced by the actuator and relay switching.

This follows the R4 external supply workflow, where a high current output is routed to an external buck converter and the regulated rail is returned to the board through the 12V input connector. R4 can then distribute the 12V to other components.

In mobile robotics, 12V rails are particularly sensitive to noise produced by high-current propulsion motors. Large inductive spikes from the traction system can propagate back into the shared battery lines. The converter isolates these disturbances by providing regulated output and input filtering, preserving stable voltage for the steering and control subsystems. Its mechanical placement on a heatsink ensures thermal stability during prolonged operation.

The converter uses a high-efficiency buck-regulator topology. Inside, a switching MOSFET repeatedly connects and disconnects the 24V input at high frequency (typically hundreds of kHz). This pulsed waveform is filtered through an LC network to produce a smooth 12V DC output. Because switching regulators operate efficiently even at high load currents, they can provide stable performance with minimal heating. The converter's feedback loop continuously monitors the output voltage and adjusts MOSFET duty cycle to compensate for load changes. This ensures that even when the steering actuator draws sudden bursts of current, the 12V rail remains stable and noise-free. Over-current, thermal shutdown, and short-circuit protection mechanisms further safeguard the vehicle.

The converter connects to the 24V battery bus through XT60 connectors for safe high-current handling. Its output feeds the 12IN1 and 24to12out1 connectors on the R4. Mechanically, it is mounted on a heatsink plate with airflow clearance.

### DC/DC 24→5V

Stable 5V power is essential because logic circuitry is highly sensitive to voltage fluctuations caused by motor loads or sudden current surges. The converter prevents brownouts, resets, and sensor malfunctions that could otherwise compromise safety during autonomous navigation. Its placement close to the R4 further reduces voltage drop and mitigates noise coupling into the microcontroller domain.

The 24→5V DC–DC converter is a critical power-regulation module responsible for supplying all 5V logic-level electronics in the OpenPodcar2 system. This includes microcontrollers, relay boards, sensors, communication interfaces, and auxiliary digital circuits. Because the vehicle operates from a 24V battery system—typical for mobility platforms and medium-sized robotics—directly powering 5V devices from the battery would result in instant damage. This converter therefore ensures safe voltage conditioning, electrical isolation, and noise filtering for low-voltage subsystems.

The converter is built around a high-efficiency buck regulator (e.g., XL4016), which uses high-frequency switching to step down 24V to a stable 5V output. Unlike linear regulators, which dissipate heat, switching regulators achieve >90% efficiency, making them suitable for continuous operation under load. Internal MOSFETs switch rapidly—typically 200–400 kHz—while an LC filter smooths the output. The built-in potentiometer allows precise tuning of output voltage to exactly 5.00V to protect downstream logic.

Thermal management is handled via an aluminum heatsink that dissipates switching losses. The converter tolerates voltage spikes coming from the traction motor system, and its input capacitors buffer transients, preventing them from reaching 5V devices.

The converter is mounted on a heatsink plate and uses XT60 connectors for secure high-current connections. Its output feeds the 5V rail on the R4. The wiring path is kept short to maintain voltage stability and reduce EMI.

### DC/DC 24→19V Converter

The 24 to 19 V converter provides an auxiliary regulated supply derived from the 24 V bus, intended to power external devices such as an onboard laptop, indicators, or other add on loads that require a stable 19 V input. It operates as a switch mode buck regulator that steps down the 24 V input to a controlled 19 V output with filtered ripple, reducing the impact of battery variation on sensitive electronics.

## Actuation

### Linear Actuator for steering

The GLA750 P linear actuator is the steering positioner for OpenPodcar2, converting commanded electrical drive into controlled linear stroke that sets the front wheel steering angle. Motion is produced by an internal DC motor driving a lead screw so extension and retraction are achieved by reversing motor polarity, with speed set by pulse width modulation from the steering driver. An integrated position feedback element provides a continuous signal proportional to stroke, allowing the R4 to close the steering loop and hold repeatable angles under load.

The GLA750-P linear actuator serves as the core element of the OpenPodcar2's steering system, enabling precise, controllable, and repeatable linear motion that translates directly into steering angle changes. As steering is a safety-critical subsystem, the actuator must maintain reliability under diverse load conditions, such as static resistance when the vehicle is stationary, dynamic steering forces when turning at speed, and vibration from rough terrain. The actuator provides high thrust, stable movement, and internal position sensing, making it ideal for autonomous vehicle steering applications where closed-loop control accuracy is essential. Its integration with the R4 and DHB12 ensures that ROS2 navigation commands can be executed with fidelity, supporting consistent trajectory tracking, obstacle avoidance, and human-robot interaction experiments.

Internally, the actuator consists of a DC motor connected to a leadscrew mechanism. When the DHB12 motor driver energizes the motor in one direction, the leadscrew rotates to extend the pushrod; reversing polarity causes retraction. The actuator's gearing provides mechanical advantage, allowing it to generate strong linear force while maintaining fine resolution of movement. Embedded within the actuator is a feedback device, typically a potentiometer or incremental encoder. This sensor outputs an analog voltage or pulse signal proportional to the actuator's extension length. The R4 reads this feedback through its ADC (AIN5), enabling the firmware to determine the exact steering position at all times. This closed-loop capability allows the system to command precise steering angles while compensating for mechanical drift, load changes, or environmental resistance.

High-current motor wires connect directly to the DHB12's A-channel terminals. Feedback wiring, including power (+12V), ground, and encoder channels, is routed to the R4. All connections use JST and screw-terminal connectors to ensure security and noise immunity.

### Traction motor and brake

The Pihsiang M3-9XL-D traction motor of the donor vehicle has a brake attachment, which is retained in OpenPodcar2 in its original form. The brake is applied by default and is removed when current is flowing though its coil. The vehicle is thus immobilized both when no current is being sent to the motor (e.g. when stopped on a hill) and also when a mechanical handbrake lever is applied which switches to break this current.

The motor has two terminals are referred to as Motor:+ve (red) and Motor:-ve (black). They are exposed in the 2 pins of the 4-to-2 connector in the vehicle's motor controller compartment, which is tapped by the OpenPodcar2 design.

The motor assembly and its connector (to the original donor controller, which is removed in OpenPodcar2) are shown in Figure 48. (This image is from a disassembled vehicle; there is no need to perform such as disassembly in the OpenPodcar2 build.)

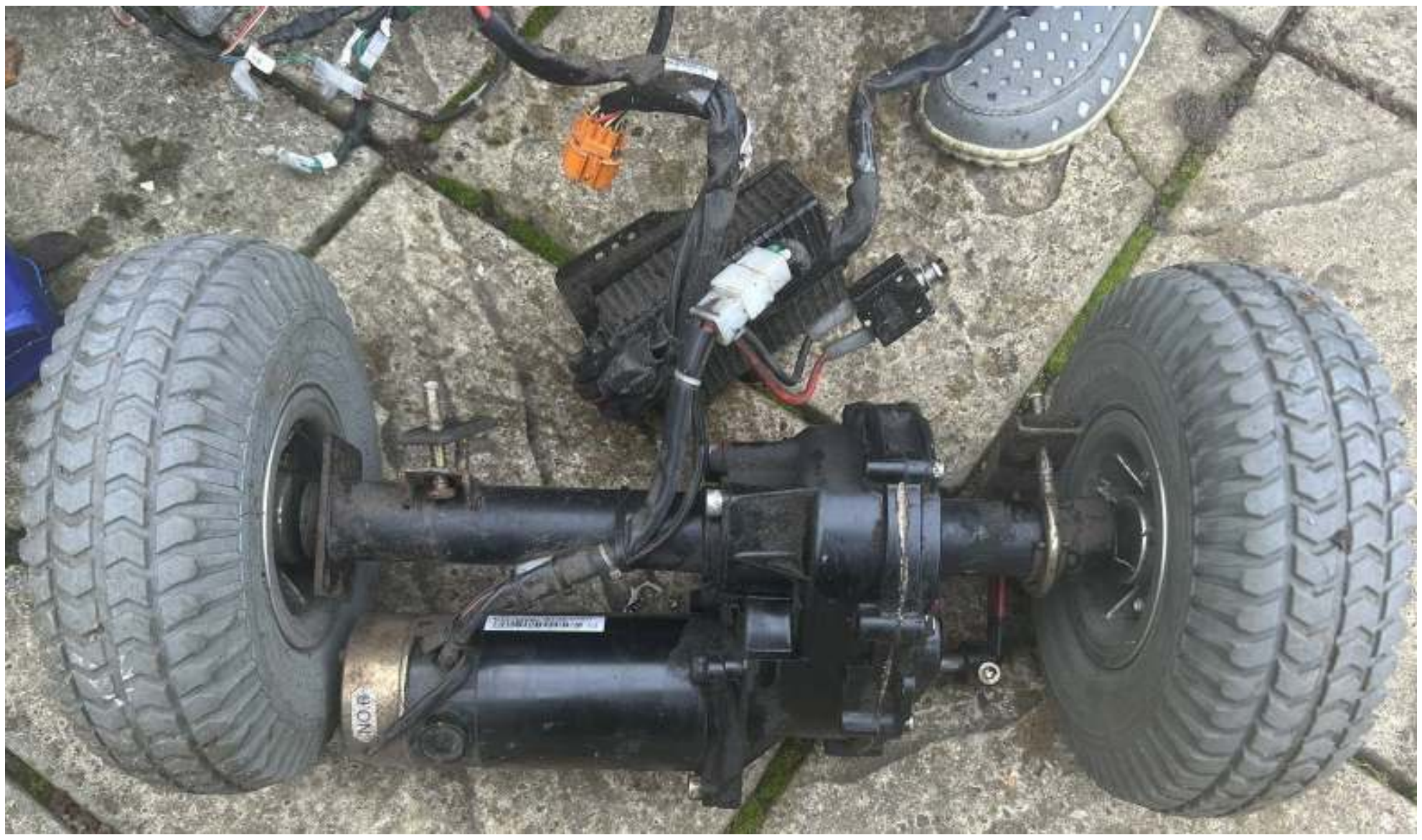

Figure 48: Donor motor assembly

## Motor Drivers

### Steering Motor Driver DHB12

The DHB12 motor driver is responsible for controlling the GLA750-P linear actuator that provides steering for the OpenPodcar2. This actuator requires reversible high-current DC power, and the DHB12 enables precise bidirectional control using commands from the R4. Because steering is a safety-critical subsystem, the DHB12 must reliably interpret the PWM and direction signals generated by the R4 firmware and convert them into appropriate voltage and current delivered to the actuator. The DHB12 is designed to operate within the 5–15V range, making it ideal for integration with the system's regulated 12V rail.

The DHB12 functions as a dual-channel H-bridge motor driver. An H-bridge uses four MOSFET switches arranged in a topology that allows controlled reversal of current flow through a DC motor. When the R4 outputs a PWM signal to the DHB12, the driver modulates motor speed by adjusting duty cycle. Direction is controlled by the DIR signal line, which determines the polarity of the current flowing through the actuator. Internally, the DHB12 uses gate-driver circuits to rapidly switch MOSFETs, ensuring efficient and low-heat operation even under load. Protection features such as thermal shutdown and current limiting help safeguard both the driver and the actuator from electrical faults.

The steering actuator often experiences mechanical load variations, especially when the vehicle is stationary or when resisting torque from the steering geometry. The DHB12 manages these variations by maintaining stable output voltage while dynamically responding to PWM updates from the R4. This results in smooth articulation without jitter or overshoot.

The DHB12 receives low-voltage control through a ribbon cable from the R4's DHB12-1 port. High-current actuator wires connect directly to the driver's A outputs. Power is supplied from the R4's 12OUT1 rail, ensuring isolated and regulated input. The module is mounted on the main board with adequate ventilation to avoid thermal buildup.

### OSMC

The OSMC (Open Source Motor Controller) serves as the primary propulsion driver for the OpenPodcar2 and is responsible for delivering high-current, bi-directional power to the vehicle's traction motor. Propulsion is the largest energy-consuming subsystem on the platform, meaning the motor driver must tolerate high instantaneous currents, voltage spikes, inductive loads, and thermal stresses. The OSMC is specifically designed for open-hardware robotics applications, making it ideal for research environments where reproducibility and modifiability are essential. Its rugged MOSFET-based design enables it to drive motors operating between 13–50V, with peak currents well above 100A depending on MOSFET configuration. This makes it suitable for medium-sized robotic vehicles like OpenPodCar2, which operates from a 24V battery and requires strong torque to move a human payload.

The OSMC is tightly integrated with the R4, receiving PWM and direction control signals while adhering to the platform's safety hierarchy. The presence of a hardware relay and circuit breaker upstream ensures that even if the OSMC were to fail, propulsion power can be physically disconnected. This layered design is consistent with safety standards in autonomous vehicle research.

Internally, the OSMC is built around a high-current H-bridge composed of large parallel MOSFETs. These MOSFETs act as electronic switches that route current through the traction motor in either forward or reverse direction. The R4 controls the OSMC by sending:

- PWM signals to regulate motor speed via duty-cycle modulation
- DIR signals to determine the rotation direction

The OSMC includes gate drivers that ensure fast MOSFET switching, reducing switching losses and heat generation. Because traction motors produce large inductive kickback, the OSMC incorporates flyback diodes, snubber networks, and multi-layer copper planes to handle transients safely. Its thermal characteristics are optimized through a thick PCB copper pour and heatsink mounting provisions. Additionally, the OSMC's built-in fan keeps the module stable under sustained load, preventing thermal runaway.

The OSMC does not implement closed-loop control internally—speed and feedback loops are handled at the ROS2 level via the R4 firmware and Nav2 stack. This separation aligns with modular open-hardware architecture: the motor driver handles raw power delivery, while intelligence resides in software.

The OSMC receives its control signals from the R4 via ribbon cable (CN5). Its power input is routed through the circuit breaker and motor relay, ensuring that human-level safety always overrides software. The motor terminals (MOT+ and MOT–) connect directly to the traction motor using bolted high-current connectors.

The fan power is provided by a 12V auxiliary line.

### Dead Man Handle (DMH)

The Dead Man Handle (DMH) is a key safety switch that ensures the vehicle can only move when the operator is actively holding or pressing it. Its role is to prevent unintended motion during testing, software faults, or communication loss. If the operator releases the DMH, propulsion is immediately disabled, making it an essential manual override in any research robot.

A simple two-wire connection links the physical DMH handheld button to the R4. The DMH is a normally-open momentary switch. When pressed, it sends a HIGH signal to the R4's DMH_IN1 input.

The R4 firmware continuously checks this signal. If it drops LOW—even briefly—R4 firmware de-energizes the 5V relay, which in turn ensures that all motor outputs are stopped. This ensures instant shutdown independent of ROS2 or onboard computing (though reliant on R4 firmware).

The 5V Relay Switch provides an electrically isolated interface between the low-voltage digital control domain of the R4 and higher-voltage enable or safety circuits within the OpenPodcar2 platform. Its primary purpose is to allow 5V logic signals generated by the R4 firmware to assert or remove permission for downstream circuits operating at elevated voltage levels without direct electrical coupling. This isolation protects the R4 from inductive transients, load-side faults, and ground disturbances originating from the power domain. Within the system safety architecture, the relay forms part of the traction enable chain, ensuring that propulsion authority is granted only when commanded by the controller and permitted by upstream safety conditions. The relay operates using an electromagnetic coil mechanism: when the control input is asserted, the coil energises and closes the normally open contact to the common terminal, completing the external control path, and when deasserted, the contact opens to remove the enable signal.

The Power Relay is the central safety and control component that enables OSMC to send power to the traction motor. Unlike software-layer controls, which merely command motor speed or direction, this relay provides a physical, hard-switch mechanism capable of completely disconnecting the main traction circuit. This design follows the standard in industrial automation and automotive engineering, where actuators must never rely solely on software for safety. In the OpenPodcar2, the relay forms part of a layered safety architecture together with the Dead Man Handle, circuit breaker, and R4 heartbeat watchdog. Only when the DMH is engaged and the R4 authorizes propulsion does the Power Relay energize, allowing current to flow to the traction motor. This ensures the vehicle cannot move unintentionally during startup, reboot, or communication dropouts.

Beyond safety, Power Relay also supports controlled startup sequencing. High-current DC motors produce large inrush currents, and energizing the relay only after system initialization helps protect downstream electronics. The relay's isolation between coil and switch sides prevents electrical noise and back-EMF from entering the low-voltage logic system.

Internally, the Power Relay consists of two isolated subsystems: a low-voltage electromagnetic coil (12V) and a set of high-current switching contacts. When the R4 energizes the coil by driving CAR_IN1 HIGH, current flows through the coil windings, generating a magnetic field that pulls the internal armature, closing the connection between terminals 30 and 87. This closure completes the 100A-capable traction circuit, enabling the battery's voltage to reach the OSMC input through the fused path.

If the R4 removes power from the coil, or if the DMH is released by the operator, the coil de-energizes and the armature springs back, opening the high-current path. This instantly cuts power to propulsion regardless of software state. Because the coil consumes relatively little current, the relay can operate continuously while maintaining a cool thermal profile. The high-current contacts are rated for the startup demands of DC traction motors and are designed to withstand arcing through hardened contact surfaces.

### 8-Way Relay Bank

The 8–way relay bank is not currently used to control the vehicle but is provided for future use.

It provides eight individually controllable electromagnetic relays, allowing the R4 to switch multiple medium-voltage circuits independently. This module is essential in a research platform like the OpenPodcar2 because it enables modular control over numerous subsystems—lighting, auxiliary sensors, ignition logic paths, fans, experimental add-ons, and future hardware expansions—without redesigning the power architecture. Each relay channel acts as an electrically isolated switch capable of safely toggling 12–24V loads while being triggered by 5V logic originating from the R4 GPIO expander. This separation ensures that high-voltage loads never reach the microcontroller domain, maintaining system reliability and operator safety.

The relay bank also standardizes wiring by consolidating multiple switching elements into a single board. Instead of scattering relays across the vehicle, all switching logic is centralized, simplifying debugging, replacement, and documentation. Because the OpenPodCar is an open-hardware platform intended for reproducibility and experimentation, having an 8-channel relay module allows researchers to add or remove components—such as additional actuators, controllers, or testing modules—without altering the main control board design.

Each relay on the board contains a 5V electromagnetic coil that actuates a single-pole changeover switch. When the R4 drives an INx pin HIGH, current energizes the corresponding coil, creating a magnetic field that draws an internal armature, switching the load side from NC (Normally Closed) to NO (Normally Open). This allows selective control of each external circuit. Flyback diodes on the board absorb inductive spikes generated by coil de-energization, protecting the R4's GPIO expander. The VCC and GND terminals power the coils, while each INx line corresponds to one relay channel.

The relay bank receives its 5V coil power and logic control signals from the R4's RelayPort_1 connector. The load terminals for each relay are wired depending on whether the downstream device is intended to be normally powered (NC) or normally off (NO). Heavy loads are kept within rated limits to avoid relay contact wear.